\documentclass{article}

    \PassOptionsToPackage{numbers, compress}{natbib}
\usepackage[final]{neurips_2021}

\usepackage{titletoc} 

\usepackage[utf8]{inputenc} 
\usepackage[T1]{fontenc}    
\usepackage{hyperref}       
\usepackage{url}            
\usepackage{booktabs}       
\usepackage{amsfonts}       
\usepackage{nicefrac}       
\usepackage{microtype}      
\usepackage{graphicx} 

\usepackage{amsmath}
\usepackage{amssymb}
\usepackage{bm}
\usepackage[dvipsnames]{xcolor}
\hypersetup{colorlinks=true,allcolors=NavyBlue}

\newtheorem{conj}{Conjecture}

\DeclareMathOperator*{\tr}{tr}

\DeclareMathOperator*{\var}{var}
\DeclareMathOperator{\cov}{cov}

\title{Asymptotics of representation learning \\ in finite Bayesian neural networks}

\author{
  Jacob A. Zavatone-Veth$^{1,2}$, Abdulkadir Canatar$^{1,2}$, Benjamin S. Ruben$^{3}$, \\ \textbf{Cengiz Pehlevan}$^{2,4}$ \\
  \textsuperscript{1}Department of Physics, \textsuperscript{2}Center for Brain Science, \textsuperscript{3}Biophysics Graduate Program, \\ \textsuperscript{4}John A. Paulson School of Engineering and Applied Sciences \\
  Harvard University\\
  Cambridge, MA 02138 \\
  \texttt{\{jzavatoneveth,canatara,benruben\}@g.harvard.edu} \\ \texttt{cpehlevan@seas.harvard.edu} 
}

\begin{document}

\maketitle

\begin{abstract}
Recent works have suggested that finite Bayesian neural networks may sometimes outperform their infinite cousins because finite networks can flexibly adapt their internal representations. However, our theoretical understanding of how the learned hidden layer representations of finite networks differ from the fixed representations of infinite networks remains incomplete. Perturbative finite-width corrections to the network prior and posterior have been studied, but the asymptotics of learned features have not been fully characterized. Here, we argue that the leading finite-width corrections to the average feature kernels for any Bayesian network with linear readout and Gaussian likelihood have a largely universal form. We illustrate this explicitly for three tractable network architectures: deep linear fully-connected and convolutional networks, and networks with a single nonlinear hidden layer. Our results begin to elucidate how task-relevant learning signals shape the hidden layer representations of wide Bayesian neural networks. 
\end{abstract}

\section{Introduction}\label{introduction}

The expressive power of deep neural networks critically depends on their ability to learn to represent the features of data \cite{goodfellow2016deep,lecun2015deep,neal1996priors,williams1997computing,lee2018deep,matthews2018gaussian,novak2019bayesian,yang2019scaling,yang2020feature,aitchison2020bigger,wilson2020bayesian,yaida2020,halverson2021neural,antognini2019finite,naveh2020predicting,li2021statistical,zv2021exact,dyer2020asymptotics,aitken2020asymptotics,wenzel2020cold,fortuin2021bayesian,izmailov2021bayesian,mackay1992practical,jacot2018neural}. However, the structure of their hidden layer representations is only theoretically well-understood in certain infinite-width limits, in which these representations cannot flexibly adapt to learn data-dependent features \cite{neal1996priors,williams1997computing,lee2018deep,matthews2018gaussian,novak2019bayesian,yang2019scaling,yang2020feature,aitchison2020bigger,wilson2020bayesian,jacot2018neural}. In the Bayesian setting, these representations are described by fixed, deterministic kernels \cite{neal1996priors,williams1997computing,lee2018deep,matthews2018gaussian,novak2019bayesian,yang2019scaling,yang2020feature,aitchison2020bigger,wilson2020bayesian}. As a result of this inflexibility, recent works have suggested that finite Bayesian neural networks (henceforth BNNs) may generalize better than their infinite counterparts because of their ability to learn representations \cite{aitchison2020bigger}.

Theoretical exploration of how finite and infinite BNNs differ has largely focused on the properties of the prior and posterior distributions over network outputs \cite{yaida2020,halverson2021neural,antognini2019finite,naveh2020predicting,li2021statistical,zv2021exact}. In particular, several works have studied the leading perturbative finite-width corrections to these distributions \cite{yaida2020,halverson2021neural,antognini2019finite,naveh2020predicting,li2021statistical}. Yet, the corresponding asymptotic corrections to the feature kernels, which measure how representations evolve from layer to layer, have only been studied in a few special cases \cite{li2021statistical}. Therefore, the structure of these corrections, as well as their dependence on network architecture, remain poorly understood. In this paper, we make the following contributions towards the goal of a complete understanding of feature learning at asymptotically large but finite widths:
\begin{itemize}
    \item 
    We argue that the leading finite-width corrections to the posterior statistics of the hidden layer kernels of any BNN with a linear readout layer and Gaussian likelihood have a largely prescribed form (Conjecture \ref{conj1}). In particular, we argue that the posterior cumulants of the kernels have well-defined asymptotic series in terms of their prior cumulants, with coefficients that have fixed dependence on the target outputs. 

    \item 
    We explicitly compute the leading finite-width corrections for deep linear fully-connected networks (\S\ref{sec:linear}), deep linear convolutional networks (\S\ref{sec:conv}), and networks with a single nonlinear hidden layer (\S\ref{sec:nonlinear}). We show that our theory yields quantitatively accurate predictions for the result of numerical experiment for tractable linear network architectures, and qualitatively accurate predictions for deep nonlinear networks, where quantitative analytical predictions are intractable. 

\end{itemize}

Our results begin to elucidate the structure of learned representations in wide BNNs. The assumptions of our general argument are satisfied in many regression settings, hence our qualitative conclusions should be broadly applicable.

\section{Preliminaries} \label{sec:preliminaries}

We begin by defining our notation, setup, and assumptions. We will index training and test examples by Greek subscripts $\mu,\nu,\ldots$, and layer dimensions (that is, neurons) by Latin subscripts $j,l,\ldots$. Layers will be indexed by the script Latin letter $\ell$. Matrix- or vector-valued quantities corresponding to a given layer will be indexed with a parenthesized superscript, while scalar quantities that depend only on the layer will be indexed with a subscript. Depending on context, $\Vert \cdot \Vert$ will denote the $\ell_2$ norm on vectors or the Frobenius norm on matrices. We denote the standard Euclidean inner product of two vectors $\mathbf{a},\mathbf{b} \in \mathbb{R}^{n}$ by $\mathbf{a} \cdot \mathbf{b}$. 

\subsection{Bayesian neural networks with linear readout}

Throughout this paper, we consider deep Bayesian neural networks with fully connected linear readout. Such a network $\mathbf{f}: \mathbb{R}^{n_0} \to \mathbb{R}^{n_d}$ with $d$ layers can be written as
\begin{align} \label{eqn:general_network}
    \mathbf{f}(\mathbf{x};W^{d},\mathcal{W}) = \frac{1}{\sqrt{n_{d-1}}} W^{(d)} \bm{\psi}(\mathbf{x};\mathcal{W}),
\end{align}
where the feature map $\bm{\psi}(\cdot ; \mathcal{W}) : \mathbb{R}^{n_0} \to \mathbb{R}^{n_{d-1}}$ includes all $d-1$ hidden layers, collectively parameterized by $\mathcal{W}$. Here, $\bm{\psi}$ can be some combination of fully-connected feedforward networks, convolutional networks, recurrent networks, et cetera; we assume only that it has a well-defined infinite-width limit in the sense of \S\ref{sec:gplimit}. We let the widths of the hidden layers be $n_{1}, n_{2}, \ldots, n_{d-1}$; we define the width of a convolutional layer to be its channel count \cite{novak2019bayesian}. We assume isotropic Gaussian priors over the trainable parameters \cite{goodfellow2016deep,lecun2015deep,neal1996priors,williams1997computing,lee2018deep,matthews2018gaussian,novak2019bayesian,yang2019scaling,yang2020feature,aitchison2020bigger,wilson2020bayesian,yaida2020,halverson2021neural,antognini2019finite,naveh2020predicting,li2021statistical,zv2021exact,dyer2020asymptotics,aitken2020asymptotics,wenzel2020cold,fortuin2021bayesian,izmailov2021bayesian,mackay1992practical}, with $W^{(d)}_{ij} \sim_{\textrm{i.i.d}} \mathcal{N}(0, \sigma_{d}^2)$ in particular. 

In our analysis, we fix an arbitrary training dataset $\mathcal{D} = \{(\mathbf{x}_{\mu},\mathbf{y}_{\mu})\}_{\mu=1}^{p}$ of $p$ examples. We define the input and output Gram matrices of this dataset as $[G_{xx}]_{\mu\nu} \equiv n_{0}^{-1} \mathbf{x}_{\mu} \cdot \mathbf{x}_{\nu}$ and $[G_{yy}]_{\mu\nu} \equiv n_{d}^{-1} \mathbf{y}_{\mu} \cdot \mathbf{y}_{\nu}$, respectively. For analytical tractability, we consider a Gaussian likelihood $p(\mathcal{D} \,|\,\Theta) \propto \exp(-\beta E)$ for
\begin{align} \label{eqn:cost}
    E(\Theta; \mathcal{D}) = \frac{1}{2} \sum_{\mu=1}^{p} \Vert \mathbf{f}(\mathbf{x}_{\mu}; \Theta) - \mathbf{y}_{\mu} \Vert^2,
\end{align}
where $\beta \geq 0$ is an inverse temperature parameter that sets the variance of the likelihood and $\Theta = \{W^{(d)},\mathcal{W}\}$ \cite{mackay1992practical}. We then introduce the Bayes posterior over parameters given these data:
\begin{align} \label{eqn:posterior}
    p(\Theta\,|\,\mathcal{D}) = \frac{p(\mathcal{D}\,|\,\Theta) p(\Theta)}{p(\mathcal{D})}; 
\end{align}
we denote averages with respect to this distribution by $\langle \cdot \rangle$. By tuning $\beta$, one can then adjust whether the posterior is dominated by the prior ($\beta \ll 1$) or the likelihood ($\beta \gg 1$). We will mostly focus on the case in which the input dimension is large and the training dataset can be linearly interpolated; the low-temperature limit $\beta \to \infty$ then enforces the interpolation constraint.

\subsection{The Gaussian process limit}\label{sec:gplimit}

We consider the limit of large hidden layer widths $n_{1},n_{2},\ldots,n_{d-1} \to \infty$ with $n_{0}$, $n_{d}$, $p$, and $d$ fixed. More precisely, we consider a limit in which $n_{\ell} = \alpha_{\ell} n$ for $\ell = 1,\ldots,d-1$, where $\alpha_{\ell} \in (0,\infty)$ and $n \to \infty$, as studied by \cite{neal1996priors,williams1997computing,lee2018deep,matthews2018gaussian,novak2019bayesian,yang2019scaling,yang2020feature,aitchison2020bigger,wilson2020bayesian,yaida2020,halverson2021neural,antognini2019finite,naveh2020predicting,zv2021exact,dyer2020asymptotics,aitken2020asymptotics,jacot2018neural} and others. Importantly, we note that size of $n_{0}$ relative to $n$ is unimportant for our results, whereas $n_{d}/n$ and $d/n$ must be small \cite{aitchison2020bigger,yaida2020,zv2021exact}. 

In this limit, for $\bm{\psi}$ built out of compositions of most standard neural network architectures, the prior over function values $\mathbf{f}$ tends to a Gaussian process (GP) \cite{neal1996priors,williams1997computing,lee2018deep,matthews2018gaussian,novak2019bayesian,yang2019scaling}. Moreover, with our choice of a Gaussian likelihood, the posterior over function values also tends weakly to the posterior induced by the limiting GP prior \cite{hron2020exact}. The kernel of the limiting GP prior is given by the deterministic limit $K_{\infty}^{(d-1)}$ of the inner product kernel of the postactivations of the final hidden layer,
\begin{equation} \label{eqn:featuremapkernel}
    K^{(d-1)}(\mathbf{x},\mathbf{x}') \equiv n_{d-1}^{-1} \bm{\psi}(\mathbf{x},\mathcal{W}) \cdot \bm{\psi}(\mathbf{x}',\mathcal{W}),
\end{equation}
multiplied by the prior variance $\sigma_{d}^{2}$ \cite{neal1996priors,williams1997computing,lee2018deep,matthews2018gaussian,novak2019bayesian,yang2019scaling}. For a broad range of network architectures, $K_{\infty}^{(d-1)}$ can be computed recursively \cite{lee2018deep,matthews2018gaussian,novak2019bayesian,yang2019scaling}. For brevity, we define the kernel matrix evaluated on the training data: $[K^{(d-1)}]_{\mu\nu} \equiv K^{(d-1)}(\mathbf{x}_{\mu},\mathbf{x}_{\nu})$. 

\section{Elementary perturbation theory for finite Bayesian neural networks}\label{sec:universal}

We first present our main result, which shows that the form of the leading perturbative correction to the average hidden layer kernels of a BNN is tightly constrained by the assumptions that the readout is linear, that the cost is quadratic, and that the GP limit is well-defined. 

\subsection{Finite-width corrections to the posterior cumulants of hidden layer observables}

Our main result is as follows:

\begin{conj}\label{conj1}

    Consider a BNN of the form \eqref{eqn:general_network}, with posterior \eqref{eqn:posterior}. Assume that this network admits a well-defined GP limit as discussed in \S\ref{sec:gplimit}. Let $O$ be a \emph{hidden layer observable}, that is, a function of the hidden layer activations that is not a function of the readout weights $W_{d}$. Assume that $O$ tends in probability to a finite, deterministic limit $O_{\infty}$ under the posterior in the GP limit. 
    
    Then, the posterior cumulants of this observable admit well-behaved asymptotic series at large widths in terms of its joint prior cumulants with the postactiviation kernel $K^{(d-1)}$. In particular, the asymptotic expansion of the posterior mean $\langle O \rangle$ has leading terms
    \begin{align} \label{eqn:generalkernel}
        \langle O \rangle = \mathbb{E}_{\mathcal{W}} O + \frac{1}{2} n_{d} \sum_{\rho,\lambda=1}^{p} [\sigma_{d}^{-2} \Gamma^{-1} G_{yy} \Gamma^{-1} - \Gamma^{-1}]_{\rho\lambda} \cov_{\mathcal{W}}(O, K^{(d-1)}_{\rho\lambda}) + \ldots,
    \end{align}
    where $\Gamma \equiv K^{(d-1)}_{\infty} + \beta^{-1} \sigma_{d}^{-2} I_{p}$. Here, the cumulants of the kernels are computed with respect to the \emph{prior}, and are themselves given by asymptotic series at large widths. The ellipsis denotes terms that are of subleading order in the inverse hidden layer widths. 

\end{conj}

In Appendix \ref{app:sec:perturbation}, we derive this result perturbatively by expanding the posterior cumulant generating function of $O$ in powers of the deviations of $O$ and $K^{(d-1)}$ from their deterministic infinite-width values. There, we also give an asymptotic formula for the posterior covariance of two observables. However, the resulting perturbation series may not rigorously be an asymptotic series, and this method does not yield quantitative bounds for the width-dependence of the terms. We therefore frame it as a conjecture. We note that similar methods can be applied to compute asymptotic corrections to the posterior predictive statistics; we comment on this possibility in Appendix \ref{app:sec:predictor}.

Though this conjecture applies to a broad class of hidden layer observables, the observables of greatest interest are the preactivation or postactivation kernels of the hidden layers within the feature map $\bm{\psi}$. We will focus on the postactivation kernels $K^{(\ell)}$, which measure how the similarities between inputs evolve as they are propagated through the network  \cite{lee2018deep,matthews2018gaussian,novak2019bayesian,yang2019scaling,yang2020feature,aitchison2020bigger}. 

Conjecture \ref{conj1} posits that there are two possible types of leading finite-width corrections to the average kernels. The first class of corrections are deviations of $\mathbb{E}_{\mathcal{W}} K^{(\ell)}$ from $K^{(\ell)}_{\infty}$. These terms reflect corrections to the prior, and do not reflect non-trivial representation learning as they are independent of the outputs. For fully-connected networks, also known as multilayer perceptrons (MLPs), work by \citet{yaida2020} and by Gur-Ari and colleagues \cite{dyer2020asymptotics,aitken2020asymptotics} shows that $\mathbb{E}_{\mathcal{W}} K^{(\ell)} = K^{(\ell)}_{\infty} + \mathcal{O}(n^{-1})$. The second type of correction is the output-dependent term that depends on $\cov_{\mathcal{W}}(K^{(\ell)}_{\mu\nu}, K^{(d-1)}_{\rho\lambda})$. For deep linear MLPs or MLPs with a single hidden layer, $\mathbb{E}_{\mathcal{W}} K^{(\ell)}$ is exactly equal to $K^{(\ell)}_{\infty}$ at any width (see Appendix \ref{app:sec:cov}) \cite{neal1996priors,yaida2020,dyer2020asymptotics}, and only the covariance term contributes. More broadly, these prior works show that $\cov_{\mathcal{W}}(K^{(\ell)}_{\mu\nu}, K^{(d-1)}_{\rho\lambda}) = \mathcal{O}(n^{-1})$ for MLPs, and that higher cumulants are of $\mathcal{O}(n^{-2})$ \cite{yaida2020,dyer2020asymptotics,aitken2020asymptotics}. Some of these results have recently been extended to convolutional networks by \citet{andreassen2020asymptotics}. Thus, the finite-width correction to the prior mean should not dominate the feature-learning covariance term, and the terms hidden in the ellipsis should indeed be suppressed. 

The leading output-dependent correction has several interesting features. First, it includes a factor of $n_{d}$, reflecting the fact that inference in wide Bayesian networks with many outputs is qualitatively different from that in networks with few outputs relative to their hidden layer width \cite{aitchison2020bigger}. If $n_{d}/n$ does not tend to zero with increasing $n$, the infinite-width behavior is not described by a standard GP \cite{yang2019scaling,aitchison2020bigger}. Moreover, we note that the matrix $\Gamma$ is invertible at any finite temperature, even when $K_{\infty}^{(d-1)}$ is singular. Therefore, provided that one can extend the GP kernel by continuity to non-invertible $G_{xx}$, Conjecture \ref{conj1} can be applied in the data-dense regime $n_{0} < p$ as well as the data-sparse regime $n_{0} > p$. Furthermore, we observe that the correction depends on the outputs only through their Gram matrix $G_{yy}$. This result is intuitively sensible, since with our choice of likelihood and prior the function-space posterior is invariant under simultaneous rotation of the output activations and targets. Finally, $G_{yy}$ is transformed by factors of the matrix $\Gamma^{-1}$, hence the correction depends on certain interactions between the output similarities and the GP kernel $K_{\infty}^{(d-1)}$.

\subsection{High- and low-temperature limits of the leading correction} \label{sec:templimits}

To gain some intuition for the properties of the leading finite-width corrections, we consider their high- and low-temperature limits. These limits correspond to tuning the posterior \eqref{eqn:posterior} to be dominated by the prior or the likelihood, respectively. At high temperatures ($\beta \ll 1$), expanding $\Gamma^{-1}$ as a Neumann series (see Appendix \ref{app:sec:tech} and \cite{horn2012matrix}) yields
\begin{align}
    \sigma_{d}^{-2} \Gamma^{-1} G_{yy} \Gamma^{-1} - \Gamma^{-1} = - \beta \sigma_{d}^{2} I_{p} + (\beta \sigma_{d}^{2})^2 (\sigma_{d}^{-2} G_{yy} + K_{\infty}^{(d-1)}) + \mathcal{O}[(\beta \sigma_{d}^{2})^{3}].
\end{align}
Thus, at high temperatures, the outputs only influence the average kernels of Conjecture \ref{conj1} to subleading order in both width and $\beta$, which reflects the fact that the likelihood is discounted relative to the prior in this regime. Moreover, the leading output-dependent contribution averages together $G_{yy}$ and $K_{\infty}^{(d-1)}$, hence, intuitively, there is no way to `cancel' the GP contributions to the average kernels. We note that, at infinite temperature ($\beta = 0$), the posterior reduces to the prior, and all finite-width corrections to the average kernels arise from the discrepancy between $\mathbb{E}_{\mathcal{W}} K^{(\ell)}$ and  $K_{\infty}^{(\ell)}$.  

At low temperatures ($\beta \gg 1$), the behavior of $\Gamma^{-1}$ differs depending on whether or not $K_{\infty}^{(d-1)}$ is of full rank. Assuming for simplicity that it is invertible, we have
\begin{align}
    \sigma_{d}^{-2} \Gamma^{-1} G_{yy} \Gamma^{-1} - \Gamma^{-1} = [K_{\infty}^{(d-1)}]^{-1} (\sigma_{d}^{-2} G_{yy} - K_{\infty}^{(d-1)}) [K_{\infty}^{(d-1)}]^{-1} + \mathcal{O}[(\beta \sigma_{d}^{2})^{-1}]; 
\end{align}
in the non-invertible case there are additional contributions involving projectors onto the null space of $K_{\infty}^{(d-1)}$. Therefore, the leading-order low temperature correction depends on the difference between the target and GP kernels, while the leading non-trivial high temperature correction depends on their sum.

\section{Learned representations in tractable network architectures}

Having derived the general form of the leading perturbative finite-width correction to the average feature kernels, we now consider several example network architectures. For these tractable examples, we provide explicit formulas for the feature-learning corrections to the hidden layer kernels, and test the accuracy of our theory with numerical experiments. 

\subsection{Deep linear fully-connected networks}\label{sec:linear}

We first consider deep linear fully-connected networks with no bias terms. Concretely, we consider a network with activations $\mathbf{h}^{(\ell)} \in \mathbb{R}^{n_{\ell}}$ recursively defined via $\mathbf{h}^{(\ell)} = n_{\ell-1}^{-1/2} W^{(\ell)} \mathbf{h}^{(\ell-1)}$ with base case $\mathbf{h}^{(0)} = \mathbf{x}$, where the prior distribution of weights is $[W^{(\ell)}]_{ij} \sim_{\textrm{i.i.d.}} \mathcal{N}(0,\sigma_{\ell}^{2})$. For such a network, the hidden layer kernels $[K^{(\ell)}]_{\mu\nu} \equiv n_{\ell}^{-1} \mathbf{h}_{\mu}^{(\ell)} \cdot \mathbf{h}_{\nu}^{(\ell)}$ have deterministic limits $K_{\infty}^{(\ell)} = m_{\ell}^{2} G_{xx}$, where $m_{\ell}^2 \equiv \sigma_{\ell}^2 \sigma_{\ell-1}^2 \cdots \sigma_{1}^2$ is the product of prior variances up to layer $\ell$. Higher prior cumulants of the kernels are easy to compute with the aid of Isserlis' theorem for Gaussian moments (see Appendix \ref{app:sec:cov}) \cite{isserlis1918formula,wick1950evaluation}, yielding
\begin{align} \label{eqn:linearkernel}
    \frac{\langle K^{(\ell)} \rangle }{m_{\ell}^2} = G_{xx} + \bigg(\sum_{\ell'=1}^{\ell} \frac{n_{d}}{n_{\ell'}}\bigg) G_{xx} \Gamma^{-1} \left(m_{d}^{-2} G_{yy} - \Gamma \right) \Gamma^{-1} G_{xx} + \mathcal{O}(n^{-2}),
\end{align}
where $\Gamma \equiv G_{xx} + I_{p}/(\beta m_{d}^{2})$ and $\ell=1,\ldots,d-1$. In Appendix \ref{app:sec:deeplinearkernel}, we show that this result can be derived directly through an \emph{ab initio} perturbative calculation of the cumulant generating function of the kernels, without relying on our heuristic argument for the general version of Conjecture \ref{conj1}. Moreover, in Appendix \ref{app:sec:skip}, we show that the form of the correction remains the same even if one allows arbitrary forward skip connections, though the dependence on width and depth is given by a more complex recurrence relation. 

Thus, the leading corrections to the normalized average kernels $\langle K^{(\ell)} \rangle / m_{\ell}^{2}$ are identical across all hidden layers up to a scalar factor that encodes the width-dependence of the correction. This sum-of-inverse-widths dependence was previously noted by \citet{yaida2020} in his study of the corrections to the prior of a deep linear network. For a network with hidden layers of equal width $n$, we have the simple linear dependence $\sum_{\ell'=1}^{\ell} (n_{d}/n_{\ell'}) = n_{d} \ell / n$. If one instead includes a narrow bottleneck in an otherwise wide network, this dependence predicts that the kernels before the bottleneck should be close to their GP values, while those after the bottleneck should deviate strongly.

This result simplifies further at low temperatures, where, by the result of \S\ref{sec:templimits}, we have
\begin{align} \label{eqn:lowtemplinear}
    \frac{\langle K^{(\ell)} \rangle}{m_{\ell}^2} = G_{xx} + \bigg(\sum_{\ell'=1}^{\ell} \frac{n_{d}}{n_{\ell'}}\bigg) \left( m_{d}^{-2} G_{yy}  - G_{xx} \right) + \mathcal{O}(n^{-2}, \beta^{-1})
\end{align}
in the regime in which $G_{xx}$ is invertible. We thus obtain the simple qualitative picture that the low-temperature average kernels linearly interpolate between the input and output Gram matrices. In Appendix \ref{app:sec:comp}, we show that this limiting result can be recovered from the recurrence relation derived through other methods by \citet{aitchison2020bigger}, who did not use it to compute finite-width corrections. We note that the low-temperature limit is peculiar in that the mean predictor reduces to the least-norm pseudoinverse solution to the underlying underdetermined linear system $XW=Y$; we comment on this property in Appendix \ref{app:sec:predictor}. 

We can gain some additional understanding of the structure of the correction by using the eigendecomposition of $G_{xx}$. As $G_{xx}$ is by definition a real positive semidefinite matrix, it admits a unitary eigendecomposition $G_{xx} = U \Lambda U^{\dagger}$ with non-negative eigenvalues $\Lambda_{\mu\mu}$. In this basis, the average kernel is 
\begin{align}
    \frac{1}{m_{\ell}^{2}} U^{\dagger} \langle K^{(\ell)} \rangle U = \Lambda + \bigg(\sum_{\ell'=1}^{\ell} \frac{n_{d}}{n_{\ell'}}\bigg) \left(m_{d}^{-2} \tilde{\Lambda} U^{\dagger} G_{yy} U \tilde{\Lambda} - \tilde{\Lambda} \Lambda \right) + \mathcal{O}(n^{-2}),
\end{align}
where we have defined the diagonal matrix $\tilde{\Lambda} \equiv \beta m_{d}^{2} \Lambda (I_{p} + \beta m_{d}^{2} \Lambda)^{-1}$. As $\beta m_{d}^{2} \Lambda \geq 0$, the diagonal elements of $\tilde{\Lambda}$ are bounded as $0 \leq \tilde{\Lambda}_{\mu\mu} \leq 1$. Thus, the factors of $\Gamma^{-1} G_{xx}$ by which $G_{yy}$ is conjugated have the effect of suppressing directions in the projection of $G_{yy}$ onto the eigenspace of $G_{xx}$ with small eigenvalues. We can see that this effect will be enhanced at high temperatures ($\beta \ll 1$) and small scalings ($m_{d}^{2} \ll 1$), and suppressed at low temperatures and large scalings. For this linear network, similarities are not enhanced, only suppressed. Moreover, if $G_{xx}$ is diagonal, then a given element of the average kernel will depend only on the corresponding element of $G_{yy}$.

We now seek to numerically probe how accurately these asymptotic corrections predict learned representations in deep fully-connected linear BNNs. Using Langevin sampling \cite{kloeden1992stochastic,paske2019pytorch}, we trained deep linear networks of varying widths, and compared the difference between the empirical and GP kernels with theory predictions. We provide a detailed discussion of our numerical methods in Appendix \ref{app:sec:numeric}. In Figure \ref{fig:linear_scaling}, we present an experiment with a 2-layer linear neural network trained on the MNIST dataset of handwritten digit images \cite{lecun2010mnist} using the Neural Tangents library \cite{neuraltangents2020}. We find an excellent agreement with our theory, confirming the inverse scaling with width and linear scaling with depth for the deviations from GP kernel.

\begin{figure}
    \centering
    \includegraphics[width=0.97\linewidth]{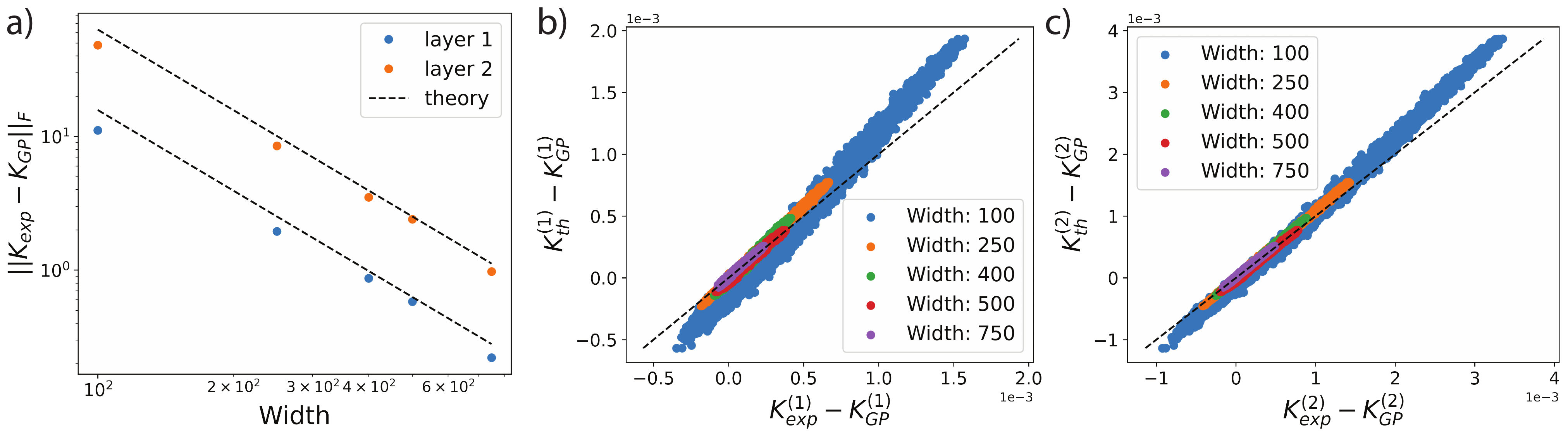}
    \caption{Learned representations in two-hidden-layer linear fully-connected neural networks with varying widths trained via Langevin sampling on 5000 MNIST images (see Appendix \ref{app:sec:numeric} for more details). \textbf{a)} The Frobenius norm of the deviation of the empirical average kernel of each layer from its GP value (in this case, simply $G_{xx}$) for varying widths. We see perfect match with theoretical predictions, which are shown as dashed lines. We obtain the predicted $1/n$ decay with increasing width and the linear scaling with the depth where the deviations for first and second layers differ by a factor of $2$. \textbf{b)-c)} Scatter plot of individual elements of the experimental (ordinate) and theoretical (abscissa) kernels for both layers. For low widths a slight deviation is visible between experiment and theory, while for larger widths the agreement is better. }
    \label{fig:linear_scaling}
\end{figure}

\subsection{Deep linear convolutional networks}\label{sec:conv}

To demonstrate the applicability of Conjecture \ref{conj1} to non-fully-connected BNNs, we consider deep convolutional linear networks with no bias terms. Here, the appropriate notion of width is the number of channels in each hidden layer \cite{novak2019bayesian}. Following the setup of \citet{novak2019bayesian} and \citet{xiao2018dynamical}, we consider a network consisting of $d-1$ linear convolutional layers followed by a fully-connected linear readout layer. For simplicity, we restrict our attention to convolutions with periodic boundary conditions, and do not include internal pooling layers (see Appendix \ref{app:sec:cov} for more details). Concretely, we consider a network with hidden layer activations $h_{i,\mathfrak{a}}^{(\ell)}$, where $i$ indexes the $n_{\ell}$ channels of the layer and $\mathfrak{a}$ is a spatial multi-index. The hidden layer activations are then defined through the recurrence 
\begin{align}
    h_{i,\mathfrak{a}}^{(\ell)}(x) = \frac{1}{\sqrt{n_{\ell-1}}} \sum_{j=1}^{n_{\ell-1}} \sum_{\mathfrak{b}} w_{ij,\mathfrak{b}}^{(\ell)} h_{j,\mathfrak{a}+\mathfrak{b}}^{(\ell-1)}(x)
\end{align}
with base case $h_{i,\mathfrak{a}}^{(0)}(x) = x_{i,\mathfrak{a}}$, where $i$ indexes the input channels (e.g., image color channels). The feature map is then formed by flattening the output of the last hidden layer into an $n_{d-1}s$-dimensional vector, where $s$ is the total dimensionality of the inputs (see Appendix \ref{app:sec:cov} for details). We fix the prior distribution of the filter elements to be $w_{ij,\mathfrak{a}}^{(\ell)} \underset{\textrm{i.i.d.}}{\sim} \mathcal{N}(0, \sigma_{\ell}^2 v_{\mathfrak{a}})$, where $v_{\mathfrak{a}} > 0$ is a weighting factor that sets the fraction of receptive field variance at location $\mathfrak{a}$ (and is thus subject to the constraint $\sum_{\mathfrak{a}} v_{\mathfrak{a}} = 1$). For inputs $[x_{\mu}]_{i,\mathfrak{a}}$ and $[x_{\nu}]_{i,\mathfrak{a}}$, we introduce the four-index hidden layer kernels 
\begin{align}
    K^{(\ell)}_{\mu\nu, \mathfrak{a}\mathfrak{b}} \equiv \frac{1}{n_{\ell}} \sum_{i=1}^{n_{\ell}} h_{i,\mathfrak{a}}^{(\ell)}(x_{\mu}) h_{i,\mathfrak{b}}^{(\ell)}(x_{\nu}).
\end{align}
With the given readout strategy, the two-index feature map kernel appearing in Conjecture \ref{conj1} is related to the four-index kernel of the last hidden layer by $K^{(d-1)}_{\mu\nu} = \frac{1}{s} \sum_{\mathfrak{a}} K^{(d-1)}_{\mu\nu,\mathfrak{a}\mathfrak{a}}$. We discuss other readout strategies in Appendix \ref{app:sec:cov}, but use this vectorization strategy in our numerical experiments.

As shown by \citet{xiao2018dynamical}, the infinite-width four-index kernel obeys the recurrence
\begin{align}\label{eq:gp_kernels_cnn}
    [K^{(\ell)}_{\infty}]_{\mu\nu,\mathfrak{a}\mathfrak{b}} = \sigma_{\ell}^{2} \sum_{\mathfrak{c}} v_{\mathfrak{c}} [K_{\infty}^{(\ell-1)}]_{\mu\nu,(\mathfrak{a}+\mathfrak{c})(\mathfrak{b}+\mathfrak{c})} 
\end{align}
with base case $[K^{0}_{\infty}]_{\mu\nu,\mathfrak{a}\mathfrak{b}} = [G_{xx}]_{\mu\nu,\mathfrak{a}\mathfrak{b}} \equiv \frac{1}{n_0} \sum_{i=1}^{n_0} [x_{\mu}]_{i,\mathfrak{a}} [x_{\nu}]_{i,\mathfrak{b}}$. This gives convolutional linear networks a sense of spatial hierarchy that is not present in the fully-connected case: even at infinite width, the kernels include iterative spatial averaging.  

\begin{figure}
    \centering
    \includegraphics[width=0.97\linewidth]{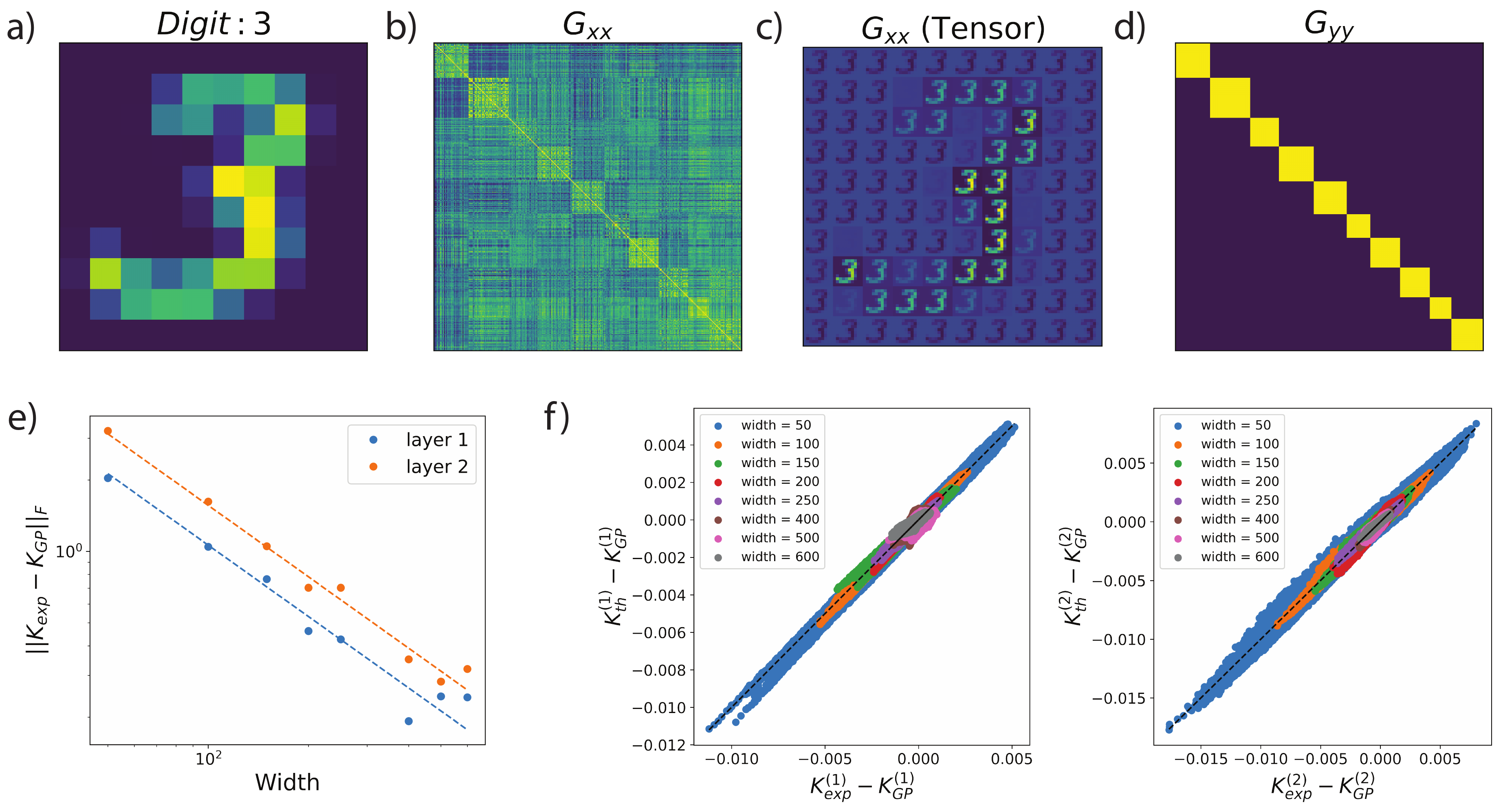}
     \caption{The MNIST image dataset and experiments for neural networks with two 1D convolutional layers. \textbf{(a)} A $10\times 10$ MNIST image downsized from $28\times 28$ pixels. \textbf{(b)} Input Gram matrix for $300$ MNIST images. \textbf{(c)} A single $(\mu,\nu)$ component of the input tensor $[G_{xx}]_{\mu\nu,\mathfrak{a}\mathfrak{b}}$ obtained using Eq. \eqref{eq:gp_kernels_cnn}. \textbf{(d)} The output Gram matrix. \textbf{(e)} The Frobenius norm of the correction to the 1D convolutional GP kernel is inversely proportional to the width. Here, the dashed lines are the theoretical predictions. \textbf{(f)} Scatter plots of individual elements of the empirical corrections to the GP kernels against the theoretical predictions for both layers show excellent agreement.}
    \label{fig:mnist_and_cnn1d}
\end{figure}

In Appendix C, we derive the kernel covariances appearing in Conjecture \ref{conj1}. As in the fully-connected case, this computation is easy to perform with the aid of Isserlis' theorem. The general result is somewhat complicated, but things simplify under the assumption that readout is performed using vectorization. Then, one finds that
\begin{equation}
    \langle K^{(\ell)}_{\mu\nu,\mathfrak{a}\mathfrak{b}} \rangle = [K^{(\ell)}_{\infty}]_{\mu\nu,\mathfrak{a}\mathfrak{b}} + \left(\prod_{\ell'=\ell}^{d-1} \sigma_{\ell}^{2}\right) \left(\sum_{\ell'=1}^{\ell} \frac{n_{d}}{n_{\ell'}}\right) \frac{1}{s}\sum_{\mathfrak{c}=1}^s \sum_{\rho,\lambda=1}^{p}[K_{\infty}^{(\ell)}]_{\mu\rho,\mathfrak{a}\mathfrak{c}} \Phi_{\rho\lambda} [K_{\infty}^{(\ell)}]_{\lambda\nu,\mathfrak{c}\mathfrak{b}} + \mathcal{O}(n^{-2}),
\end{equation}
where we have defined $\Phi_{\rho\lambda} \equiv [\sigma_{d}^{-2} \Gamma^{-1} G_{yy} \Gamma^{-1} - \Gamma^{-1}]_{\rho\lambda}$ for brevity. Thus, the correction to the convolutional kernel is quite similar to that obtained in the fully-connected case. To this order, the difference between these network architectures manifests itself largely through the difference in the infinite-width kernels. In Appendix C, we show that a similar simplification holds if readout is performed using global average pooling over space. 

As we did for fully-connected networks, we test whether our theory accurately predicts the results of numerical experiment, using the MNIST digit images illustrated in \ref{fig:mnist_and_cnn1d}(a-d). We consider a network with one-dimensional (Figure \ref{fig:mnist_and_cnn1d}e and f) and two-dimensional (Figure \ref{fig:cnn_scaling_kernels}) convolutional hidden layers, trained to classify $50$ MNIST images (see Appendix \ref{app:sec:numeric} for details of our numerical methods). As shown in Figure \ref{fig:mnist_and_cnn1d}(e, f) (Figure \ref{fig:cnn_scaling_kernels}(a,b) for 2D convolutions), we again obtain good quantitative agreement between the predictions of our asymptotic theory and the results of numerical experiment. In Figure \ref{fig:cnn_scaling_kernels}c, we directly visualize the learned feature kernels for 2D convolutional layers, illustrating the good agreement between theory and experiment. Therefore, our asymptotic theory can be applied to accurately predict learned representations in deep convolutional linear networks. 

\begin{figure}
    \centering
    \includegraphics[width=0.97\linewidth]{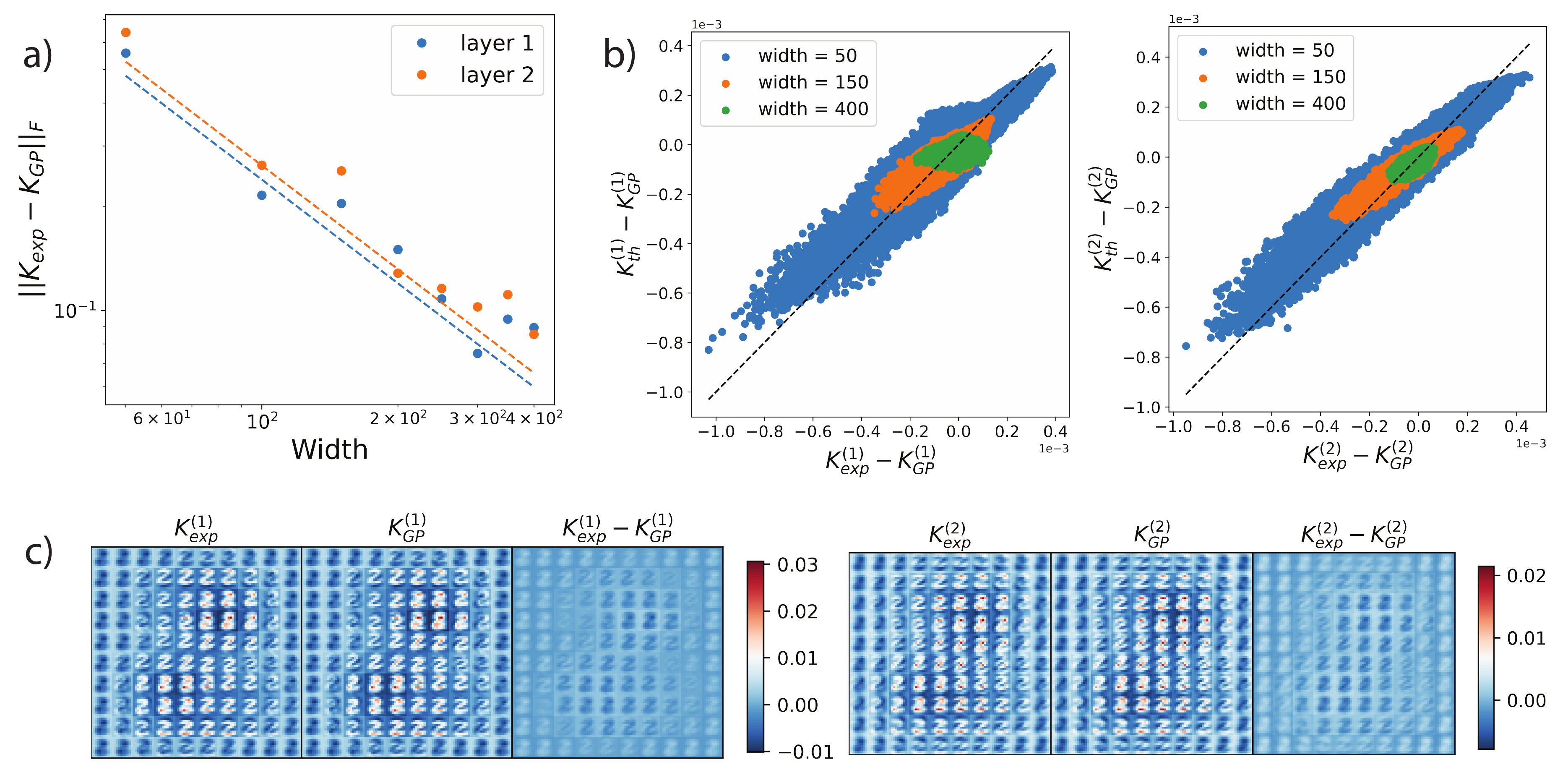}
    \caption{Learned representations in two-hidden-layer linear 2-D convolutional networks of varying channel widths. \textbf{(a)} The Frobenius norm of the correction to the GP kernel is inversely proportional to the width. Here, the dashed lines represent theory predictions. \textbf{(b)} Scatter plots of individual elements of the empirical corrections to the GP kernels against the theoretical predictions for both layers show good agreement. \textbf{(c)} A single component $(\mu,\nu)$ of the learned feature kernels in 2-layer CNN experiments for both convolutional layers. While the experimental kernel looks quite similar the GP (first and second columns), their difference shows the finite width corrections to the GP (last column).}
    \label{fig:cnn_scaling_kernels}
\end{figure}

\subsection{Networks with a single nonlinear hidden layer}\label{sec:nonlinear}

Finally, we would like to gain some understanding of how including nonlinearity affects the structure of learned representations. However, for a nonlinear MLP, it is usually not possible to analytically compute $\cov_{\mathcal{W}}(K^{(\ell)}_{\mu\nu},K^{(d-1)}_{\rho\lambda})$ to the required order \cite{yang2020feature,yaida2020,dyer2020asymptotics,aitken2020asymptotics}. Here, we consider the case of a network with a single nonlinear layer and no bias terms, in which we can both summarize the key obstacles to studying deep nonlinear networks and gain some intuitions about how they might differ from linear BNNs. Concretely, we consider a network with feature map $\bm{\psi}(\mathbf{x};W^{(1)}) = \phi(n_{0}^{-1/2} W^{(1)} \mathbf{x})$ for an elementwise activation function $\phi$, where the weight matrix $W^{(1)}$ has prior distribution $[W^{(1)}]_{ij} \sim_{\textrm{i.i.d.}} \mathcal{N}(0,\sigma_{1}^{2})$. The only hidden layer kernel of this network is the feature map postactivation kernel $K_{\mu\nu}$ defined in \eqref{eqn:featuremapkernel}, where we drop the layer index for brevity. As detailed in Appendix \ref{app:sec:nonlinkernel}, for such a network we have the exact expressions
\begin{align}
    [K_{\infty}]_{\mu\nu} = \mathbb{E}_{\mathcal{W}} K_{\mu\nu} &= \mathbb{E}[\phi(h_{\mu}) \phi(h_{\nu}) ],
    \\
    n_{1} \cov_{\mathcal{W}}(K_{\mu\nu},K_{\rho\lambda}) &= \mathbb{E}[ \phi(h_{\mu}) \phi(h_{\nu}) \phi(h_{\rho}) \phi(h_{\lambda}) ] - [K_{\infty}]_{\mu\nu} [K_{\infty}]_{\rho\lambda},
\end{align}
where expectations are taken over the $p$-dimensional Gaussian random vector $h_{\mu}$, which has mean zero and covariance $\cov(h_{\mu},h_{\nu}) = \sigma_{1}^{2} [G_{xx}]_{\mu\nu}$. Unlike for deeper nonlinear networks, here there are no finite-width corrections to the prior expectations \cite{neal1996priors,yaida2020,dyer2020asymptotics}. 

Though these expressions are easy to define, it is not possible to evaluate the four-point expectation in closed form for general Gram matrices $G_{xx}$ and activation functions $\phi$, including ReLU and erf. This obstacle has been noted in previous studies \cite{yang2020feature,yaida2020,naveh2020predicting}, and makes it challenging to extend approaches similar to those used here to deeper nonlinear networks. For polynomial activation functions, the required expectations can be evaluated using Isserlis' theorem (see Appendix \ref{app:sec:tech}). However, even for a quadratic activation function $\phi(x) = x^2$, the resulting formula for the kernel will involve many elementwise matrix products, and cannot be simplified into an intuitively comprehensible form. 

If the input Gram matrix $G_{xx}$ is diagonal, the four-point expectation becomes tractable because the required expectations factor across sample indices. In this simple case, there is an interesting distinction between the behavior of activation functions that yield $\mathbb{E}\phi(h) = 0$ and those that yield $\mathbb{E}\phi(h) \neq 0$. As detailed in Appendix \ref{app:sec:deeplinearkernel}, if $\mathbb{E}\phi(h) = 0$, $K_{\infty}$ is diagonal, and a given element of the leading finite-width correction to $\langle K \rangle$ depends only on the corresponding element of $G_{yy}$. However, if $\mathbb{E} \phi(h) \neq 0$, then $K_{\infty}$ includes a rank-1 component, and each element of the correction depends on all elements of $G_{yy}$. This means that the case in which $G_{xx}$ is diagonal is qualitatively distinct from the case in which there is only a single training input for such activation functions.

\section{Learned representations in deep nonlinear networks}

\begin{figure}
    \centering
    \includegraphics[height=4in]{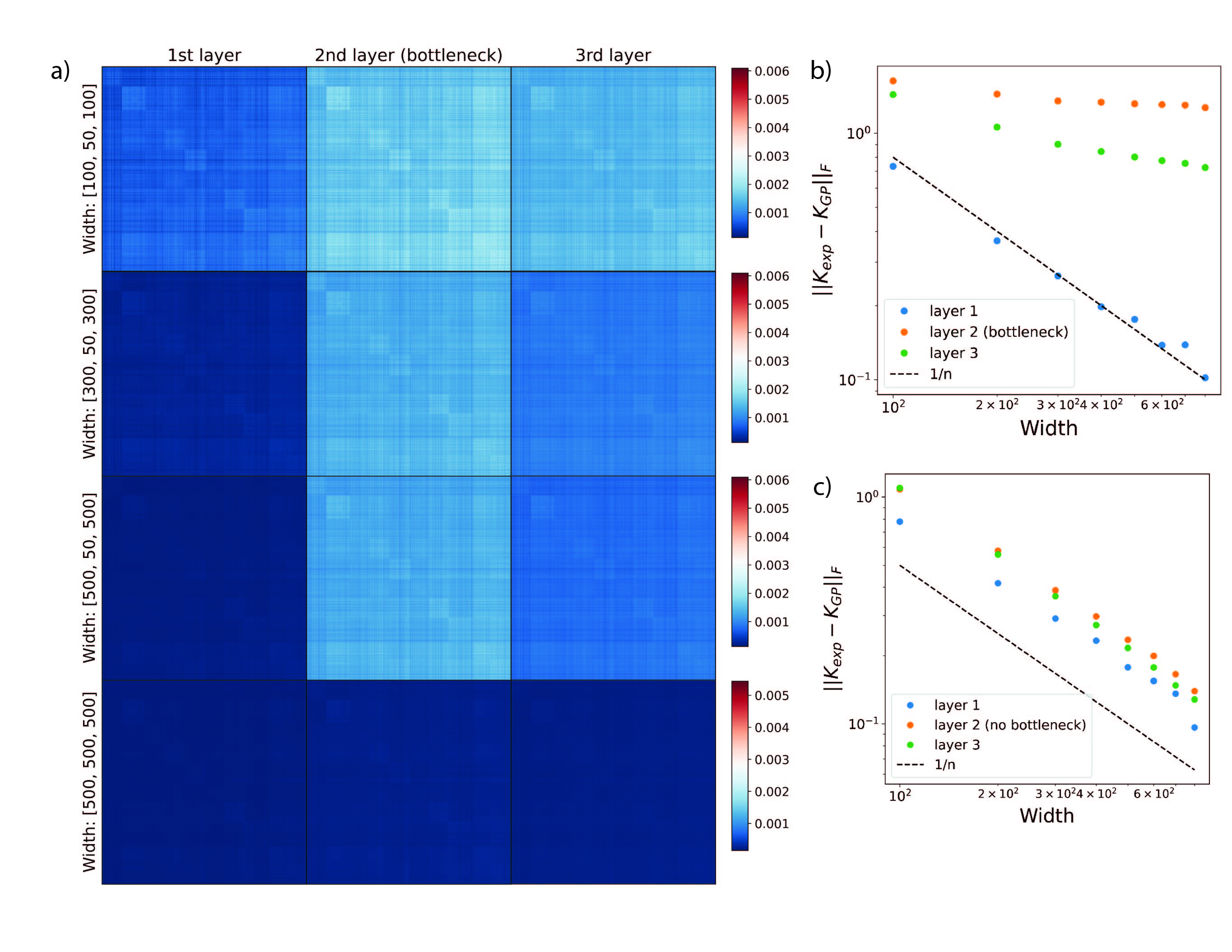}
    \setlength{\belowcaptionskip}{-8pt}
    \caption{3-hidden layer neural network with ReLU activations trained via Langevin sampling on 1000 MNIST images (see Appendix \ref{app:sec:numeric}). \textbf{(a)} The empirical average kernels subtracted from their corresponding GP kernels for all layers with varying widths. Labels on the y-axes indicate the widths of each layer. We observe that for networks with bottleneck layers, the deviation from $K^{(\ell)}_{\infty}$ is largest at the bottleneck indicating representation learning; without a bottleneck deviations are considerably less (the last row). \textbf{(b)} Hidden layer kernel deviation from GP kernels as a function of width for bottleneck networks. While the first layer shows $1/n$ scaling, the bottleneck layer and the 3\textsuperscript{rd} layer deviations stay almost constant. This behavior is predicted analytically for linear networks. \textbf{(c)} As in (b) for networks without a bottleneck. Consistent with our theory, all layers display $1/n$ decay.}
    \label{fig:relu_bottleneck}
\end{figure}

In the preceding section, we noted that analytical study of learned representations in deep nonlinear BNNs is generally quite challenging. Here, we use numerical experiments to explore whether any of the intuitions gained in the linear setting carry over to nonlinear networks. Concretely, we study how narrow bottlenecks affect representation learning in a more realistic nonlinear network. We train a network with three hidden layers and ReLU activations on a subset of the MNIST dataset \cite{lecun2010mnist}. Despite its analytical simplicity, ReLU is among the activation functions for which the covariance term in Conjecture \ref{conj1} cannot be evaluated in closed form (see \S\ref{sec:nonlinear}). However, it is straightforward to simulate numerically. Consistent with the predictions of our theory for linear networks, we find that introducing a narrow bottleneck leads to more representation learning in subsequent hidden layers, even if those layers are quite wide (Figure \ref{fig:relu_bottleneck}). Quantitatively, if one increases the width of the hidden layers between which the fixed-width bottleneck is sandwiched, the deviation of the first layer's kernel from its GP value decays roughly as $1/n$ with increasing width, while the deviations for the bottleneck and subsequent layers remain roughly constant. In contrast, the kernel deviations throughout a network with equal-width hidden layers decay roughly as $1/n$ (Figure \ref{fig:relu_bottleneck}). These observations are qualitatively consistent with the width-dependence of the linear network kernel \eqref{eqn:linearkernel}, as well as with previous studies of networks with infinitely-wide layers separated by a finite bottleneck \cite{agrawal2020bottleneck}. Keeping in mind the obstacles noted in \S\ref{sec:nonlinear}, precise characterization of nonlinear networks will be an interesting objective for future work. 

\section{Related work}

Our work is closely related to several recent analytical studies of finite-width BNNs. First, \citet{aitchison2020bigger} argued that the flexibility afforded by finite-width BNNs can be advantageous. He derived a recurrence relation for the learned feature kernels in deep linear networks, which he solved in the limits of infinite width and few outputs, narrow width and many outputs, and infinite width and many outputs. As discussed in \S\ref{sec:linear} and in Appendix \ref{app:sec:comp}, our results on deep linear networks extend those of his work. Furthermore, our numerical results support his suggestion that networks with narrow bottlenecks may learn interesting features. 

Moreover, our analytical approach and the asymptotic regime we consider mirror recent perturbative studies of finite-width BNNs. As noted in \S\ref{sec:universal} and Appendix \ref{app:sec:perturbation}, we make use of the results of \citet{yaida2020}, who derived recurrence relations for the perturbative corrections to the cumulants of the finite-width prior for an MLP. However, Yaida did not attempt to study the statistics of learned features; the goal of his work was to establish a general framework for the study of finite-width corrections. Bounds on the prior cumulants of a broader class of observables have been studied by Gur-Ari and colleagues \cite{dyer2020asymptotics,aitken2020asymptotics,andreassen2020asymptotics}; these results could allow for the identification of observables to which Conjecture \ref{conj1} should apply. Finally, perturbative corrections to the network prior and posterior have been studied by \citet{halverson2021neural} and \citet{naveh2020predicting}, respectively. Our work builds upon these studies by perturbatively characterizing the internal representations that are learned upon inference. 

Following the appearance of our work in preprint form, \citet{roberts2021principles} announced an alternative derivation of the zero-temperature limit of Conjecture \ref{conj1} for MLPs; we have adopted their terminology of hidden layer observables. As in \citet{yaida2020}'s earlier work, they rely on sequential perturbative approximation of the prior over preactivations as the hidden layers are marginalized out in order from the first to the last. While our elementary perturbative argument for Conjecture \ref{conj1} does not require assuming a particular network architecture for the hidden layers, it takes as input information regarding the prior cumulants that would have to be approximated using such methods. Moreover, the approach of layer-by-layer approximation to the prior could enable a fully rigorous version of Conjecture \ref{conj1} to be proved on an architecture-by-architecture basis \cite{hanin2021random}. 

Our work, like most studies of wide BNNs \cite{neal1996priors,williams1997computing,lee2018deep,matthews2018gaussian,novak2019bayesian,yang2019scaling,yang2020feature,aitchison2020bigger,wilson2020bayesian,yaida2020,halverson2021neural,antognini2019finite,naveh2020predicting,zv2021exact,dyer2020asymptotics,aitken2020asymptotics,jacot2018neural}, focuses on the regime in which the sample size $p$ is held fixed while the hidden layer width scale $n$ tends to infinity, i.e., $p \ll n$. One can instead consider regimes in which $p$ is not negligible relative to $n$, in which the posterior would be expected to concentrate. The behavior of deep linear BNNs in this regime was recently studied by \citet{li2021statistical}, who computed asymptotic approximations for the predictor statistics and hidden layer kernels. In Appendix \ref{app:sec:comp}, we show that our result \eqref{eqn:lowtemplinear} for the zero-temperature kernel can be recovered as the $p/n\downarrow 0$ limit of their result. As the dataset size $p$ appears only implicitly in our approach, we leave the incorporation of large-$p$ corrections as an interesting objective for future work. We note, however, that alternative methods developed to study the large-$p$ regime \cite{li2021statistical,naveh2021selfconsistent} cannot overcome the obstacles to analytical study of deep nonlinear networks encountered here.

\section{Conclusions}

In this paper, we have shown that the leading perturbative feature learning corrections to the infinite-width kernels of wide BNNs with linear readout and least-squares cost should be of a tightly constrained form. We demonstrate analytically and with numerical experiments that these results hold for certain tractable network architectures, and conjecture that they should extend to more general network architectures that admit a well-defined GP limit. 

\textit{Limitations.} We emphasize that our perturbative argument for Conjecture \ref{conj1} is not rigorous, and that we have not obtained quantitative bounds on the remainder for general network architectures. It is possible that there are non-perturbative contributions to the posterior statistics that are not captured by Conjecture \ref{conj1}; non-perturbative investigation of feature learning in finite BNNs will be an interesting objective for future work \cite{zv2021exact,noci2021precise}. More broadly, we leave rigorous proofs of the applicability of our results to more general architectures and of the smallness of the remainder as objective for future work. As mentioned above, one could attempt such a proof on an architecture-by-architecture basis \cite{yaida2020,roberts2021principles,hanin2021random}. Alternatively, one could attempt to treat all sufficiently sensible architectures uniformly \cite{yang2019scaling,yang2020feature}. Furthermore, we have considered only one possible asymptotic regime: that in which the width is taken to infinity with a finite training dataset and small output dimensionality. As discussed above in reference to the work of \citet{aitchison2020bigger} and \citet{li2021statistical}, investigation of alternative limits in which output dimension, dataset size, depth, and hidden layer width are all taken to infinity with fixed ratios may be an interesting subject for future work.

\begin{ack}
We thank B. Bordelon for helpful comments on our manuscript. JAZ-V acknowledges partial support from the NSF-Simons Center for Mathematical and Statistical Analysis of Biology at Harvard and the Harvard Quantitative Biology Initiative. This work was further supported by the Harvard Data Science Initiative Competitive Research Fund, the Harvard Dean’s Competitive Fund for Promising Scholarship, and a Google Faculty Research Award. The authors declare no conflict of interest. 
\end{ack}

\bibliography{references}

\begin{thebibliography}{39}
\providecommand{\natexlab}[1]{#1}
\providecommand{\url}[1]{\texttt{#1}}
\expandafter\ifx\csname urlstyle\endcsname\relax
  \providecommand{\doi}[1]{doi: #1}\else
  \providecommand{\doi}{doi: \begingroup \urlstyle{rm}\Url}\fi

\bibitem[Goodfellow et~al.(2016)Goodfellow, Bengio, Courville, and
  Bengio]{goodfellow2016deep}
Ian Goodfellow, Yoshua Bengio, Aaron Courville, and Yoshua Bengio.
\newblock \emph{Deep learning}.
\newblock MIT Press, Cambridge, MA, USA, 2016.

\bibitem[LeCun et~al.(2015)LeCun, Bengio, and Hinton]{lecun2015deep}
Yann LeCun, Yoshua Bengio, and Geoffrey Hinton.
\newblock Deep learning.
\newblock \emph{Nature}, 521\penalty0 (7553):\penalty0 436--444, 2015.

\bibitem[Neal(1996)]{neal1996priors}
Radford~M Neal.
\newblock Priors for infinite networks.
\newblock In \emph{Bayesian Learning for Neural Networks}, pages 29--53.
  Springer, 1996.

\bibitem[Williams(1997)]{williams1997computing}
Christopher~KI Williams.
\newblock Computing with infinite networks.
\newblock \emph{Advances in Neural Information Processing Systems}, pages
  295--301, 1997.

\bibitem[Lee et~al.(2018)Lee, Sohl-Dickstein, Pennington, Novak, Schoenholz,
  and Bahri]{lee2018deep}
Jaehoon Lee, Jascha Sohl-Dickstein, Jeffrey Pennington, Roman Novak, Sam
  Schoenholz, and Yasaman Bahri.
\newblock Deep neural networks as {Gaussian} processes.
\newblock In \emph{International Conference on Learning Representations}, 2018.
\newblock URL \url{https://openreview.net/forum?id=B1EA-M-0Z}.

\bibitem[Matthews et~al.(2018)Matthews, Hron, Rowland, Turner, and
  Ghahramani]{matthews2018gaussian}
Alexander G. de~G. Matthews, Jiri Hron, Mark Rowland, Richard~E. Turner, and
  Zoubin Ghahramani.
\newblock Gaussian process behaviour in wide deep neural networks.
\newblock In \emph{International Conference on Learning Representations}, 2018.
\newblock URL \url{https://openreview.net/forum?id=H1-nGgWC-}.

\bibitem[Novak et~al.(2019)Novak, Xiao, Bahri, Lee, Yang, Abolafia, Pennington,
  and Sohl-Dickstein]{novak2019bayesian}
Roman Novak, Lechao Xiao, Yasaman Bahri, Jaehoon Lee, Greg Yang, Daniel~A.
  Abolafia, Jeffrey Pennington, and Jascha Sohl-Dickstein.
\newblock Bayesian deep convolutional networks with many channels are
  {G}aussian processes.
\newblock In \emph{International Conference on Learning Representations}, 2019.
\newblock URL \url{https://openreview.net/forum?id=B1g30j0qF7}.

\bibitem[Yang(2019)]{yang2019scaling}
Greg Yang.
\newblock Scaling limits of wide neural networks with weight sharing:
  {Gaussian} process behavior, gradient independence, and neural tangent kernel
  derivation.
\newblock \emph{arXiv preprint arXiv:1902.04760}, 2019.

\bibitem[Yang and Hu(2020)]{yang2020feature}
Greg Yang and Edward~J Hu.
\newblock Feature learning in infinite-width neural networks.
\newblock \emph{arXiv preprint arXiv:2011.14522}, 2020.

\bibitem[Aitchison(2020)]{aitchison2020bigger}
Laurence Aitchison.
\newblock Why bigger is not always better: on finite and infinite neural
  networks.
\newblock In Hal~Daum\'e III and Aarti Singh, editors, \emph{Proceedings of the
  37th International Conference on Machine Learning}, volume 119 of
  \emph{Proceedings of Machine Learning Research}, pages 156--164. PMLR, July
  2020.
\newblock URL \url{http://proceedings.mlr.press/v119/aitchison20a.html}.

\bibitem[Wilson and Izmailov(2020)]{wilson2020bayesian}
Andrew~Gordon Wilson and Pavel Izmailov.
\newblock Bayesian deep learning and a probabilistic perspective of
  generalization.
\newblock \emph{arXiv preprint arXiv:2002.08791}, 2020.

\bibitem[Yaida(2020)]{yaida2020}
Sho Yaida.
\newblock Non-{G}aussian processes and neural networks at finite widths.
\newblock In Jianfeng Lu and Rachel Ward, editors, \emph{Proceedings of The
  First Mathematical and Scientific Machine Learning Conference}, volume 107 of
  \emph{Proceedings of Machine Learning Research}, pages 165--192, Princeton
  University, Princeton, NJ, USA, July 2020. PMLR.
\newblock URL \url{http://proceedings.mlr.press/v107/yaida20a.html}.

\bibitem[Halverson et~al.(2021)Halverson, Maiti, and
  Stoner]{halverson2021neural}
James Halverson, Anindita Maiti, and Keegan Stoner.
\newblock Neural networks and quantum field theory.
\newblock \emph{Machine Learning: Science and Technology}, 2021.

\bibitem[Antognini(2019)]{antognini2019finite}
Joseph~M Antognini.
\newblock Finite size corrections for neural network {G}aussian processes.
\newblock \emph{arXiv preprint arXiv:1908.10030}, 2019.

\bibitem[Naveh et~al.(2020)Naveh, Ben-David, Sompolinsky, and
  Ringel]{naveh2020predicting}
Gadi Naveh, Oded Ben-David, Haim Sompolinsky, and Zohar Ringel.
\newblock Predicting the outputs of finite networks trained with noisy
  gradients.
\newblock \emph{arXiv preprint arXiv:2004.01190}, 2020.

\bibitem[Li and Sompolinsky(2021)]{li2021statistical}
Qianyi Li and Haim Sompolinsky.
\newblock Statistical mechanics of deep linear neural networks: The
  backpropagating kernel renormalization.
\newblock \emph{Phys. Rev. X}, 11:\penalty0 031059, Sep 2021.
\newblock \doi{10.1103/PhysRevX.11.031059}.
\newblock URL \url{https://link.aps.org/doi/10.1103/PhysRevX.11.031059}.

\bibitem[Zavatone-Veth and Pehlevan(2021)]{zv2021exact}
Jacob~A Zavatone-Veth and Cengiz Pehlevan.
\newblock Exact marginal prior distributions of finite {Bayesian} neural
  networks.
\newblock In Marc'Aurelio Ranzato, Alina Beygelzimer, Percy Liang, Jenn~Wortman
  Vaughan, and Yann Dauphin, editors, \emph{Advances in Neural Information
  Processing Systems}, volume~34. Curran Associates, Inc., 2021.
\newblock URL \url{https://arxiv.org/abs/2104.11734}.

\bibitem[Dyer and Gur-Ari(2020)]{dyer2020asymptotics}
Ethan Dyer and Guy Gur-Ari.
\newblock Asymptotics of wide networks from {F}eynman diagrams.
\newblock In \emph{International Conference on Learning Representations}, 2020.
\newblock URL \url{https://openreview.net/forum?id=S1gFvANKDS}.

\bibitem[Aitken and Gur-Ari(2020)]{aitken2020asymptotics}
Kyle Aitken and Guy Gur-Ari.
\newblock On the asymptotics of wide networks with polynomial activations.
\newblock \emph{arXiv preprint arXiv:2006.06687}, 2020.

\bibitem[Wenzel et~al.(2020)Wenzel, Roth, Veeling, Swiatkowski, Tran, Mandt,
  Snoek, Salimans, Jenatton, and Nowozin]{wenzel2020cold}
Florian Wenzel, Kevin Roth, Bastiaan Veeling, Jakub Swiatkowski, Linh Tran,
  Stephan Mandt, Jasper Snoek, Tim Salimans, Rodolphe Jenatton, and Sebastian
  Nowozin.
\newblock How good is the {B}ayes posterior in deep neural networks really?
\newblock In Hal~Daumé III and Aarti Singh, editors, \emph{Proceedings of the
  37th International Conference on Machine Learning}, volume 119 of
  \emph{Proceedings of Machine Learning Research}, pages 10248--10259. PMLR,
  13--18 Jul 2020.
\newblock URL \url{http://proceedings.mlr.press/v119/wenzel20a.html}.

\bibitem[Fortuin et~al.(2021)Fortuin, Garriga-Alonso, Wenzel, R{\"a}tsch,
  Turner, van~der Wilk, and Aitchison]{fortuin2021bayesian}
Vincent Fortuin, Adri{\`a} Garriga-Alonso, Florian Wenzel, Gunnar R{\"a}tsch,
  Richard Turner, Mark van~der Wilk, and Laurence Aitchison.
\newblock Bayesian neural network priors revisited.
\newblock \emph{arXiv preprint arXiv:2102.06571}, 2021.

\bibitem[Izmailov et~al.(2021)Izmailov, Vikram, Hoffman, and
  Wilson]{izmailov2021bayesian}
Pavel Izmailov, Sharad Vikram, Matthew~D Hoffman, and Andrew Gordon~Gordon
  Wilson.
\newblock What are {Bayesian} neural network posteriors really like?
\newblock In Marina Meila and Tong Zhang, editors, \emph{Proceedings of the
  38th International Conference on Machine Learning}, volume 139 of
  \emph{Proceedings of Machine Learning Research}, pages 4629--4640. PMLR,
  18--24 Jul 2021.
\newblock URL \url{http://proceedings.mlr.press/v139/izmailov21a.html}.

\bibitem[MacKay(1992)]{mackay1992practical}
David~JC MacKay.
\newblock A practical {Bayesian} framework for backpropagation networks.
\newblock \emph{Neural Computation}, 4\penalty0 (3):\penalty0 448--472, 1992.

\bibitem[Jacot et~al.(2018)Jacot, Gabriel, and Hongler]{jacot2018neural}
Arthur Jacot, Franck Gabriel, and Cl{\'e}ment Hongler.
\newblock Neural tangent kernel: Convergence and generalization in neural
  networks.
\newblock \emph{arXiv preprint arXiv:1806.07572}, 2018.

\bibitem[Hron et~al.(2020)Hron, Bahri, Novak, Pennington, and
  Sohl-Dickstein]{hron2020exact}
Jiri Hron, Yasaman Bahri, Roman Novak, Jeffrey Pennington, and Jascha
  Sohl-Dickstein.
\newblock Exact posterior distributions of wide {Bayesian} neural networks.
\newblock \emph{arXiv preprint arXiv:2006.10541}, 2020.

\bibitem[Andreassen and Dyer(2020)]{andreassen2020asymptotics}
Anders Andreassen and Ethan Dyer.
\newblock Asymptotics of wide convolutional neural networks.
\newblock \emph{arXiv preprint arXiv:2008.08675}, 2020.

\bibitem[Horn and Johnson(2012)]{horn2012matrix}
Roger~A Horn and Charles~R Johnson.
\newblock \emph{Matrix Analysis}.
\newblock Cambridge University Press, 2012.

\bibitem[Isserlis(1918)]{isserlis1918formula}
Leon Isserlis.
\newblock On a formula for the product-moment coefficient of any order of a
  normal frequency distribution in any number of variables.
\newblock \emph{Biometrika}, 12\penalty0 (1/2):\penalty0 134--139, 1918.

\bibitem[Wick(1950)]{wick1950evaluation}
Gian-Carlo Wick.
\newblock The evaluation of the collision matrix.
\newblock \emph{Physical Review}, 80\penalty0 (2):\penalty0 268, 1950.

\bibitem[Kloeden and Platen(1992)]{kloeden1992stochastic}
Peter~E Kloeden and Eckhard Platen.
\newblock Stochastic differential equations.
\newblock In \emph{Numerical Solution of Stochastic Differential Equations},
  pages 103--160. Springer, 1992.

\bibitem[Paszke et~al.(2019)Paszke, Gross, Massa, Lerer, Bradbury, Chanan,
  Killeen, Lin, Gimelshein, Antiga, Desmaison, Kopf, Yang, DeVito, Raison,
  Tejani, Chilamkurthy, Steiner, Fang, Bai, and Chintala]{paske2019pytorch}
Adam Paszke, Sam Gross, Francisco Massa, Adam Lerer, James Bradbury, Gregory
  Chanan, Trevor Killeen, Zeming Lin, Natalia Gimelshein, Luca Antiga, Alban
  Desmaison, Andreas Kopf, Edward Yang, Zachary DeVito, Martin Raison, Alykhan
  Tejani, Sasank Chilamkurthy, Benoit Steiner, Lu~Fang, Junjie Bai, and Soumith
  Chintala.
\newblock Pytorch: An imperative style, high-performance deep learning library.
\newblock In H.~Wallach, H.~Larochelle, A.~Beygelzimer, F.~d\textquotesingle
  Alch\'{e}-Buc, E.~Fox, and R.~Garnett, editors, \emph{Advances in Neural
  Information Processing Systems}, volume~32. Curran Associates, Inc., 2019.
\newblock URL
  \url{https://proceedings.neurips.cc/paper/2019/file/bdbca288fee7f92f2bfa9f7012727740-Paper.pdf}.

\bibitem[LeCun et~al.(2010)LeCun, Cortes, and Burges]{lecun2010mnist}
Yann LeCun, Corinna Cortes, and CJ~Burges.
\newblock {MNIST} handwritten digit database.
\newblock \emph{ATT Labs [Online]. Available:
  http://yann.lecun.com/exdb/mnist}, 2, 2010.

\bibitem[Novak et~al.(2020)Novak, Xiao, Hron, Lee, Alemi, Sohl-Dickstein, and
  Schoenholz]{neuraltangents2020}
Roman Novak, Lechao Xiao, Jiri Hron, Jaehoon Lee, Alexander~A. Alemi, Jascha
  Sohl-Dickstein, and Samuel~S. Schoenholz.
\newblock Neural tangents: Fast and easy infinite neural networks in python.
\newblock In \emph{International Conference on Learning Representations}, 2020.
\newblock URL \url{https://github.com/google/neural-tangents}.

\bibitem[Xiao et~al.(2018)Xiao, Bahri, Sohl-Dickstein, Schoenholz, and
  Pennington]{xiao2018dynamical}
Lechao Xiao, Yasaman Bahri, Jascha Sohl-Dickstein, Samuel Schoenholz, and
  Jeffrey Pennington.
\newblock Dynamical isometry and a mean field theory of {CNNs}: How to train
  10,000-layer vanilla convolutional neural networks.
\newblock In \emph{International Conference on Machine Learning}, pages
  5393--5402. PMLR, 2018.

\bibitem[Agrawal et~al.(2020)Agrawal, Papamarkou, and
  Hinkle]{agrawal2020bottleneck}
Devanshu Agrawal, Theodore Papamarkou, and Jacob Hinkle.
\newblock Wide neural networks with bottlenecks are deep {Gaussian} processes.
\newblock \emph{Journal of Machine Learning Research}, 21\penalty0
  (175):\penalty0 1--66, 2020.
\newblock URL \url{http://jmlr.org/papers/v21/20-017.html}.

\bibitem[Roberts et~al.(2021)Roberts, Yaida, and Hanin]{roberts2021principles}
Daniel~A Roberts, Sho Yaida, and Boris Hanin.
\newblock The principles of deep learning theory.
\newblock \emph{arXiv preprint arXiv:2106.10165}, 2021.

\bibitem[Hanin(2021)]{hanin2021random}
Boris Hanin.
\newblock Random neural networks in the infinite width limit as {Gaussian}
  processes.
\newblock \emph{arXiv preprint arXiv:2107.01562}, 2021.

\bibitem[Naveh and Ringel(2021)]{naveh2021selfconsistent}
Gadi Naveh and Zohar Ringel.
\newblock A self consistent theory of {G}aussian processes captures feature
  learning effects in finite {CNN}s.
\newblock \emph{arXiv preprint arXiv:2106.04110}, 2021.

\bibitem[Noci et~al.(2021)Noci, Bachmann, Roth, Nowozin, and
  Hofmann]{noci2021precise}
Lorenzo Noci, Gregor Bachmann, Kevin Roth, Sebastian Nowozin, and Thomas
  Hofmann.
\newblock Precise characterization of the prior predictive distribution of deep
  {ReLU} networks.
\newblock \emph{arXiv preprint arXiv:2106.06615}, 2021.

\end{thebibliography}

\newpage 

\appendix

\setcounter{page}{1}
\renewcommand*{\thepage}{S\arabic{page}}

\numberwithin{equation}{section}
\numberwithin{figure}{section}

\centerline{\Large\textbf{Supplemental Information}}

\startcontents[sections]
\printcontents[sections]{l}{1}{\setcounter{tocdepth}{1}}

\section{Preliminary technical results}\label{app:sec:tech}

In this appendix, we review useful technical results upon which our calculations rely.

\subsection{Isserlis' theorem for Gaussian moments}

Let $(x_{1},x_{2},\ldots,x_{n})$ be a zero-mean Gaussian random vector. Then, Isserlis' theorem \cite{isserlis1918formula} states that
\begin{align}
    \mathbb{E}[x_{1} x_{2} \cdots x_{n}] = \begin{cases} \sum_{p \in P_{n}^2} \prod_{(i,j) \in p} \cov(x_{i},x_{j}) & n\ \textrm{even} \\ 0 & n\ \textrm{odd}, \end{cases}
\end{align}
where the sum is over all pairings $p$ of $\{1,2,\ldots,n\}$ and the product is over all pairs contained in $p$. In particular, for $n=4$, we have
\begin{align}
    \mathbb{E}[x_{1} x_{2} x_{3} x_{4}] = \cov(x_{1},x_{2}) \cov(x_{3},x_{4}) + \cov(x_{1}, x_{3}) \cov(x_{2}, x_{4}) + \cov(x_{1},x_{4}) \cov(x_{2},x_{3}).
\end{align}
In physics, Isserlis' theorem is often known as Wick's probability theorem \cite{wick1950evaluation}. 

\subsection{Neumann series for matrix inverses near the identity}

The Neumann series is the generalization of the geometric series to bounded linear operators, including square matrices. In particular, let $A$ be a $p \times p$ square matrix. Then, we have
\begin{align}
    (I_{p} - A)^{-1} = \sum_{k=0}^{\infty} A^{k}
\end{align}
provided that the series converges in the operator norm \cite{horn2012matrix}. We will use this result without concern for rigorous convergence conditions, as we are interested only in asymptotic expansions.

\subsection{Series expansion of the log-determinant near the identity}

Let $A$ be a $p \times p$ square matrix, and let $t$ be a small parameter. Then, we have
\begin{align}
    \log \det(I_{p} + t A) = \sum_{k=1}^{\infty} \frac{(-1)^{k+1}}{k} \tr(A^{k}) t^{k}
\end{align}
assuming that the series converges. We will not concern ourselves with rigorous convergence conditions, as we will use this expansion formally.

This result follows from the fact that
\begin{align}
    \frac{\partial^{k}}{\partial t^{k}} \log\det(I_{p} + t A) = (-1)^{k+1} (k-1)! \tr( (I_{p}+t A)^{-k} A^{k}) \quad (k = 1,2,\ldots).
\end{align}
The base case $k=1$ is given by Jacobi's formula \cite{horn2012matrix}:
\begin{align}
    \frac{\partial}{\partial t} \log\det(I_{p} + t A) = \tr( (I_{p}+t A)^{-1} A).
\end{align}
Then, using the identity
\begin{align}
    \frac{\partial}{\partial t} (I_{p}+t A)^{-1} = - (I_{p}+t A)^{-1} A (I_{p}+t A)^{-1}
\end{align}
and the fact that $A$ commutes with $(I_{p} + t A)^{-1}$, we find that the claim holds by induction. As $\log\det(I_{p} + t A) |_{t=0} = 0$, this implies the desired Maclaurin series.

\section{Perturbation theory for wide Bayesian neural networks with linear readout}\label{app:sec:perturbation}

In this appendix, we derive Conjecture \ref{conj1}. As outlined in the main text, we consider a depth-$d$ neural network $\mathbf{f}: \mathbb{R}^{n_0} \to \mathbb{R}^{n_d}$ with linear readout, written as
\begin{align}
    \mathbf{f}(\mathbf{x};W^{d},\mathcal{W}) = \frac{1}{\sqrt{n_{d-1}}} W^{(d)} \bm{\psi}(\mathbf{x};\mathcal{W})
\end{align}
in terms of the hidden layer feature map $\bm{\psi}(\cdot ; \mathcal{W}) : \mathbb{R}^{n_0} \to \mathbb{R}^{n_{d-1}}$. The full set of trainable parameters is then $\Theta = \{W^{(d)},\mathcal{W}\}$, where $\mathcal{W}$ is the set of feature map parameters. We assume isotropic Gaussian priors over these parameters, with, for instance, 
\begin{align} 
    W^{(d)}_{ij} \underset{\textrm{i.i.d.}}{\sim} \mathcal{N}(0, \sigma_{d}^2).
\end{align}
We fix an arbitrary training dataset $\mathcal{D} = \{(\mathbf{x}_{\mu},\mathbf{y}_{\mu})\}_{\mu=1}^{p}$ of $p$ examples, and use a Gaussian likelihood $p(\mathcal{D} \,|\,\Theta) \propto \exp(-\beta E)$, where 
\begin{align}
    E(\Theta; \mathcal{D}) = \frac{1}{2} \sum_{\mu=1}^{p} \Vert \mathbf{f}(\mathbf{x}_{\mu}) - \mathbf{y}_{\mu} \Vert^2
\end{align}
is a quadratic cost. We then introduce the Bayes posterior
\begin{align} 
    p(\Theta\,|\,\mathcal{D}) = \frac{p(\mathcal{D}\,|\,\Theta) p(\Theta)}{p(\mathcal{D})};
\end{align}
averages with respect to this distribution will be denoted by $\langle \cdot \rangle$.

We define the postactivation feature map kernel
\begin{align}
    K^{(d-1)}(\mathbf{x},\mathbf{x}') \equiv n_{d-1}^{-1} \bm{\psi}(\mathbf{x},\mathcal{W}) \cdot \bm{\psi}(\mathbf{x}',\mathcal{W}),
\end{align}
and write $[K^{(d-1)}]_{\mu\nu} \equiv K^{(d-1)}(\mathbf{x}_{\mu},\mathbf{x}_{\nu})$ for the kernel evaluated on the training set. For brevity, we will frequently abbreviate $K \equiv K^{(d-1)}$ throughout this appendix.

We denote expectation by $\mathbb{E}$, and prior expectation by $\mathbb{E}_{\mathcal{W}}$. We also introduce the joint cumulant operator $\mathbb{K}$ and its prior counterpart $\mathbb{K}_{\mathcal{W}}$. We will only require the second and third joint cumulants, which, for random variables $A$, $B$, and $C$, are given as
\begin{align}
    \mathbb{K}(A,B) = \mathbb{E}[(A-\mathbb{E} A) (B - \mathbb{E} B)]
\end{align}
and
\begin{align}
    \mathbb{K}(A,B,C) = \mathbb{E}[(A-\mathbb{E} A) (B - \mathbb{E} B) (C - \mathbb{E} C)],
\end{align}
respectively. 

Our starting point is the partition function $Z$ of the Bayes posterior \eqref{eqn:posterior} for the network \eqref{eqn:general_network}, including a source term for the (generically matrix-valued) observable $O$:
\begin{align}
    Z(J) = \mathbb{E}_{W^{(d)}} \mathbb{E}_{\mathcal{W}} \, \exp\left(-\beta E + \tr(J^{\top} O) \right),
\end{align}
where $\mathcal{W}$ denotes all of the parameters except for the readout weight matrix $W^{(d)}$ and expectation is taken with respect to the Gaussian prior. The logarithm of the partition function is the posterior cumulant generating function of the observable $O$, with 
\begin{align}
    \langle O \rangle = \frac{\delta \log Z}{\delta J} \bigg|_{J = 0}
\end{align}
and covariance
\begin{align}
    \cov(O_{\rho \gamma},O_{\omega \chi}) = \frac{\partial^2 \log Z}{\partial J_{\rho\gamma} \partial J_{\omega \chi}} \bigg\vert_{J=0}.
\end{align}

\subsection{Integrating out the readout layer}

We first show that the readout layer can be integrated out exactly. As the source term is independent of $W^{(d)}$, Fubini's theorem yields
\begin{align}
    Z = \mathbb{E}_{\mathcal{W}}\bigg[ \exp(\tr(J^{\top} O))\ \mathbb{E}_{W^{(d)}} \exp(-\beta E) \bigg].
\end{align}
The expectation over $W^{d}$ is a Gaussian integral, hence it is easy to evaluate exactly:
\begin{align}
    &\mathbb{E}_{W^{(d)}} \exp(-\beta E) 
    \\
    &= \mathbb{E}_{W^{(d)}} \exp\left(-\frac{1}{2} \beta \sum_{\mu=1}^{p} \left\Vert \frac{1}{\sqrt{n_{d-1}}} W^{(d)} \bm{\psi}_{\mu} - \mathbf{y}_{\mu} \right\Vert^{2} \right)
    \\
    &= \exp\left(- \frac{1}{2} \beta \tr(Y^{\top} Y) \right) \nonumber\\&\quad \times \prod_{j=1}^{n_d} \left[ \int \frac{d\mathbf{w}_{j}}{(2\pi \sigma_{d}^{2})^{n_{d-1}/2}} \exp\left(-\frac{1}{2} \mathbf{w}_{j}^{\top} \left(\sigma_{d}^{-2} I_{n} + \frac{ \beta}{n_{d-1}} \Psi^{\top}\Psi\right) \mathbf{w}_{j} + \frac{\beta}{\sqrt{n_{d-1}}} (Y^{\top}\Psi)_{j \cdot} \mathbf{w}_{j} \right) \right]
    \\
    &= \det\left(I_{n} + \frac{\beta \sigma_{d}^{2}}{n_{d-1}} \Psi^{\top} \Psi\right)^{-n_{d}/2} \nonumber\\&\quad \times \exp\left(\frac{1}{2} \frac{\beta^{2} \sigma_{d}^{2}}{n_{d-1}} \tr\left[Y^{\top} \Psi \left(I_{n} + \frac{\beta \sigma_{d}^{2}}{n_{d-1}} \Psi^{\top}\Psi\right)^{-1} \Psi^{\top} Y\right] - \frac{1}{2} \beta \tr(Y^{\top} Y)\right),
\end{align}
where we abbreviate $\bm{\psi}_{\mu} \equiv \bm{\psi}(\mathbf{x}_{\mu};\mathcal{W})$ and introduce the matrices $\Psi_{\mu j} \equiv \psi_{\mu,j}$ and $Y_{\mu j} \equiv y_{\mu,j}$. Here, we have used the fact that the matrix $I_{n} + (\beta \sigma_{d}^{2}/n_{d-1}) \Psi^{\top} \Psi$ is invertible at any finite temperature. By the Weinstein–Aronszajn identity \cite{horn2012matrix},
\begin{align}
    \det\left(I_{n} + \frac{\beta \sigma_{d}^{2}}{n_{d-1}} \Psi^{\top} \Psi\right)
    = \det\left(I_{p} + \frac{\beta \sigma_{d}^{2}}{n_{d-1}} \Psi\Psi^{\top} \right)
    = \det( I_{p} + \beta \sigma_{d}^{2} K),
\end{align}
where we introduce the (non-constant) kernel matrix
\begin{align}
    K = K^{(d-1)} \equiv \frac{1}{n_{d-1}} \Psi\Psi^{\top};
\end{align}
as mentioned above, we abbreviate $K \equiv K^{(d-1)}$ for brevity. By the push-through identity \cite{horn2012matrix},
\begin{align}
    \frac{1}{n_{d-1}} \Psi \left(I_{n} + \frac{\beta \sigma_{d}^{2}}{n_{d-1}} \Psi^{\top}\Psi\right)^{-1} \Psi^{\top} 
    = \left(I_{p} + \frac{\beta \sigma_{d}^{2}}{n_{d-1}} \Psi \Psi^{\top}\right)^{-1} \frac{1}{n_{d-1}} \Psi \Psi^{\top}
    = (I_{p} + \beta \sigma_{d}^{2} K)^{-1} K,
\end{align}
hence, using the cyclic property of the trace,
\begin{align}
    & \frac{1}{2} \frac{\beta^{2} \sigma_{d}^{2}}{n_{d-1}} \tr\left[Y^{\top} \Psi \left(I_{n} + \frac{\beta \sigma_{d}^{2}}{n_{d-1}} \Psi^{\top}\Psi\right)^{-1} \Psi^{\top} Y\right] - \frac{1}{2} \beta \tr(Y^{\top} Y)
    \nonumber\\&\quad = \frac{1}{2} \beta n_{d} \tr\left[ \left(\beta \sigma_{d}^{2} (I_{p} + \beta \sigma_{d}^{2} K)^{-1} K - I_{p} \right) G_{yy} \right]
    \\&\quad = - \frac{1}{2} \beta n_{d} \tr[(I_{p} + \beta \sigma_{d}^{2} K)^{-1} G_{yy}],
\end{align}
where we have defined the normalized Gram matrix of the outputs
\begin{align}
    G_{yy} \equiv \frac{1}{n_d} Y Y^{\top} 
\end{align}
and noticed that
\begin{align}
    I_{p} - \beta \sigma_{d}^{2} (I_{p} + \beta \sigma_{d}^{2} K)^{-1} K = (I_{p} + \beta \sigma_{d}^{2} K)^{-1} .
\end{align}
Therefore, we conclude that
\begin{align}
    Z = \mathbb{E}_{\mathcal{W}} \exp\left[\tr(J^{\top} O) - \frac{n_{d}}{2} \left(\beta \tr[(I_{p} + \beta \sigma_{d}^{2} K)^{-1} G_{yy}] + \log\det( I_{p} + \beta \sigma_{d}^{2} K)\right) \right]
\end{align}
at any width.

\subsection{Perturbative expansion}

We now consider how this expression behaves in the large-width limit. We assume that this limit is well-defined in the sense that the readout kernel $K$ tends in probability to the constant GP kernel $K_{\infty}$  \cite{lee2018deep,matthews2018gaussian,novak2019bayesian,yang2019scaling}, and that the observable $O$ similarly tends to a deterministic limit $O_{\infty}$. Then, we formally write $K$ and $O$ as their infinite-width limits plus corrections which are small at large hidden layer widths:
\begin{align}
    K &= K_{\infty} + \lambda\, \delta K,
    \\
    O &= O_{\infty} + \lambda\, \delta O,
\end{align}
where the parameter $\lambda$ is used to track powers of the small deviations. 

We first expand the term resulting from integrating out the readout layer into its infinite-width limit and a finite-width correction. We define the constant matrix 
\begin{align}
    \Gamma \equiv K_{\infty} + \frac{1}{\beta \sigma_{d}^{2}} I_{p},
\end{align}
which is invertible at any finite temperature. Then, by the Woodbury identity \cite{horn2012matrix}, we have,
\begin{align}
    \beta \sigma_{d}^{2} (I_{p} + \beta \sigma_{d}^{2} K)^{-1} 
    = (\Gamma + \lambda \delta K)^{-1}
    = \Gamma^{-1} - \lambda \Gamma^{-1} \delta K (\Gamma + \lambda \delta K)^{-1}
\end{align}
and, similarly,
\begin{align}
    \log \det(I_{p} + \beta \sigma_{d}^{2} K) = \log\det(\beta \sigma_{d}^{2} \Gamma) + \log\det(I_{p} + \lambda \Gamma^{-1} \delta K).
\end{align}
Noting that that both $\lambda \Gamma^{-1} \delta K (\Gamma + \lambda \delta K)^{-1}$ and $\log\det(I_{p} + \lambda \Gamma^{-1} \delta K)$ are $\mathcal{O}(\lambda)$, we expand the logarithm of the partition function as
\begin{align}
    \log Z = \log Z_{\infty} + \tr(J^{\top} O_{\infty}) + \log \mathbb{E}_{\mathcal{W}} \exp[\lambda \tr( J^{\top} \delta O ) + \lambda \Omega],
\end{align}
where
\begin{align}
    Z_{\infty} \equiv \det(\beta \sigma_{d}^{2} \Gamma)^{-n_{d}/2} \exp\left(-\frac{1}{2} n_{d} \sigma_{d}^{-2} \tr(\Gamma^{-1} G_{yy}) \right)
\end{align}
is the GP partition function and
\begin{align}
    \Omega \equiv \frac{1}{2} n_{d} \tr[ \sigma_{d}^{-2} \Gamma^{-1} \delta K (\Gamma + \lambda \delta K)^{-1} G_{yy}] - \frac{1}{2} n_{d} \lambda^{-1} \log\det(I_{p} + \lambda \Gamma^{-1} \delta K)
\end{align}
is the remainder. $\log \mathbb{E}_{\mathcal{W}} \exp[\lambda \tr( J^{\top} \delta O ) + \lambda \Omega]$ has the form of a cumulant generating function, hence it has a formal series expansion in $\lambda$ given by
\begin{align}
    \log \mathbb{E}_{\mathcal{W}} \exp[\lambda \tr( J^{\top} \delta O ) + \lambda \Omega] 
    &= \lambda \mathbb{E}_{\mathcal{W}} [\tr( J^{\top} \delta O ) + \Omega] 
    \nonumber\\&\quad + \frac{1}{2} \lambda^2\, \mathbb{E}_{\mathcal{W}}  \{\tr[ J^{\top} (\delta O - \mathbb{E}_{\mathcal{W}} \delta O)] + \Omega - \mathbb{E}_{\mathcal{W}} \Omega] \}^2 
    \nonumber\\&\quad + \frac{1}{6} \lambda^{3} \mathbb{E}_{\mathcal{W}} \{\tr[ J^{\top} (\delta O - \mathbb{E}_{\mathcal{W}} \delta O)] + \Omega - \mathbb{E}_{\mathcal{W}} \Omega] \}^3 
    \nonumber\\&\quad + \mathcal{O}(\lambda^4). 
\end{align}
We can then see that the $k$-th cumulant is $\mathcal{O}(J^k)$, hence the $k$-th posterior cumulant of $O$ will be $\mathcal{O}(\lambda^k)$. Specifically, we can read off the posterior mean
\begin{align}
    \langle O \rangle = O_{\infty} + \lambda \mathbb{E}_{\mathcal{W}} \delta O + \lambda^2 \mathbb{K}_{\mathcal{W}}(\delta O, \Omega) + \frac{1}{2} \lambda^3 \mathbb{K}_{\mathcal{W}}(\delta O, \Omega,\Omega) + \mathcal{O}(\lambda^4).
\end{align}
and covariance
\begin{align}
    \cov(O_{\rho \gamma},O_{\omega \chi}) = \lambda^2 \mathbb{K}_{\mathcal{W}}(\delta O_{\rho\gamma}, \delta O_{\omega \chi} ) + \lambda^3 \mathbb{K}_{\mathcal{W}}( \delta O_{\rho\gamma}, \delta O_{\omega \chi}, \Omega) + \mathcal{O}(\lambda^4).
\end{align}
To make further progress, we expand $\Omega$ in powers of $\lambda$. Using the Neumann series for the matrix inverse (see Appendix \ref{app:sec:tech}), we have
\begin{align}
    (\Gamma + \lambda \delta K)^{-1}
    = \Gamma^{-1} - \lambda \Gamma^{-1} \delta K \Gamma^{-1} + \mathcal{O}(\lambda^2),
\end{align}
and, using the series expansion of the log-determinant near the identity (see Appendix \ref{app:sec:tech}), we have
\begin{align}
    \lambda^{-1} \log\det(I_{p} + \lambda \Gamma^{-1} \delta K) = \tr(\Gamma^{-1} \delta K) - \frac{1}{2} \lambda \tr(\Gamma^{-1} \delta K \Gamma^{-1} \delta K) + \mathcal{O}(\lambda^2).
\end{align}
This yields
\begin{align}
    \Omega 
    &= \frac{n_{d}}{2} \tr[ (\sigma_{d}^{-2} \Gamma^{-1} G_{yy} \Gamma^{-1} - \Gamma^{-1}) \delta K ] 
    \nonumber\\&\quad - \frac{n_{d}}{2} \lambda \tr\left[ \left(\sigma_{d}^{-2} \Gamma^{-1} G_{yy} \Gamma^{-1}   - \frac{1}{2} \Gamma^{-1} \right) \delta K \Gamma^{-1} \delta K \right]
    \nonumber\\&\quad + \mathcal{O}(\lambda^2).
\end{align}
The leading term is simple because it is linear in $\delta K$. Then, keeping only the leading non-trivial corrections and recognizing that
\begin{align}
    O_{\infty} + \lambda \mathbb{E}_{\mathcal{W}} \delta O &= \mathbb{E}_{\mathcal{W}} O,
    \\
    \lambda^2 \mathbb{K}_{\mathcal{W}}(\delta O, \delta K_{\mu\nu} ) &= \mathbb{K}_{\mathcal{W}}(O, K_{\mu\nu}),
    \\
    \lambda^2 \mathbb{K}_{\mathcal{W}}(\delta O_{\rho\gamma}, \delta O_{\omega \chi} ) &= \mathbb{K}_{\mathcal{W}}(O_{\rho\gamma}, O_{\omega \chi} ), \quad \textrm{and}
    \\
    \lambda^3 \mathbb{K}_{\mathcal{W}}( \delta O_{\rho\gamma}, \delta O_{\omega \chi}, \delta K_{\mu\nu} ) &= \mathbb{K}_{\mathcal{W}}(O_{\rho\gamma}, O_{\omega \chi}, K_{\mu\nu} ),
\end{align}
we have
\begin{align}
    \langle O \rangle = \mathbb{E}_{\mathcal{W}} O + \frac{1}{2} n_{d} \sum_{\mu,\nu=1}^{p} (\sigma_{d}^{-2} \Gamma^{-1} G_{yy} \Gamma^{-1} - \Gamma^{-1})_{\mu\nu} \mathbb{K}_{\mathcal{W}}(O, K_{\mu\nu}) + \mathcal{O}(\lambda^3)
\end{align}
and
\begin{align}
    \cov(O_{\rho \gamma},O_{\omega \chi}) &= \mathbb{K}_{\mathcal{W}}(O_{\rho\gamma}, O_{\omega \chi} ) \nonumber\\&\quad + \frac{1}{2} n_{d} \sum_{\mu,\nu=1}^{p} (\sigma_{d}^{-2} \Gamma^{-1} G_{yy} \Gamma^{-1} - \Gamma^{-1})_{\mu\nu} \mathbb{K}_{\mathcal{W}}(O_{\rho\gamma}, O_{\omega \chi}, K_{\mu\nu} ) \nonumber\\&\quad + \mathcal{O}(\lambda^4).
\end{align}
Restoring the layer indices to $K = K^{(d-1)}$, the above result for $\langle O \rangle$ yields the expression \eqref{eqn:generalkernel} given in the main text. From the structure of these expressions, we can see that higher-order terms (in $\lambda$) will involve higher joint cumulants of the kernel deviations $\delta K^{(\ell)}$, which can in turn be converted into joint cumulants of the kernels $K^{(\ell)}$. Therefore, to show that the perturbative expansion yields a valid asymptotic series , one would need to show that these joint cumulants themselves have asymptotic series expansions at large width, with leading terms that are successively suppressed by powers of $n^{-1}$. 

\section{Explicit covariance computations in deep linear networks}\label{app:sec:cov}

In this appendix, we detail how to compute the prior covariances appearing in \eqref{eqn:generalkernel} for the hidden layer kernels of deep linear fully-connected and convolutional networks.  

\subsection{Fully-connected linear networks}

In this brief subsection, we provide a self-contained derivation of the behavior of the prior cumulants of the kernels of a deep fully-connected linear network with no bias terms. This is a special case of \citet{yaida2020}'s results, and provides some intuition for his results on general MLPs. As in the main text, we consider a network with activations $\mathbf{h}^{(\ell)} \in \mathbb{R}^{n_{\ell}}$ recursively defined as 
\begin{align}
    \mathbf{h}^{(\ell)} = n_{\ell-1}^{-1/2} W^{(\ell)} \mathbf{h}^{(\ell-1)} \qquad (\ell=1,\ldots,d)
\end{align}
with base case $\mathbf{h}^{(0)} = \mathbf{x}$. We take the prior distribution over weights to be $[W^{(\ell)}]_{ij} \sim_{\textrm{i.i.d.}} \mathcal{N}(0,\sigma_{\ell}^{(2)})$, and define the hidden layer kernels $[K^{(\ell)}]_{\mu\nu} \equiv n_{\ell}^{-1} \mathbf{h}_{\mu}^{(\ell)} \cdot \mathbf{h}_{\nu}^{(\ell)}$ for $\ell = 1, \ldots, d-1$. Then, we have
\begin{align}
    \mathbb{E}_{\mathcal{W}} K^{(\ell)}_{\mu\nu} 
    &=  \frac{1}{n_{\ell} \cdots n_{0}} \mathbb{E}_{\mathcal{W}} \mathbf{x}_{\mu}^{\top} (W^{(1)})^{\top} \cdots (W^{(\ell)})^{\top} W^{(\ell)} \cdots W^{(1)} \mathbf{x}_{\nu} 
    \\
    &= \sigma_{1}^{2} \cdots \sigma_{\ell}^{2} \frac{\mathbf{x}_{\mu} \cdot \mathbf{x}_{\nu}}{n_0} 
    \\
    &= [K^{(\ell)}_{\infty}]_{\mu\nu}
\end{align}
at any width, as $\mathbb{E}_{W^{(\ell)}} (W^{(\ell)})^{\top} W^{(\ell)} / n_{\ell} = \sigma_{\ell}^{2} I_{n_{\ell-1}}$. We now consider the second moments of the kernels. We first note that 
\begin{align}
    \mathbb{E}_{\mathcal{W}} K_{\mu\nu}^{(\ell)} K_{\rho\lambda}^{(\ell+\tau)} = \sigma_{\ell+\tau}^{2} \cdots \sigma_{\ell+1}^{2} \mathbb{E}_{\mathcal{W}} K_{\mu\nu}^{(\ell)} K_{\rho\lambda}^{(\ell)}
\end{align}
for any $\tau \geq 1$. By Isserlis' theorem (see Appendix \ref{app:sec:tech}), we have
\begin{align}
    \mathbb{E}_{W^{(\ell)}} W^{(\ell)}_{ik} W^{(\ell)}_{il} W^{(\ell)}_{jm} W^{(\ell)}_{jr} 
    = \sigma_{\ell}^{4} \delta_{ij} (\delta_{km} \delta_{lr} + \delta_{kr} \delta_{lm}) + \sigma_{\ell}^{4} \delta_{kl} \delta_{mr},
\end{align}
hence we have the exact recursion
\begin{align}
    \mathbb{E}_{\mathcal{W}} K_{\mu\nu}^{(\ell)} K_{\rho\lambda}^{(\ell)} 
    &= \frac{1}{(n_{\ell} \cdots n_{0})^{2}} \mathbb{E}_{\mathcal{W}} \sum_{i,j=1}^{n_{\ell}} \sum_{k,l,m,r=1}^{n_{\ell-1}} W^{(\ell)}_{ik} W^{(\ell)}_{il} W^{(\ell)}_{jm} W^{(\ell)}_{jr} 
    \nonumber\\&\qquad\qquad\qquad\qquad\qquad \times [W^{(\ell-1)} \cdots W^{(1)} \mathbf{x}_{\mu}]_{k} [W^{(\ell-1)} \cdots W^{(1)} \mathbf{x}_{\nu}]_{l} \nonumber\\&\qquad\qquad\qquad\qquad\qquad \times [W^{(\ell-1)} \cdots W^{(1)} \mathbf{x}_{\rho}]_{m}
     [W^{(\ell-1)} \cdots W^{(1)} \mathbf{x}_{\lambda}]_{r}
    \\
    &= \sigma_{\ell}^{4} \mathbb{E}_{\mathcal{W}} K_{\mu\nu}^{(\ell-1)} K_{\rho\lambda}^{(\ell-1)} + \frac{1}{n_{\ell}} \sigma_{\ell}^{4} (\mathbb{E}_{\mathcal{W}} K_{\mu\rho}^{(\ell-1)} K_{\nu\lambda}^{(\ell-1)} + \mathbb{E}_{\mathcal{W}} K_{\mu\lambda}^{(\ell-1)} K_{\nu\rho}^{(\ell-1)})
\end{align}
with base case
\begin{align}
    \mathbb{E}_{\mathcal{W}} K_{\mu\nu}^{(1)} K_{\rho\lambda}^{(1)} 
    &= \frac{1}{(n_{1} n_{0})^{2}}  \sum_{i,j=1}^{n_{1}} \sum_{k,l,m,r=1}^{n_{0}} \mathbb{E}_{\mathcal{W}} W^{(1)}_{ik} W^{(1)}_{il} W^{(1)}_{jm} W^{(1)}_{jr} x_{\mu,k} x_{\nu,l} x_{\rho,m} x_{\lambda,r}
    \\
    &= \sigma_{1}^{4} \frac{\mathbf{x}_{\mu} \cdot \mathbf{x}_{\nu}}{n_0} \frac{\mathbf{x}_{\rho} \cdot \mathbf{x}_{\lambda}}{n_0} + \frac{1}{n_1} \sigma_{1}^{4} \left(\frac{\mathbf{x}_{\mu} \cdot \mathbf{x}_{\rho}}{n_0} \frac{\mathbf{x}_{\nu} \cdot \mathbf{x}_{\lambda}}{n_0} + \frac{\mathbf{x}_{\mu} \cdot \mathbf{x}_{\lambda}}{n_0} \frac{\mathbf{x}_{\nu} \cdot \mathbf{x}_{\rho}}{n_0} \right)
    \\
    &= [K_{\infty}^{(1)}]_{\mu\nu} [K_{\infty}^{(1)}]_{\rho\lambda} + \frac{1}{n_1} \left([K_{\infty}^{(1)}]_{\mu\rho} [K_{\infty}^{(1)}]_{\nu\lambda} + [K_{\infty}^{(1)}]_{\mu\lambda} [K_{\infty}^{(1)}]_{\nu\rho}\right)
\end{align}
for the second moments of the kernels at each layer. This recurrence relation is in principle exactly solvable for any finite width, but we are interested only in its leading-order behavior at large widths. In particular, we can read off that 
\begin{align}
    \cov_{\mathcal{W}}(K_{\mu\nu}^{(\ell)}, K_{\rho\lambda}^{(\ell+\tau)}) &= \sigma_{\ell+\tau}^{2} \cdots \sigma_{\ell+1}^{2} \left(\sum_{\ell'=1}^{\ell} \frac{1}{n_{\ell'}}\right) \left([K_{\infty}^{(\ell)}]_{\mu\rho} [K_{\infty}^{(\ell)}]_{\nu\lambda} + [K_{\infty}^{(\ell)}]_{\mu\lambda} [K_{\infty}^{(\ell)}]_{\nu\rho}\right) \nonumber\\&\quad + \mathcal{O}(n^{-2}).
\end{align}
Moreover, one can see by Isserlis' theorem that the third and higher cumulants will be $\mathcal{O}(n^{-2})$. Substituting this result into \eqref{eqn:generalkernel} with the hidden layer kernel as the observable of interest, we obtain the expression \eqref{eqn:linearkernel} given in the main text. 

\subsection{Convolutional linear networks}

In this subsection, we derive the prior cumulants required to compute corrections to the average feature kernels of deep convolutional linear networks. As described in the main text, following the setup of \citet{novak2019bayesian} and \citet{xiao2018dynamical}, we consider a network consisting of $d-1$ linear convolutional layers followed by a fully-connected linear readout layer. For simplicity, we assume circular padding and no internal pooling. As discussed in \citet{novak2019bayesian}, this setup could be easily extended to other padding strategies, strided convolutions, and average pooling in intermediate layers.

We write the activations at the $\ell$-th hidden layer as $h_{i,\mathfrak{a}}^{(\ell)}$, where $i$ indexes the $n_{\ell}$ channels of the layer and $\mathfrak{a}$ is a $q$-dimensional spatial multi-index. We take the filters to be of size $(2k+1) \times \dots \times (2k+1)$ in all convolutional layers; the extension to differently-sized filters would be straightforward but notationally cumbersome. The ranges of all spatial summations will be implied.

The hidden layer activations are then defined through the recurrence
\begin{align}
    h_{i,\mathfrak{a}}^{(\ell)}(x) = \frac{1}{\sqrt{n_{\ell-1}}} \sum_{j=1}^{n_{\ell-1}} \sum_{\mathfrak{b}} w_{ij,\mathfrak{b}}^{(\ell)} h_{j,\mathfrak{a}+\mathfrak{b}}^{(\ell-1)}(x)
\end{align}
with base case $h_{i,\mathfrak{a}}^{(0)}(x) = x_{i,\mathfrak{a}}$. We fix the prior distribution of the filter elements to be
\begin{align}
    w_{ij,\mathfrak{a}}^{(\ell)} \underset{\textrm{i.i.d.}}{\sim} \mathcal{N}(0, \sigma_{\ell}^2 v_{\mathfrak{a}}),
\end{align}
where $v_{\mathfrak{a}} > 0$ is a weighting factor that sets the fraction of receptive field variance at location $\mathfrak{a}$ (and is thus subject to the constraint $\sum_{\mathfrak{a}} v_{\mathfrak{a}} = 1$). For inputs $[x_{\mu}]_{i,\mathfrak{a}}$ and $[x_{\nu}]_{i,\mathfrak{a}}$, we introduce the hidden layer kernels 
\begin{align}
    K^{(\ell)}_{\mu\nu, \mathfrak{a}\mathfrak{b}} \equiv \frac{1}{n_{\ell}} \sum_{i=1}^{n_{\ell}} h_{i,\mathfrak{a}}^{(\ell)}(x_{\mu}) h_{i,\mathfrak{b}}^{(\ell)}(x_{\nu}).
\end{align}
We will first compute the prior mean and covariance of these four-indexed kernels, and then address how to handle readout across space. 

As shown by \citet{xiao2018dynamical}, the prior mean obeys the recurrence
\begin{align}
    &\mathbb{E}_{\mathcal{W}} K_{\mu\nu,\mathfrak{a}\mathfrak{b}}^{(\ell)}
    \nonumber\\&\quad = \mathbb{E}_{W^{(1)}\cdots W^{(\ell-1)}} \frac{1}{n_{\ell} n_{\ell-1}} \sum_{i=1}^{n_{\ell}}  \sum_{j,j'=1}^{n_{\ell-1}} \sum_{\mathfrak{c},\mathfrak{d}} h_{j,\mathfrak{a}+\mathfrak{c}}^{(\ell-1)}(x_{\mu}) h_{j',\mathfrak{b}+\mathfrak{d}}^{(\ell-1)}(x_{\nu}) \mathbb{E}_{W^{(\ell)}} w_{ij,\mathfrak{c}}^{(\ell)} w_{ij',\mathfrak{d}}^{(\ell)}
    \\
    &\quad = \sigma_{\ell}^{2} \mathbb{E}_{W^{(1)}\cdots W^{(\ell-1)}} \sum_{\mathfrak{c}} v_{\mathfrak{c}} \frac{1}{n_{\ell-1}} \sum_{j=1}^{n_{\ell-1}}  h_{j,\mathfrak{a}+\mathfrak{c}}^{(\ell-1)}(x_{\mu}) h_{j,\mathfrak{b}+\mathfrak{c}}^{(\ell-1)}(x_{\nu})
    \\
    &\quad = \sigma_{\ell}^{2} \sum_{\mathfrak{c}} v_{\mathfrak{c}} \mathbb{E}_{\mathcal{W}} K_{\mu\nu,(\mathfrak{a}+\mathfrak{c})(\mathfrak{b}+\mathfrak{c})}^{(\ell-1)}
\end{align}
with base case
\begin{align}
    \mathbb{E}_{\mathcal{W}} K_{\mu\nu,\mathfrak{a}\mathfrak{b}}^{(1)} = \sigma_{1}^{2} \sum_{\mathfrak{c}} v_{\mathfrak{c}} [G_{xx}]_{\mu\nu,(\mathfrak{a}+\mathfrak{c})(\mathfrak{b}+\mathfrak{c})}
\end{align}
for
\begin{align}\label{eq:gram_matrix_tensor}
    [G_{xx}]_{\mu\nu,\mathfrak{a}\mathfrak{b}} \equiv \frac{1}{n_0} \sum_{i=1}^{n_0} [x_{\mu}]_{i,\mathfrak{a}} [x_{\nu}]_{i,\mathfrak{b}}.
\end{align}
This recurrence yields
\begin{align}
    \mathbb{E}_{\mathcal{W}} K_{\mu\nu,\mathfrak{a}\mathfrak{b}}^{(\ell)}
    = \sigma_{1}^{2} \cdots \sigma_{\ell}^{2} \sum_{\mathfrak{c}_{1},\ldots,\mathfrak{c}_{\ell}} v_{\mathfrak{c}_{1}} \cdots v_{\mathfrak{c}_{\ell}} [G_{xx}]_{\mu\nu,(\mathfrak{a} + \mathfrak{c}_{1} + \cdots + \mathfrak{c}_{\ell}) (\mathfrak{b} + \mathfrak{c}_{1} + \cdots + \mathfrak{c}_{\ell})}.
\end{align}
Moreover, as in the fully-connected case considered in the preceding section, we have
\begin{align}
    [K_{\infty}^{(\ell)}]_{\mu\nu,\mathfrak{a}\mathfrak{b}} = \mathbb{E}_{\mathcal{W}} K_{\mu\nu,\mathfrak{a}\mathfrak{b}}^{(\ell)}
\end{align}
at any width. 

We now consider the prior covariance of the kernels of two different hidden layers $\ell$ and $\ell+\tau$. As the weight prior factors across layers, we have
\begin{align}
    \mathbb{E}_{\mathcal{W}} K_{\mu\nu,\mathfrak{a}\mathfrak{b}}^{(\ell)} K_{\rho\lambda,\mathfrak{c}\mathfrak{d}}^{(\ell+\tau)} &= \sigma_{\ell+1}^{2} \cdots \sigma_{\ell+\tau}^{2} \sum_{\mathfrak{e}_{1},\ldots,\mathfrak{e}_{\tau}} v_{\mathfrak{e}_{1}} \cdots v_{\mathfrak{e}_{\tau}} \nonumber\\&\qquad\qquad\qquad\qquad \times \mathbb{E}_{\mathcal{W}} K_{\mu\nu,\mathfrak{a}\mathfrak{b}}^{(\ell)} K_{\rho\lambda,(\mathfrak{c}+\mathfrak{e}_{1}+\cdots+\mathfrak{e}_{\tau})(\mathfrak{d}+\mathfrak{e}_{1}+\cdots+\mathfrak{e}_{\tau})}^{(\ell)} .
\end{align}
By Isserlis' theorem (see Appendix \ref{app:sec:tech}), 
\begin{align}
    \mathbb{E}_{W^{(\ell)}} w_{ij,\mathfrak{e}}^{(\ell)} w_{ij',\mathfrak{f}}^{(\ell)} w_{i'j'',\mathfrak{g}}^{(\ell)} w_{i'j''',\mathfrak{h}}^{(\ell)}
    &= \sigma_{\ell}^{4} v_{\mathfrak{e}} v_{\mathfrak{g}} \delta_{jj'}  \delta_{j''j'''} \delta_{\mathfrak{e}\mathfrak{f}} \delta_{\mathfrak{g}\mathfrak{h}}
    \nonumber\\&\quad + \sigma_{\ell}^{4} v_{\mathfrak{e}} v_{\mathfrak{f}} \delta_{ii'} \delta_{jj''} \delta_{j'j'''} \delta_{\mathfrak{e}\mathfrak{g}} \delta_{\mathfrak{f}\mathfrak{h}}
    \nonumber\\&\quad + \sigma_{\ell}^{4} v_{\mathfrak{e}} v_{\mathfrak{f}} \delta_{ii'} \delta_{jj'''} \delta_{j'j''} \delta_{\mathfrak{e}\mathfrak{h}} \delta_{\mathfrak{f}\mathfrak{g}},
\end{align}
hence we have the recurrence
\begin{align}
    & \mathbb{E}_{\mathcal{W}} K_{\mu\nu,\mathfrak{a}\mathfrak{b}}^{(\ell)} K_{\rho\lambda,\mathfrak{c}\mathfrak{d}}^{(\ell)}
    \nonumber\\&\quad = \mathbb{E}_{W^{(1)} \cdots W^{(\ell-1)}} \frac{1}{n_{\ell}^2 n_{\ell-1}^{2}} \sum_{i,i'=1}^{n_{\ell}} \sum_{j,j',j'',j'''=1}^{n_{\ell-1}} \sum_{\mathfrak{e},\mathfrak{f},\mathfrak{g},\mathfrak{h}} \nonumber\\&\qquad\qquad\qquad\qquad\qquad\qquad \times h_{j,\mathfrak{a}+\mathfrak{e}}^{(\ell-1)}(x_{\mu}) h_{j',\mathfrak{b}+\mathfrak{f}}^{(\ell-1)}(x_{\nu}) h_{j'',\mathfrak{c}+\mathfrak{g}}^{(\ell-1)}(x_{\rho}) h_{j''',\mathfrak{d}+\mathfrak{h}}^{(\ell-1)}(x_{\lambda}) \nonumber\\&\qquad\qquad\qquad\qquad\qquad\qquad \times \mathbb{E}_{W^{(\ell)}} w_{ij,\mathfrak{e}}^{(\ell)} w_{ij',\mathfrak{f}}^{(\ell)} w_{i'j'',\mathfrak{g}}^{(\ell)} w_{i'j''',\mathfrak{h}}^{(\ell)}
    \\&\quad = \sigma_{\ell}^{4} \sum_{\mathfrak{e},\mathfrak{f}} v_{\mathfrak{e}} v_{\mathfrak{f}} \bigg[  \mathbb{E}_{\mathcal{W}} K^{(\ell-1)}_{\mu\nu,(\mathfrak{a}+\mathfrak{e}) (\mathfrak{b}+\mathfrak{e})} K^{(\ell-1)}_{\rho\lambda,(\mathfrak{c}+\mathfrak{f})(\mathfrak{d}+\mathfrak{f})} \nonumber\\&\qquad\qquad\qquad\qquad\qquad + \frac{1}{n_{\ell}} \mathbb{E}_{\mathcal{W}}  K^{(\ell-1)}_{\mu\rho,(\mathfrak{a}+\mathfrak{e})(\mathfrak{c}+\mathfrak{e})} K^{(\ell-1)}_{\nu\lambda,(\mathfrak{b}+\mathfrak{f})(\mathfrak{d}+\mathfrak{f})} \nonumber\\&\qquad\qquad\qquad\qquad\qquad + \frac{1}{n_{\ell}} \mathbb{E}_{\mathcal{W}}  K^{(\ell-1)}_{\mu\lambda,(\mathfrak{a}+\mathfrak{e})(\mathfrak{d}+\mathfrak{e})}  K^{(\ell-1)}_{\nu\rho,(\mathfrak{a}+\mathfrak{f})(\mathfrak{c}+\mathfrak{f})} \bigg] 
\end{align}
with base case
\begin{align}
    \mathbb{E}_{\mathcal{W}} K_{\mu\nu,\mathfrak{a}\mathfrak{b}}^{(1)} K_{\rho\lambda,\mathfrak{c}\mathfrak{d}}^{(1)} 
    &= \sigma_{1}^{4} \sum_{\mathfrak{e},\mathfrak{f}} v_{\mathfrak{e}} v_{\mathfrak{f}} \bigg[  [G_{xx}]_{\mu\nu,(\mathfrak{a}+\mathfrak{e}) (\mathfrak{b}+\mathfrak{e})} [G_{xx}]_{\rho\lambda,(\mathfrak{c}+\mathfrak{f})(\mathfrak{d}+\mathfrak{f})} \nonumber\\&\qquad\qquad\qquad\qquad\quad + \frac{1}{n_{\ell}} [G_{xx}]_{\mu\rho,(\mathfrak{a}+\mathfrak{e})(\mathfrak{c}+\mathfrak{e})} [G_{xx}]_{\nu\lambda,(\mathfrak{b}+\mathfrak{f})(\mathfrak{d}+\mathfrak{f})} \nonumber\\&\qquad\qquad\qquad\qquad\quad + \frac{1}{n_{\ell}} [G_{xx}]_{\mu\lambda,(\mathfrak{a}+\mathfrak{e})(\mathfrak{d}+\mathfrak{e})} [G_{xx}]_{\nu\rho,(\mathfrak{a}+\mathfrak{f})(\mathfrak{c}+\mathfrak{f})} \bigg] 
    \\
    &= [K_{\infty}^{(1)}]_{\mu\nu,\mathfrak{a}\mathfrak{b}} [K_{\infty}^{(1)}]_{\rho\lambda,\mathfrak{c}\mathfrak{d}}  \nonumber\\&\quad + \frac{1}{n_{\ell}} \bigg[ [K_{\infty}^{(1)}]_{\mu\rho,\mathfrak{a}\mathfrak{c}} [K_{\infty}^{(1)}]_{\nu\lambda,\mathfrak{b}\mathfrak{d}} + [K_{\infty}^{(1)}]_{\mu\lambda,\mathfrak{a}\mathfrak{d}} [K_{\infty}^{(1)}]_{\nu\rho,\mathfrak{b}\mathfrak{c}} \bigg]
\end{align}
for the second prior moments of the kernels. As in the fully-connected case, these recurrence relations could in principle be solved exactly, but we are only interested in their large-width behavior. Using the forward recurrence for the GP kernels, we can easily read off that
\begin{align}\label{eq:inter_covariance_conv}
    \cov_{\mathcal{W}}(K_{\mu\nu,\mathfrak{a}\mathfrak{b}}^{(\ell)},  K_{\rho\lambda,\mathfrak{c}\mathfrak{d}}^{(\ell)}) 
    &= \left(\sum_{\ell'=1}^{\ell} \frac{1}{n_{\ell'}}\right) \left( [K_{\infty}^{(\ell)}]_{\mu\rho,\mathfrak{a}\mathfrak{c}} [K_{\infty}^{(\ell)}]_{\nu\lambda,\mathfrak{b}\mathfrak{d}} + [K_{\infty}^{(\ell)}]_{\mu\lambda,\mathfrak{a}\mathfrak{d}} [K_{\infty}^{(\ell)}]_{\nu\rho,\mathfrak{b}\mathfrak{c}} \right) \nonumber\\&\quad  + \mathcal{O}(n^{-2}),
\end{align}
which can then be substituted into the desired cross-layer covariance:
\begin{align}\label{eq:cross_covariance_conv}
    \cov_{\mathcal{W}}(K_{\mu\nu,\mathfrak{a}\mathfrak{b}}^{(\ell)},  K_{\rho\lambda,\mathfrak{c}\mathfrak{d}}^{(\ell+\tau)})
    &= \sigma_{\ell+1}^{2} \cdots \sigma_{\ell+\tau}^{2} \sum_{\mathfrak{e}_{1},\ldots,\mathfrak{e}_{\tau}} v_{\mathfrak{e}_{1}} \cdots v_{\mathfrak{e}_{\tau}} \nonumber\\&\qquad\qquad\qquad\qquad \times  \cov_{\mathcal{W}}( K_{\mu\nu,\mathfrak{a}\mathfrak{b}}^{(\ell)}, K_{\rho\lambda,(\mathfrak{c}+\mathfrak{e}_{1}+\cdots+\mathfrak{e}_{\tau})(\mathfrak{d}+\mathfrak{e}_{1}+\cdots+\mathfrak{e}_{\tau})}^{(\ell)} ).
\end{align}

We now address the question of how to read out the convolutional layer activities across space. Following \citet{novak2019bayesian}, we consider two strategies: vectorization and projection. With vectorization, the output of the final convolutional layer is flattened into a $n_{d-1} s$-dimensional vector before readout, i.e., $\psi_{i + s(\mathfrak{a}-1)}(x) = h_{i,\mathfrak{a}}^{(d-1)}(x)$ or $\psi_{n_{d}(i-1)+\mathfrak{a}}(x) = h_{i,\mathfrak{a}}^{(d-1)}(x)$. The two-index feature map kernel appearing in Conjecture \ref{conj1} is then related to the four-index convolutional hidden layer kernel analyzed above via
\begin{align}\label{eq:feature_conv}
    K^{(d-1)}_{\mu\nu} = \frac{1}{s} \sum_{\mathfrak{a}} K^{(d-1)}_{\mu\nu,\mathfrak{a}\mathfrak{a}}. 
\end{align}
With projection, the feature map is formed by contracting the final convolutional layer with a fixed vector $\mathbf{u}$, i.e.,
\begin{align}
    \psi_{i}(x) = \sum_{\mathfrak{a}} u_{\mathfrak{a}} h_{i,\mathfrak{a}}^{(d-1)}(x).
\end{align}
The feature map kernel is then given as
\begin{align}\label{eq:feature_map}
    K_{\mu\nu}^{(d-1)} = \sum_{\mathfrak{a},\mathfrak{b}} u_{\mathfrak{a}} u_{\mathfrak{b}} K_{\mu\nu,\mathfrak{a}\mathfrak{b}}^{(d-1)}.
\end{align}
Examples of common projection readout strategies include global average pooling ($u_{\mathfrak{a}} = 1/s$) and single-pixel subsampling ($u_{\mathfrak{a}} = \delta_{\mathfrak{a}\mathfrak{c}}$ for some desired location $\mathfrak{c}$). These readout approaches endow the network with differing properties under spatial transformations; global average pooling has the particular property of making the output translation-invariant.

We now seek to simplify the resulting expression for the leading-order correction to the posterior mean of some four-index feature kernel $K^{(\ell)}_{\mu\nu,\mathfrak{a}\mathfrak{b}}$. Per Conjecture 1, the general form of this correction is
\begin{align}
    \frac{1}{2} n_{d} \sum_{\rho,\lambda=1}^{p} \Phi_{\rho\lambda} \cov_{\mathcal{W}}(K^{(\ell)}_{\mu\nu,\mathfrak{a}\mathfrak{b}}, K^{(d-1)}_{\rho\lambda}),
\end{align}
where we have defined $\Phi_{\rho\lambda} = [\sigma_{d}^{-2} \Gamma^{-1} G_{yy} \Gamma^{-1} - \Gamma^{-1}]_{\rho\lambda}$ for notational convenience. As elsewhere, $\Gamma \equiv K^{(d-1)}_{\infty} + \beta^{-1} \sigma_{d}^{-2} I_{p}$ for $K^{(d-1)}_{\infty}$ the two-index kernel determined by the chosen readout strategy. Depending on the chosen readout strategy, this general expression can be simplified dramatically. In particular, for vectorization or global average pooling, the correction does not depend on the particular form of $v_{\mathfrak{a}}$. 

To show this for vectorization (the strategy used in our experiments), we substitute the definition of $K^{(d-1)}_{\infty}$ from \eqref{eq:feature_conv} and the expression for the cross-layer kernel covariance from \eqref{eq:cross_covariance_conv} into the general expression for the correction to obtain
\begin{align}
    &\frac{n_{d}}{2s}\sigma_{\ell+1}^{2} \cdots \sigma_{d-1}^{2} \nonumber\\ &\times \sum_{\rho,\lambda} \Phi_{\rho\lambda} \sum_{\mathfrak{e}_{1},\cdots,\mathfrak{e}_{d-\ell-1}} v_{\mathfrak{e}_{1}} \cdots v_{\mathfrak{e}_{d-\ell-1}} \sum_{\mathfrak{c}} \cov_{\mathcal{W}}(K^{(\ell)}_{\mu\nu,\mathfrak{a}\mathfrak{b}}, K^{(\ell)}_{\rho\lambda,(\mathfrak{c}+\mathfrak{e}_{1}+\cdots+\mathfrak{e}_{d-\ell-1})(\mathfrak{c}+\mathfrak{e}_{1}+\cdots+\mathfrak{e}_{d-\ell-1})}).
\end{align}
Thanks to the periodic boundary conditions, the summation over $\mathfrak{c}$ is independent of the index shift $\mathfrak{e}_{1}+\cdots+\mathfrak{e}_{d-\ell-1}$. Then, the sums over  $\mathfrak{e}_{1},\cdots,\mathfrak{e}_{d-\ell-1}$ factor, yielding
\begin{align}
    \frac{n_{d}}{2s}&\sigma_{\ell+1}^{2} \cdots  \sigma_{d-1}^{2}\sum_{\rho,\lambda=1}^{p}\Phi_{\rho\lambda} \sum_{\mathfrak{c}} \cov_{\mathcal{W}}(K^{(\ell)}_{\mu\nu,\mathfrak{a}\mathfrak{b}}, K^{(\ell)}_{\rho\lambda,\mathfrak{c}\mathfrak{c}})
\end{align}
thanks to the normalization constraint $\sum_{\mathfrak{e}} v_{\mathfrak{e}} = 1$. We now notice that $\Phi_{\rho\lambda}$ is a symmetric matrix, and that the kernel remains invariant under the simultaneous exchange of indices $\rho \leftrightarrow \lambda$ and $\mathfrak{c} \leftrightarrow \mathfrak{d}$. Then, substituting in the expression for the same-layer kernel covariance \eqref{eq:inter_covariance_conv}, it is easy to show that the correction reduces to
\begin{align}
    \sigma_{\ell+1}^{2} \cdots \sigma_{d-1}^{2}\left(\sum_{\ell'=1}^{\ell} \frac{n_{d}}{n_{\ell'}}\right)& \frac{1}{s}\sum_{\mathfrak{c}} \sum_{\rho,\lambda=1}^{p}[K_{\infty}^{(\ell)}]_{\mu\rho,\mathfrak{a}\mathfrak{c}} \Phi_{\rho\lambda} [K_{\infty}^{(\ell)}]_{\lambda\nu,\mathfrak{c}\mathfrak{b}}.
\end{align}
This yields the expression given in the main text. 

For projection, an analogous simplification is possible in the case of global average pooling ($u_{\mathfrak{a}} = 1/s$). Substituting the definition of $K_\infty^{(d-1)}$ from \eqref{eq:feature_map} and expression for the cross-layer kernel covariance \eqref{eq:cross_covariance_conv} into the correction, we have
\begin{align}
    \frac{n_{d}}{2s^2}&\sigma_{\ell+1}^{2} \cdots  \sigma_{d-1}^{2}\sum_{\rho,\lambda=1}^{p}\Phi_{\rho\lambda} \sum_{\mathfrak{c},\mathfrak{d}} \cov_{\mathcal{W}}(K^{(\ell)}_{\mu\nu,\mathfrak{a}\mathfrak{b}}, K^{(\ell)}_{\rho\lambda,\mathfrak{c}\mathfrak{d}}).
\end{align}
Substituting in the expression for the same-layer kernel covariance \eqref{eq:inter_covariance_conv}, it is again easy to show that the correction reduces to
\begin{align}
    \sigma_{\ell+1}^{2} \cdots \sigma_{d-1}^{2}\left(\sum_{\ell'=1}^{\ell} \frac{n_{d}}{n_{\ell'}}\right)& \frac{1}{s^2}\sum_{\mathfrak{c},\mathfrak{d}} \sum_{\rho,\lambda=1}^{p}[K_{\infty}^{(\ell)}]_{\mu\rho,\mathfrak{a}\mathfrak{c}} \Phi_{\rho\lambda} [K_{\infty}^{(\ell)}]_{\lambda\nu,\mathfrak{d}\mathfrak{b}}.
\end{align}
For projection strategies other than global average pooling (more precisely, for strategies for which $u_{\mathfrak{a}}$ is not constant), the sum over indices in the cross-layer covariance is not independent of the shift, hence we cannot simplify the correction in a similar fashion. This can be seen explicitly when treating the case of single-pixel subsampling ($u_{\mathfrak{a}} = \delta_{\mathfrak{a}\mathfrak{c}}$ for some desired location $\mathfrak{c}$). In this case, the correction reduces to
\begin{align}
    &\frac{n_{d}}{2}\sigma_{\ell+1}^{2} \cdots \sigma_{d-1}^{2} \nonumber\\ &\times \sum_{\rho,\lambda} \Phi_{\rho\lambda} \sum_{\mathfrak{e}_{1},\cdots,\mathfrak{e}_{d-\ell-1}} v_{\mathfrak{e}_{1}} \cdots v_{\mathfrak{e}_{d-\ell-1}} \cov_{\mathcal{W}}(K^{(\ell)}_{\mu\nu,\mathfrak{a}\mathfrak{b}}, K^{(\ell)}_{\rho\lambda,(\mathfrak{c}+\mathfrak{e}_{1}+\cdots+\mathfrak{e}_{d-\ell-1})(\mathfrak{c}+\mathfrak{e}_{1}+\cdots+\mathfrak{e}_{d-\ell-1})}).
\end{align}
Unlike for vectorization or for projection using global average pooling, this expression is manifestly dependent on the form of $v_{\mathfrak{a}}$. 

Na{\"i}vely, the computation of the corrections to the linear convolutional kernels requires the computation of $\cov_{\mathcal{W}}( K_{\mu\nu,\mathfrak{a}\mathfrak{b}}^{(\ell)}, K_{\rho\lambda,(\mathfrak{c}+\mathfrak{e}_{1}+\cdots+\mathfrak{e}_{d-\ell-1})(\mathfrak{d}+\mathfrak{e}_{1}+\cdots+\mathfrak{e}_{d-\ell-1})}^{(\ell)} )$ for each index, which takes impractical amounts of compute time and storage. We only found it practical to compute the theoretical kernels in the special cases presented above.

\section{Direct computation of the average hidden layer kernels of a deep linear MLP}\label{app:sec:deeplinearkernel}

In this appendix, we provide a self-contained derivation of the average hidden layer kernels of a deep linear fully-connected network (MLP). This derivation relies upon neither the results of Appendices \ref{app:sec:perturbation} and \ref{app:sec:cov} nor those of \citet{yaida2020}. 

\subsection{The cumulant generating function of learned features for a MLP}\label{app:sec:mlp}

In this section, we briefly describe the full partition function of the Bayes posterior for a general fully connected network, or multi-layer perceptron (MLP), with no bias terms. An MLP $\mathbf{f}: \mathbb{R}^{n_{0}} \to \mathbb{R}^{n_{d}}$ with $d$ layers, no biases, and parameters $\Theta = \{W^{(\ell)}\}_{\ell=1}^{d}$ can be defined recursively in terms of its layer-wise preactivations $\mathbf{h}^{(\ell)} \in \mathbb{R}^{n_{\ell}}$ as
\begin{align}
    \mathbf{h}^{(0)} &= \mathbf{x},
    \\
    \mathbf{h}^{(\ell)} &= \frac{1}{\sqrt{n_{\ell-1}}} W^{(\ell)} \phi_{\ell-1}(\mathbf{h}^{(\ell-1)}) \quad (\ell = 1, \ldots, d), 
    \\
    \mathbf{f} &= \phi_{d}(\mathbf{h}^{(d)}),
\end{align}
where the activation functions $\phi_{\ell}$ act elementwise. As always, we focus on networks with linear readout, i.e., $\phi_{d}(x) = x$, and assume Gaussian priors over the weights:
\begin{align}
    W^{(\ell)}_{ij} \underset{\textrm{i.i.d.}}{\sim} \mathcal{N}(0, \sigma_{\ell}^2).
\end{align}
We enforce the definition of the network architecture via Fourier representations of the Dirac distribution, with $\mathbf{q}^{(\ell)}_{\mu}$ being the Lagrange multiplier that enforces the definition of the preactivation $\mathbf{h}^{(\ell)}_{\mu}$. Then, after integrating out the weights using the fact that the relevant integrals are Gaussian, this allows us to write the partition function as
\begin{align}
    Z &= \int \prod_{\mu=1}^{p} \prod_{\ell=1}^{d} \frac{d\mathbf{h}_{\mu}^{(\ell)}\, d\mathbf{q}_{\mu}^{(\ell)}}{(2\pi)^{n_{\ell}}} \exp\left[S(\{\mathbf{h}_{\mu}^{(\ell)}\},\{\mathbf{q}_{\mu}^{(\ell)}\}) \right], 
\end{align}
where the ``effective action'' for the preactivations and Lagrange multipliers is
\begin{align}
    S &= -\frac{1}{2} \beta \sum_{\mu=1}^{p} \Vert \mathbf{h}^{(d)}_{\mu} - \mathbf{y}_{\mu} \Vert^2 + \sum_{\ell=1}^{d} \sum_{\mu=1}^{p} i \mathbf{q}_{\mu}^{(\ell)} \cdot \mathbf{h}_{\mu}^{(\ell)} \nonumber\\&\quad - \frac{1}{2} \sum_{\ell=1}^{d} \frac{\sigma_{\ell}^2}{n_{\ell-1}}  \sum_{\mu,\nu=1}^{p} \mathbf{q}_{\mu}^{(\ell)} \cdot \mathbf{q}_{\nu}^{(\ell)}  \phi_{\ell-1}(\mathbf{h}^{(\ell-1)}_{\mu}) \cdot \phi_{\ell-1}(\mathbf{h}^{(\ell-1)}_{\nu}).
\end{align}
As described in Appendix \ref{app:sec:perturbation}, source terms can be added to the effective action to allow computation of various averages. For deep linear networks, it is convenient to scale the source terms by an overall factor of $-1/2$, for which we must correct when computing the averages:
\begin{align}
    S_{\textrm{J}} = - \frac{1}{2} \sum_{\ell=1}^{d-1} \sum_{\mu,\nu=1}^{p} J_{\mu\nu}^{(\ell)} \phi_{\ell}(\mathbf{h}_{\mu}^{(\ell)}) \cdot \phi_{\ell}(\mathbf{h}_{\nu}^{(\ell)}).
\end{align}
For an MLP, our task is therefore to integrate out the preactivations and corresponding Lagrange multipliers. We will do so sequentially from the first layer to the last, keeping terms up to the desired order at each step, akin to the approach of \citet{yaida2020}. So long as $n_{d}$ and $d$ are fixed and small relative to the width of the hidden layers, this is a consistent perturbative approach, as noted by \citet{yaida2020}.

\subsection{General form of the perturbative layer integrals for a deep linear network}\label{app:subsec:layerintegral}

In this section, we evaluate the general form of the integrals required to perturbatively marginalize out a given layer of a deep linear network to $\mathcal{O}(n^{-1})$. These integrals are generically of the form
\begin{align}
    I = \int \prod_{\mu=1}^{p} \frac{d\mathbf{h}_{\mu}\,d\mathbf{q}_{\mu}}{(2\pi)^{n_{2}}} \exp\Bigg(&\sum_{\mu=1}^{p} i \mathbf{q}_{\mu} \cdot \mathbf{h}_{\mu} - \frac{1}{2} \sum_{\mu,\nu=1}^{p} G_{\mu\nu} (\mathbf{q}_{\mu} \cdot \mathbf{q}_{\nu}) + \sum_{\mu=1}^{p} \mathbf{j}_{\mu} \cdot \mathbf{h}_{\mu} 
    \nonumber\\&\quad - \frac{1}{2} \frac{1}{n_{2}} \sum_{\mu,\nu=1}^{p} A_{\mu\nu} (\mathbf{h}_{\mu} \cdot \mathbf{h}_{\nu}) \nonumber\\&\quad + \frac{1}{4} \frac{g}{n_{1}} \sum_{\mu,\nu,\rho,\lambda=1}^{p} G_{\mu\nu} (\mathbf{q}_{\nu} \cdot \mathbf{q}_{\rho}) G_{\rho\lambda} (\mathbf{q}_{\lambda} \cdot \mathbf{q}_{\mu}) \nonumber\\&\quad + \frac{1}{2} \frac{1}{n_{1}} \sum_{\mu,\nu=1}^{p} B_{\mu\nu}  (\mathbf{q}_{\mu} \cdot \mathbf{q}_{\nu}) \Bigg),
\end{align}
where $\mathbf{h}_{\mu},\mathbf{q}_{\mu} \in \mathbb{R}^{n_2}$. Here, $G$ is a positive semidefinite matrix, while $A$ and $B$ are symmetric matrices that need not be positive semidefinite. Furthermore, $\mathbf{j}_{\mu}$ is some source, while $g$ is a coupling constant. We will first evaluate this integral up to terms of $\mathcal{O}(n_{1}^{-1})$ for $n_{1} \gg 1$, assuming that $G$, $A$, $B$, $\mathbf{j}_{\mu}$, and $g$ are $\mathcal{O}(1)$ functions of $n_{1}$, and then evaluate it up to terms of $\mathcal{O}(n_{1}^{-1},n_{2}^{-1})$ for $n_{1},n_{2} \gg 1$, assuming that $G$, $A$, $B$, $\mathbf{j}_{\mu}$, and $g$ are also $\mathcal{O}(1)$ functions of $n_{2}$.

We will proceed by evaluating the integrals for $G$ invertible, and then infer the general case by a continuity argument. We treat the quartic term perturbatively, and all other terms directly. Writing
\begin{align}
    C \equiv G - \frac{1}{n_1} B,
\end{align}
the leading term in the integral over $\mathbf{q}_{\mu}$ is
\begin{align}
    \frac{1}{(2\pi)^{n_{2}p/2}\det(C)^{n_{2}/2}} \exp\left(-\frac{1}{2} \sum_{\mu,\nu=1}^{p} C^{-1}_{\mu\nu} (\mathbf{h}_{\mu} \cdot \mathbf{h}_{\nu}) \right).
\end{align}
Multiplying and dividing by this quantity, we can compute the perturbative correction from the quartic term using the fact that $\mathbf{q}_{\mu}$ then behaves as a Gaussian random vector of mean $\bar{\mathbf{q}}_{\mu} = i \sum_{\nu=1}^{p} C^{-1}_{\mu\nu} \mathbf{h}_{\nu}$ and covariance $C^{-1}_{\mu\nu} I_{n_2}$. Denoting expectation with respect to this distribution as $\langle\!\langle \cdot \rangle\!\rangle_{q}$ and writing $\tilde{\mathbf{q}}_{\mu} \equiv \mathbf{q}_{\mu} - \bar{\mathbf{q}}_{\mu}$, Isserlis' theorem yields
\begin{align}
    \langle\!\langle (\mathbf{q}_{\nu} \cdot \mathbf{q}_{\rho})  (\mathbf{q}_{\lambda} \cdot \mathbf{q}_{\mu}) \rangle\!\rangle_{q}
    &= \langle\!\langle ([\tilde{\mathbf{q}}_{\nu} + \bar{\mathbf{q}}_{\nu}] \cdot [\tilde{\mathbf{q}}_{\rho} + \bar{\mathbf{q}}_{\rho}])  ([\tilde{\mathbf{q}}_{\lambda} + \bar{\mathbf{q}}_{\lambda}] \cdot [\tilde{\mathbf{q}}_{\mu} + \bar{\mathbf{q}}_{\mu}]) \rangle\!\rangle_{q}
    \\
    &= \langle\!\langle (\tilde{\mathbf{q}}_{\nu} \cdot \tilde{\mathbf{q}}_{\rho} + \tilde{\mathbf{q}}_{\nu} \cdot \bar{\mathbf{q}}_{\rho} + \bar{\mathbf{q}}_{\nu} \cdot \tilde{\mathbf{q}}_{\rho} + \bar{\mathbf{q}}_{\nu} \cdot \bar{\mathbf{q}}_{\rho}) \nonumber\\&\qquad\qquad \times (\tilde{\mathbf{q}}_{\lambda} \cdot \tilde{\mathbf{q}}_{\mu} + \tilde{\mathbf{q}}_{\lambda} \cdot \bar{\mathbf{q}}_{\mu} + \bar{\mathbf{q}}_{\lambda} \cdot \tilde{\mathbf{q}}_{\mu} + \bar{\mathbf{q}}_{\lambda} \cdot \bar{\mathbf{q}}_{\mu}) \rangle\!\rangle_{q}
    \\
    &= \langle\!\langle (\tilde{\mathbf{q}}_{\nu} \cdot \tilde{\mathbf{q}}_{\rho}) (\tilde{\mathbf{q}}_{\lambda} \cdot \tilde{\mathbf{q}}_{\mu}) \rangle\!\rangle_{q} + \langle\!\langle (\tilde{\mathbf{q}}_{\nu} \cdot \tilde{\mathbf{q}}_{\rho}) \rangle\!\rangle_{q} (\bar{\mathbf{q}}_{\lambda} \cdot \bar{\mathbf{q}}_{\mu})
    \nonumber\\&\quad + \langle\!\langle (\tilde{\mathbf{q}}_{\nu} \cdot \bar{\mathbf{q}}_{\rho}) (\tilde{\mathbf{q}}_{\lambda} \cdot \bar{\mathbf{q}}_{\mu}) \rangle\!\rangle_{q} + \langle\!\langle (\tilde{\mathbf{q}}_{\nu} \cdot \bar{\mathbf{q}}_{\rho}) (\bar{\mathbf{q}}_{\lambda} \cdot \tilde{\mathbf{q}}_{\mu}) \rangle\!\rangle_{q}
    \nonumber\\&\quad + \langle\!\langle  (\bar{\mathbf{q}}_{\nu} \cdot \tilde{\mathbf{q}}_{\rho}) (\tilde{\mathbf{q}}_{\lambda} \cdot \bar{\mathbf{q}}_{\mu}) \rangle\!\rangle_{q} + \langle\!\langle  (\bar{\mathbf{q}}_{\nu} \cdot \tilde{\mathbf{q}}_{\rho}) (\bar{\mathbf{q}}_{\lambda} \cdot \tilde{\mathbf{q}}_{\mu})  \rangle\!\rangle_{q}
    \nonumber\\&\quad + (\bar{\mathbf{q}}_{\nu} \cdot \bar{\mathbf{q}}_{\rho}) \langle\!\langle (\tilde{\mathbf{q}}_{\lambda} \cdot \tilde{\mathbf{q}}_{\mu}) \rangle\!\rangle_{q} + (\bar{\mathbf{q}}_{\nu} \cdot \bar{\mathbf{q}}_{\rho}) (\bar{\mathbf{q}}_{\lambda} \cdot \bar{\mathbf{q}}_{\mu})
    \\
    &= n_{2}^{2} C^{-1}_{\nu\rho} C^{-1}_{\lambda\mu} + n_{2} C^{-1}_{\nu\lambda} C^{-1}_{\rho\mu} + n_{2} C^{-1}_{\nu\mu} C^{-1}_{\rho\lambda} + n_{2} C^{-1}_{\nu\rho} (\bar{\mathbf{q}}_{\lambda} \cdot \bar{\mathbf{q}}_{\mu})
    \nonumber\\&\quad + C^{-1}_{\nu\lambda} (\bar{\mathbf{q}}_{\rho} \cdot \bar{\mathbf{q}}_{\mu}) + C^{-1}_{\nu\mu} (\bar{\mathbf{q}}_{\rho} \cdot \bar{\mathbf{q}}_{\lambda})
    \nonumber\\&\quad + C^{-1}_{\rho\lambda} (\bar{\mathbf{q}}_{\nu} \cdot \bar{\mathbf{q}}_{\mu}) + C^{-1}_{\rho\mu} (\bar{\mathbf{q}}_{\nu} \cdot \bar{\mathbf{q}}_{\lambda}) 
    \nonumber\\&\quad + n_{2} (\bar{\mathbf{q}}_{\nu} \cdot \bar{\mathbf{q}}_{\rho}) C^{-1}_{\mu\lambda} + (\bar{\mathbf{q}}_{\nu} \cdot \bar{\mathbf{q}}_{\rho}) (\bar{\mathbf{q}}_{\lambda} \cdot \bar{\mathbf{q}}_{\mu}).
\end{align}
Then, the quartic correction to the integral over $\mathbf{q}_{\mu}$ is proportional to
\begin{align}
    \sum_{\mu,\nu,\rho,\lambda=1}^{p} G_{\mu\nu} G_{\rho\lambda} \langle\!\langle (\mathbf{q}_{\nu} \cdot \mathbf{q}_{\rho})  (\mathbf{q}_{\lambda} \cdot \mathbf{q}_{\mu}) \rangle\!\rangle_{q}
    &= n_{2} (n_{2}+1) \tr(G C^{-1} G C^{-1}) + n_{2} \tr(G C^{-1})^2 
    \nonumber\\&\quad - 2 (n_{2} + 1) \tr(G C^{-1} G C^{-1} H C^{-1}) 
    \nonumber\\&\quad - 2 \tr(G C^{-1}) \tr(G C^{-1} H C^{-1})
    \nonumber\\&\quad + \tr(G C^{-1} H C^{-1} G C^{-1} H C^{-1}),
\end{align}
where we write $H_{\mu\nu} \equiv \mathbf{h}_{\mu} \cdot \mathbf{h}_{\nu}$. 

We now must integrate over $\mathbf{h}_{\mu}$. The leading term is simply
\begin{align}
    \det(CD)^{-n_{2}/2} \exp\left(\frac{1}{2} \sum_{\mu,\nu=1}^{p} D^{-1}_{\mu\nu} J_{\mu\nu}\right)
\end{align}
where we have defined
\begin{align}
    D \equiv C^{-1} + \frac{1}{n_{2}} A.
\end{align}
and $J_{\mu\nu} \equiv \mathbf{j}_{\mu} \cdot \mathbf{j}_{\nu}$. Multiplying and dividing by this quantity, we can compute the perturbative correction from the quartic term using the fact that $\mathbf{h}_{\mu}$ then behaves as a Gaussian random vector of mean $\bar{\mathbf{h}}_{\mu} = \sum_{\nu=1}^{p} D^{-1}_{\mu\nu} \mathbf{j}_{\nu}$ and covariance $D^{-1}_{\mu\nu} I_{n_2}$. We denote expectations with respect to this distribution by $\langle\!\langle \cdot \rangle\!\rangle_{h}$, and define $\tilde{\mathbf{h}}_{\mu} \equiv \mathbf{h}_{\mu} - \bar{\mathbf{h}}_{\mu}$. Then, we have
\begin{align}
    \langle\!\langle H_{\mu\nu} \rangle\!\rangle_{h} 
    = \langle\!\langle \mathbf{h}_{\mu} \cdot \mathbf{h}_{\nu} \rangle\!\rangle_{h} = \bar{\mathbf{h}}_{\mu} \cdot \bar{\mathbf{h}}_{\nu} + n_{2} D^{-1}_{\mu\nu},
\end{align}
and, by analogy to the corresponding four-point average for $\mathbf{q}_{\mu}$, 
\begin{align}
    \langle\!\langle (\mathbf{h}_{\nu} \cdot \mathbf{h}_{\rho})  (\mathbf{h}_{\lambda} \cdot \mathbf{h}_{\mu}) \rangle\!\rangle_{h}
    &= n_{2}^{2} D^{-1}_{\nu\rho} D^{-1}_{\lambda\mu} + n_{2} D^{-1}_{\nu\lambda} D^{-1}_{\rho\mu} + n_{2} D^{-1}_{\nu\mu} D^{-1}_{\rho\lambda} + n_{2} D^{-1}_{\nu\rho} (\bar{\mathbf{h}}_{\lambda} \cdot \bar{\mathbf{h}}_{\mu})
    \nonumber\\&\quad + D^{-1}_{\nu\lambda} (\bar{\mathbf{h}}_{\rho} \cdot \bar{\mathbf{h}}_{\mu}) + D^{-1}_{\nu\mu} (\bar{\mathbf{h}}_{\rho} \cdot \bar{\mathbf{h}}_{\lambda})
    \nonumber\\&\quad + D^{-1}_{\rho\lambda} (\bar{\mathbf{h}}_{\nu} \cdot \bar{\mathbf{h}}_{\mu}) + D^{-1}_{\rho\mu} (\bar{\mathbf{h}}_{\nu} \cdot \bar{\mathbf{h}}_{\lambda}) 
    \nonumber\\&\quad + n_{2} (\bar{\mathbf{h}}_{\nu} \cdot \bar{\mathbf{h}}_{\rho}) D^{-1}_{\mu\lambda} + (\bar{\mathbf{h}}_{\nu} \cdot \bar{\mathbf{h}}_{\rho}) (\bar{\mathbf{h}}_{\lambda} \cdot \bar{\mathbf{h}}_{\mu}).
\end{align}
Then, the correction to the integral over $\mathbf{h}_{\mu}$ is proportional to
\begin{align}
    \sum_{\mu,\nu,\rho,\lambda=1}^{p} G_{\mu\nu} G_{\rho\lambda} \langle\!\langle (\mathbf{q}_{\nu} \cdot \mathbf{q}_{\rho})  (\mathbf{q}_{\lambda} \cdot \mathbf{q}_{\mu}) \rangle\!\rangle
    &= n_{2} (n_{2}+1) \tr(G C^{-1} G C^{-1}) + n_{2} \tr(G C^{-1})^2 
    \nonumber\\&\quad - 2 (n_{2} + 1) \tr(G C^{-1} G C^{-1} D^{-1} J D^{-1} C^{-1}) 
    \nonumber\\&\quad - 2 n_{2} (n_{2} + 1) \tr(G C^{-1} G C^{-1} D^{-1} C^{-1}) 
    \nonumber\\&\quad - 2 \tr(G C^{-1}) \tr(G C^{-1} D^{-1} J D^{-1} C^{-1}) 
    \nonumber\\&\quad - 2 n_{2} \tr(G C^{-1}) \tr(G C^{-1} D^{-1} C^{-1})
    \nonumber\\&\quad + n_{2} (n_{2}+1) \tr(C^{-1} G C^{-1} D^{-1} C^{-1} G C^{-1} D^{-1}) 
    \nonumber\\&\quad + n_{2} \tr(C^{-1} G C^{-1} D^{-1})^2 
    \nonumber\\&\quad + 2 (n_{2} + 1) \tr(C^{-1} G C^{-1} D^{-1} C^{-1} G C^{-1} D^{-1} J D^{-1}) 
    \nonumber\\&\quad + 2 \tr(C^{-1} G C^{-1} D^{-1}) \tr(C^{-1} G C^{-1} D^{-1} J D^{-1})
    \nonumber\\&\quad + \tr(C^{-1} G C^{-1} D^{-1} J D^{-1} C^{-1} G C^{-1} D^{-1} J D^{-1}),
\end{align}
where we have noted that
\begin{align}
    &\langle\!\langle \tr(G C^{-1} H C^{-1} G C^{-1} H C^{-1}) \rangle\!\rangle_{h} \nonumber\\&\quad = \sum_{\mu,\nu,\rho,\lambda=1}^{p} (C^{-1} G C^{-1})_{\mu\nu} (C^{-1} G C^{-1})_{\rho\lambda} \langle\!\langle (\mathbf{h}_{\nu} \cdot \mathbf{h}_{\rho})  (\mathbf{h}_{\lambda} \cdot \mathbf{h}_{\mu}) \rangle\!\rangle_{h}
    \\&\quad = n_{2} (n_{2}+1) \tr(C^{-1} G C^{-1} D^{-1} C^{-1} G C^{-1} D^{-1}) + n_{2} \tr(C^{-1} G C^{-1} D^{-1})^2 
    \nonumber\\&\qquad + 2 (n_{2} + 1) \tr(C^{-1} G C^{-1} D^{-1} C^{-1} G C^{-1} D^{-1} J D^{-1}) 
    \nonumber\\&\qquad + 2 \tr(C^{-1} G C^{-1} D^{-1}) \tr(C^{-1} G C^{-1} D^{-1} J D^{-1})
    \nonumber\\&\qquad + \tr(C^{-1} G C^{-1} D^{-1} J D^{-1} C^{-1} G C^{-1} D^{-1} J D^{-1})
\end{align}
by analogy with the corresponding quartic expectation for $\mathbf{q}_{\mu}$.

We must now expand our results in $n_{1}^{-1}$. The inverses of the matrices $C$ and $D$ have Neumann series
\begin{align}
    C^{-1} = G^{-1} + \frac{1}{n_1} G^{-1} B G^{-1} + \mathcal{O}(n_1^{-2})
\end{align}
and
\begin{align}
    D^{-1} 
    &= \left(C^{-1} + \frac{1}{n_2} A\right)^{-1} 
    \\
    &= \left(G^{-1} + \frac{1}{n_1} G^{-1} B G^{-1} + \frac{1}{n_2} A  + \mathcal{O}(n_1^{-2}) \right)^{-1} 
    \\
    &= F^{-1} G - \frac{1}{n_1} F^{-1} B F^{-\top} + \mathcal{O}(n_{1}^{-2})
\end{align}
where we have defined
\begin{align}
    F \equiv I_{p} + \frac{1}{n_2} G A
\end{align}
and we write $F^{-\top} = (F^{-1})^{\top} = (F^{\top})^{-1}$. Then, using the series expansion of the log-determinant, we find that the logarithm of the leading term expands as
\begin{align}
    \frac{1}{2} \tr( D^{-1} J ) - \frac{1}{2} n_{2} \log \det(CD)
    &= \frac{1}{2} \tr( F^{-1} G J ) - \frac{1}{2} n_{2} \log\det(F) 
    \nonumber\\&\quad - \frac{1}{2} \frac{1}{n_{1}} \tr(F^{-1} B F^{-\top} J) + \frac{1}{2} \frac{1}{n_{1}} \tr(F^{-1} B A)
    \nonumber\\&\quad + \mathcal{O}(n_{1}^{-2}),
\end{align}
while the quartic correction simplifies to
\begin{align}
    &\frac{1}{4} \frac{g}{n_1} \sum_{\mu,\nu,\rho,\lambda=1}^{p} G_{\mu\nu} G_{\rho\lambda} \langle\!\langle (\mathbf{q}_{\nu} \cdot \mathbf{q}_{\rho})  (\mathbf{q}_{\lambda} \cdot \mathbf{q}_{\mu}) \rangle\!\rangle
    \nonumber\\&\quad= \frac{1}{4} \frac{g}{n_1} n_{2} (n_{2}+p+1) p
    \nonumber\\&\qquad
    + \frac{1}{4} \frac{n_{2} g}{n_1} \bigg( (n_{2}+1) \tr(F^{-2}) + \tr(F^{-1})^2 - 2 (n_{2} + p + 1) \tr(F^{-1}) \bigg)
    \nonumber\\&\qquad + \frac{1}{2} \frac{g}{n_1}\bigg((n_{2} + 1) \tr( F^{-3} G J ) + \tr(F^{-1}) \tr(F^{-2} G J) - (n_{2} + p + 1) \tr(F^{-2} G J) \bigg)
    \nonumber\\&\qquad + \frac{1}{4} \frac{g}{n_1} \tr( F^{-2} G J F^{-2} G J)
    \nonumber\\&\qquad + \mathcal{O}(n_{1}^{-2}).
\end{align}
Combining these results, we find that the result of integrating out the layer to $\mathcal{O}(n_{1}^{-1})$ is
\begin{align}
    \log I &= \frac{1}{2} \tr( F^{-1} G J ) - \frac{1}{2} n_{2} \log\det(F) 
    \nonumber\\&\quad - \frac{1}{2} \frac{1}{n_{1}} \tr(F^{-1} B F^{-\top} J) + \frac{1}{2} \frac{1}{n_{1}} \tr(F^{-1} B A)
    \nonumber\\&\quad + \frac{1}{4} \frac{g}{n_1} n_{2} (n_{2}+p+1) p
    \nonumber\\&\quad
    + \frac{1}{4} \frac{n_{2} g}{n_1} \bigg( (n_{2}+1) \tr(F^{-2}) + \tr(F^{-1})^2 - 2 (n_{2} + p + 1) \tr(F^{-1}) \bigg)
    \nonumber\\&\quad + \frac{1}{2} \frac{g}{n_1}\bigg((n_{2} + 1) \tr( F^{-3} G J ) + \tr(F^{-1}) \tr(F^{-2} G J) - (n_{2} + p + 1) \tr(F^{-2} G J) \bigg)
    \nonumber\\&\quad + \frac{1}{4} \frac{g}{n_1} \tr( F^{-2} G J F^{-2} G J)
    \nonumber\\&\quad + \mathcal{O}(n_{1}^{-2}).
\end{align}
As this result is a continuous function of $G$, as the set of full-rank positive definite matrices is dense in the space of positive semidefinite matrices, this result holds for all positive-semidefinite $G$.

We now further expand this result in $n_{2}^{-1}$. This yields
\begin{align}
    F^{-1} = I_{p} - \frac{1}{n_2} G A + \frac{1}{n_2^2} G A G A + \mathcal{O}(n_{2}^{-3})
\end{align}
and
\begin{align}
    \log\det(F) = \frac{1}{n_2} \tr(G A) - \frac{1}{2} \frac{1}{n_2^2} \tr(G A G A) + \mathcal{O}(n_{2}^{-3}),
\end{align}
hence we find that the logarithm of the leading term yields
\begin{align}
    \frac{1}{2} \tr( D^{-1} J ) - \frac{1}{2} n_{2} \log \det(CD)
    &= \frac{1}{2} \tr( G J ) - \frac{1}{2} \tr(G A) + \frac{1}{4} \frac{1}{n_2} \tr(G A G A) 
    \nonumber\\&\quad - \frac{1}{2} \frac{1}{n_2} \tr( G A G J ) + \frac{1}{2} \frac{1}{n_{1}} \tr(B (A-J))
    \nonumber\\&\quad + \mathcal{O}(n_{1}^{-2},n_{2}^{-2},n_{1}^{-1} n_{2}^{-1}).
\end{align}
After some straightforward but tedious algebra, the quartic term reduces to
\begin{align}
    \frac{1}{4} \frac{g}{n_1} \sum_{\mu,\nu,\rho,\lambda=1}^{p} G_{\mu\nu} G_{\rho\lambda} \langle\!\langle (\mathbf{q}_{\nu} \cdot \mathbf{q}_{\rho})  (\mathbf{q}_{\lambda} \cdot \mathbf{q}_{\mu}) \rangle\!\rangle &= \frac{1}{4} \frac{g}{n_1} \tr(G (A-J) G (A-J) ) \nonumber\\&\quad + \mathcal{O}(n_{1}^{-2},n_{2}^{-2},n_{1}^{-1} n_{2}^{-1}).
\end{align}
Combining these results, we find that the result of integrating out the layer is
\begin{align}
    \log I &= \frac{1}{2} \tr( G J ) - \frac{1}{2} \tr(G A) + \frac{1}{4} \frac{1}{n_2} \left(1 + \frac{n_2}{n_1} g\right) \tr(G A G A) 
    \nonumber\\&\quad - \frac{1}{2} \left(1 + \frac{n_2}{n_1} g\right) \tr( G A G J ) + \frac{1}{2} \frac{1}{n_{1}} \tr(B (A-J))
    + \frac{1}{4} \frac{1}{n_1} g \tr(G J G J)
    \nonumber\\&\quad + \mathcal{O}(n_{1}^{-2},n_{2}^{-2},n_{1}^{-1} n_{2}^{-1}).
\end{align}
Again, this result is continuous in $G$, hence it holds even if $G$ is rank-deficient. 

\subsection{Perturbative computation of the partition function of a deep linear network}

We now apply the results of Appendix \ref{app:subsec:layerintegral} to compute the partition function for a deep linear network to the desired order. Our starting point is the effective action before any of the layers have been integrated out, including a source term:
\begin{align}
    S &= -\frac{1}{2} \beta \sum_{\mu=1}^{p} \Vert \mathbf{h}^{(d)}_{\mu} - \mathbf{y}_{\mu} \Vert^2 + \sum_{\ell=1}^{d} \sum_{\mu=1}^{p} i \mathbf{q}_{\mu}^{(\ell)} \cdot \mathbf{h}_{\mu}^{(\ell)}  - \frac{1}{2} \sum_{\mu,\nu=1}^{p} (\sigma_{1}^{2} G_{xx})_{\mu\nu} (\mathbf{q}_{\mu}^{(1)} \cdot \mathbf{q}_{\nu}^{(1)})
    \nonumber\\&\quad - \frac{1}{2} \sum_{\ell=1}^{d-1} \frac{1}{n_{\ell}}  \sum_{\mu,\nu=1}^{p} ( J_{\mu\nu}^{(\ell)} + \sigma_{\ell+1}^2 \mathbf{q}_{\mu}^{(\ell+1)} \cdot \mathbf{q}_{\nu}^{(\ell+1)} )  (\mathbf{h}^{(\ell)}_{\mu} \cdot \mathbf{h}^{(\ell)}_{\nu}) .
\end{align}
Applying the results of Appendix \ref{app:subsec:layerintegral} with
\begin{align}
\begin{split}
    G &= \sigma_{1}^{2} G_{xx},
    \\
    \mathbf{j}_{\mu} &= 0,
    \\
    A &= J^{(1)} + \sigma_{2}^{2} Q^{(2)},
    \\
    B &= 0, \quad \textrm{and}
    \\
    g &= 0,
\end{split}
\end{align}
we find that the effective action after integrating out the first layer is
\begin{align}
    S^{(1)} &= -\frac{1}{2} \beta \sum_{\mu=1}^{p} \Vert \mathbf{h}^{(d)}_{\mu} - \mathbf{y}_{\mu} \Vert^2 + \sum_{\ell=2}^{d} \sum_{\mu=1}^{p} i \mathbf{q}_{\mu}^{(\ell)} \cdot \mathbf{h}_{\mu}^{(\ell)} -  \frac{1}{2} \sum_{\mu,\nu=1}^{p} (m_{2}^{2} G_{xx})_{\mu\nu} (\mathbf{q}_{\mu}^{(2)} \cdot \mathbf{q}_{\nu}^{(2)})
    \nonumber\\&\quad - \frac{1}{2} \sum_{\ell=2}^{d-1} \frac{1}{n_{\ell}}  \sum_{\mu,\nu=1}^{p} ( J_{\mu\nu}^{(\ell)} + \sigma_{\ell+1}^2 \mathbf{q}_{\mu}^{(\ell+1)} \cdot \mathbf{q}_{\nu}^{(\ell+1)} )  (\mathbf{h}^{(\ell)}_{\mu} \cdot \mathbf{h}^{(\ell)}_{\nu})
    \nonumber\\&\quad + \frac{1}{4} \frac{g_{1}}{n_{1}} m_{2}^{4} \tr(G_{xx} Q^{(2)} G_{xx} Q^{(2)}) + \frac{1}{2} \frac{g_{1}}{n_1} m_{2}^{2} \tr(G_{xx} \tilde{J}_{1} G_{xx} Q^{(2)})
    \nonumber\\&\quad - \frac{1}{2} \tr(m_{1}^{2} G_{xx} J^{(1)}) + \frac{1}{4} \frac{g_{1}}{n_{1}} m_{1}^{4} \tr(G_{xx} J^{(1)} G_{xx} J^{(1)})
    \nonumber\\&\quad + \mathcal{O}(n^{-2}),
\end{align}
where we have defined
\begin{align}
\begin{split}
    m_{1} &\equiv \sigma_{1},
    \\
    m_{2} &\equiv \sigma_{2} m_{1},
    \\
    g_{1} &\equiv 1, \quad \textrm{and}
    \\
    \tilde{J}_{1} &\equiv m_{1}^{2} J^{(1)}.
\end{split}
\end{align}
Assuming that the network has more than one hidden layer, if we now again apply the results of Appendix \ref{app:subsec:layerintegral} with
\begin{align}
\begin{split}
    G &= m_{2}^{2} G_{xx},
    \\
    \mathbf{j}_{\mu} &= 0,
    \\
    A &= J^{(2)} + \sigma_{3}^{2} Q^{(3)},
    \\
    B &= g_{1} m_{2}^{2} G_{xx} \tilde{J}_{1} G_{xx}, \quad \textrm{and}
    \\
    g &= g_{1},
\end{split}
\end{align}
we find that the effective action after integrating out the first two layers is
\begin{align}
    S^{(2)} &= -\frac{1}{2} \beta \sum_{\mu=1}^{p} \Vert \mathbf{h}^{(d)}_{\mu} - \mathbf{y}_{\mu} \Vert^2 + \sum_{\ell=3}^{d} \sum_{\mu=1}^{p} i \mathbf{q}_{\mu}^{(\ell)} \cdot \mathbf{h}_{\mu}^{(\ell)} - \frac{1}{2} \sum_{\mu,\nu=1}^{p} (m_{3}^{2} G_{xx})_{\mu\nu} (\mathbf{q}_{\mu}^{(3)} \cdot \mathbf{q}_{\nu}^{(3)})
    \nonumber\\&\quad - \frac{1}{2} \sum_{\ell=3}^{d-1} \frac{1}{n_{\ell}}  \sum_{\mu,\nu=1}^{p} ( J_{\mu\nu}^{(\ell)} + \sigma_{\ell+1}^2 \mathbf{q}_{\mu}^{(\ell+1)} \cdot \mathbf{q}_{\nu}^{(\ell+1)} )  (\mathbf{h}^{(\ell)}_{\mu} \cdot \mathbf{h}^{(\ell)}_{\nu})
    \nonumber\\&\quad + \frac{1}{4} \frac{g_{2}}{n_{2}} m_{3}^{4} \tr(G_{xx} Q^{(3)} G_{xx} Q^{(3)}) 
    \nonumber\\&\quad + \frac{1}{2} \frac{g_2}{n_2} m_{3}^{2} \tr(G_{xx} \tilde{J}_{2} G_{xx} Q^{(3)})
    \nonumber\\&\quad - \frac{1}{2} \tr(m_{1}^{2} G_{xx} J^{(1)}) - \frac{1}{2} \tr(m_{2}^{2} G_{xx} J^{(2)})
    \nonumber\\&\quad + \frac{1}{4} \frac{g_{1}}{n_{1}} m_{1}^{4} \tr(G_{xx} J^{(1)} G_{xx} J^{(1)}) + \frac{1}{4} \frac{g_{2}}{n_{2}} m_{2}^{4} \tr(G_{xx} J^{(2)} G_{xx} J^{(2)})
    \nonumber\\&\quad + \frac{1}{2} \frac{g_{1}}{n_1} m_{2}^{2} \tr(G_{xx} \tilde{J}_{1} G_{xx} J^{(2)})
    \nonumber\\&\quad + \mathcal{O}(n^{-2}),
\end{align}
where we have defined
\begin{align}
\begin{split}
    m_{3} &\equiv \sigma_{3} m_{2},
    \\
    g_{2} &\equiv 1 + \frac{n_{2}}{n_{1}} g_{1}, \quad \textrm{and}
    \\
    \tilde{J}_{2} &\equiv m_{2}^{2} J^{(2)} + \frac{n_2}{n_1} \frac{g_1}{g_2} \tilde{J}_{1}.
\end{split}
\end{align}
Then, by induction, we can see that we can iterate this procedure to integrate out all of the hidden layers, yielding
\begin{align}
    S^{(d-1)} &= -\frac{1}{2} \beta \sum_{\mu=1}^{p} \Vert \mathbf{h}^{(d)}_{\mu} - \mathbf{y}_{\mu} \Vert^2 + \sum_{\mu=1}^{p} i \mathbf{q}_{\mu}^{(d)} \cdot \mathbf{h}_{\mu}^{(d)} - \frac{1}{2} \sum_{\mu,\nu=1}^{p} (m_{d}^{2} G_{xx})_{\mu\nu} (\mathbf{q}_{\mu}^{(d)} \cdot \mathbf{q}_{\nu}^{(d)})
    \nonumber\\&\quad + \frac{1}{4} \frac{g_{d-1}}{n_{d-1}} m_{d}^{4} \tr(G_{xx} Q^{(d)} G_{xx} Q^{(d)}) 
    \nonumber\\&\quad + \frac{1}{2} \frac{g_{d-1}}{n_{d-1}} m_{d}^{2} \tr(G_{xx} \tilde{J}_{d-1} G_{xx} Q^{(d)})
    \nonumber\\&\quad - \frac{1}{2} \sum_{\ell=1}^{d-1} \tr(m_{\ell}^{2} G_{xx} J^{(\ell)})
    \nonumber\\&\quad + \frac{1}{4} \sum_{\ell=1}^{d-1} \frac{g_{\ell}}{n_{\ell}} m_{\ell}^{4} \tr(G_{xx} J^{(\ell)} G_{xx} J^{(\ell)}) 
    \nonumber\\&\quad + \frac{1}{2} \sum_{\ell=1}^{d-2} \frac{g_{\ell}}{n_{\ell}} m_{\ell+1}^{2} \tr(G_{xx} \tilde{J}_{\ell} G_{xx} J^{(\ell+1)})
    \nonumber\\&\quad + \mathcal{O}(n^{-2}),
\end{align}
where $m_{d}$, $g_{d-1}$, and $\tilde{J}_{d-1}$ are defined by the closed recurrences
\begin{align}
    m_{\ell} &\equiv \sigma_{\ell} m_{\ell-1},
    \\
    g_{\ell} &\equiv 1 + \frac{n_{\ell}}{n_{\ell-1}} g_{\ell-1}, \quad \textrm{and}
    \\
    \tilde{J}_{\ell} &\equiv m_{\ell}^{2} J^{(\ell)} + \frac{n_{\ell}}{n_{\ell-1}} \frac{g_{\ell-1}}{g_{\ell}} \tilde{J}_{\ell-1}.
\end{align}
Applying the results of Appendix \ref{app:subsec:layerintegral} one final time with
\begin{align}
\begin{split}
    G &= m_{d}^{2} G_{xx},
    \\
    \mathbf{j}_{\mu} &= \beta \mathbf{y},
    \\
    A &= \beta n_{d} I_{p},
    \\
    B &= g_{d-1} m_{d}^{2} G_{xx} \tilde{J}_{d-1} G_{xx}, \quad \textrm{and}
    \\
    g &= g_{d-1},
\end{split}
\end{align}
we conclude that
\begin{align}
    \log Z &= - \frac{1}{2} \beta n_{d} \tr(\tilde{\Gamma}^{-1} G_{yy}) - \frac{1}{2} n_{d} \log\det(\tilde{\Gamma}) 
    \nonumber\\&\quad + \frac{1}{4} \frac{n_{d} g_{d-1}}{n_{d-1}} \bigg((n_{d}+p+1) p + (n_{d}+1) \tr(\tilde{\Gamma}^{-2}) + \tr(\tilde{\Gamma}^{-1})^2 - 2 (n_{d} + p + 1) \tr(\tilde{\Gamma}^{-1}) \bigg)
    \nonumber\\&\quad + \frac{1}{2} \frac{g_{d-1}}{n_{d-1}} \beta^{2} n_{d} m_{d}^{2} \bigg((n_{d} + 1) \tr( \tilde{\Gamma}^{-3} G_{xx} G_{yy} ) + \tr(\tilde{\Gamma}^{-1}) \tr(\tilde{\Gamma}^{-2} G_{xx} G_{yy}) \nonumber\\&\qquad\qquad\qquad\qquad\qquad - (n_{d} + p + 1) \tr(\tilde{\Gamma}^{-2} G_{xx} G_{yy}) \bigg)
    \nonumber\\&\quad + \frac{1}{4} \frac{g_{d-1}}{n_{d-1}} \beta^{4} n_{d}^{2} m_{d}^{4} \tr( \tilde{\Gamma}^{-2} G_{xx} G_{yy} \tilde{\Gamma}^{-2} G_{xx} G_{yy})
    \nonumber\\&\quad - \frac{1}{2} \frac{g_{d-1}}{n_{d-1}} n_{d}  m_{d}^{2} \tr\left[\bigg(\beta^{2} G_{xx} \tilde{\Gamma}^{-1} G_{yy} \tilde{\Gamma}^{-1} G_{xx} - \beta G_{xx} \tilde{\Gamma}^{-1} G_{xx}\bigg) \tilde{J}_{d-1} \right]
    \nonumber\\&\quad - \frac{1}{2} \sum_{\ell=1}^{d-1} \tr(m_{\ell}^{2} G_{xx} J^{(\ell)})
    \nonumber\\&\quad + \frac{1}{4} \sum_{\ell=1}^{d-1} \frac{g_{\ell}}{n_{\ell}} m_{\ell}^{4} \tr(G_{xx} J^{(\ell)} G_{xx} J^{(\ell)}) 
    \nonumber\\&\quad + \frac{1}{2} \sum_{\ell=1}^{d-2} \frac{g_{\ell}}{n_{\ell}} m_{\ell+1}^{2} \tr(G_{xx} \tilde{J}_{\ell} G_{xx} J^{(\ell+1)})
    \nonumber\\&\quad + \mathcal{O}(n^{-2}),
\end{align}
where we have defined the matrix
\begin{align}
    \tilde{\Gamma} \equiv I_{p} + \beta m_{d}^{2} G_{xx}
\end{align}
and absorbed the normalizing constant using the fact that $I_{p} - \beta m_{d}^{2} \tilde{\Gamma}^{-1} G_{xx} = \tilde{\Gamma}^{-1}$. As was the case for the individual layer integrals, a continuity argument implies that this expression can be applied even if $G_{xx}$ is rank-deficient. 

\subsection{Computing the average hidden layer kernels of a deep linear network}

With the relevant partition function in hand, we can finally compute the average hidden layer kernels. In particular, we can immediately read off that
\begin{align}
    \langle K^{(\ell)} \rangle &= m_{\ell}^{2} G_{xx} \nonumber\\&\quad + \frac{g_{d-1}}{n_{d-1}} n_{d}  m_{d}^{2} \tr\left[\bigg(\beta^{2} G_{xx} \tilde{\Gamma}^{-1} G_{yy} \tilde{\Gamma}^{-1} G_{xx} - \beta G_{xx} \tilde{\Gamma}^{-1} G_{xx}\bigg)  \frac{\delta \tilde{J}_{d-1}}{\delta J^{(\ell)}} \bigg|_{J^{(\ell)} = 0}\right]  \nonumber\\&\quad + \mathcal{O}(n^{-2}),
\end{align}
hence our only task is to determine how the effective source $\tilde{J}_{d-1}$ depends on the source for a given layer. Fortunately, the recurrence relation for the effective source is extremely easy to solve, yielding
\begin{align}
    \tilde{J}_{d-1} = \sum_{\ell=1}^{d-1} m_{\ell}^{2} \frac{n_{d-1}}{n_{\ell}} \frac{g_{\ell}}{g_{d-1}} J^{(\ell)}.
\end{align}
Thus, defining the matrix
\begin{align}
    \Gamma \equiv \frac{1}{\beta m_{d}^{2}} \tilde{\Gamma} = G_{xx} + \frac{1}{\beta m_{d}^{2}} I_{p},
\end{align}
we find that
\begin{align}
    \langle K^{(\ell)} \rangle &= m_{\ell}^{2} G_{xx} + \frac{g_{\ell}}{n_{\ell}} n_{d}  m_{\ell}^{2} \bigg(m_{d}^{-2} G_{xx} \Gamma^{-1} G_{yy} \Gamma^{-1} G_{xx} - G_{xx} \Gamma^{-1} G_{xx}\bigg) + \mathcal{O}(n^{-2}).
\end{align}
To obtain the expression listed in the main text, we note that
\begin{align}
    \frac{g_{\ell}}{n_{\ell}} = \frac{1}{n_{\ell}} + \frac{g_{\ell-1}}{n_{\ell-1}},
\end{align}
hence we have
\begin{align}
    \frac{g_{\ell}}{n_{\ell}} = \sum_{\ell'=1}^{\ell} \frac{1}{n_{\ell'}},
\end{align}
mirroring the width dependence found by \citet{yaida2020} in his study of the prior of deep linear networks.

\section{Average kernels in a deep feedforward linear network with skip connections}\label{app:sec:skip}

In this appendix, we show that Conjecture \ref{conj1} holds perturbatively for a linear feedforward network with arbitrary skip connections, following the method of Appendix \ref{app:sec:deeplinearkernel}. Concretely, we consider a network defined as
\begin{align}
    \mathbf{h}^{(0)} &= \mathbf{x}
    \\
    \mathbf{h}^{(\ell)} &= \sum_{\ell'=0}^{\ell-1} \frac{\sigma_{\ell,\ell'}}{\sqrt{n_{\ell'}}} W^{(\ell,\ell')} \mathbf{h}^{(\ell')} \qquad \ell = 1,\ldots,d
    \\
    \mathbf{f} &= \mathbf{h}^{(d)},
\end{align}
where $\sigma_{\ell,\ell'}$ is positive if layer $\ell$ receives input from an earlier layer $\ell' < \ell$, and zero otherwise. 

\subsection{Perturbative computation of the partition function}

Upon integrating out the weights, we obtain an effective action for the preactivations and the corresponding Lagrange multipliers of
\begin{align}
    S &= -\beta \sum_{\mu=1}^{p} \varepsilon(\mathbf{h}^{(d)}_{\mu}, \mathbf{y}_{\mu}) + \sum_{\mu=1}^{p} \sum_{\ell=1}^{d} i \mathbf{q}_{\mu}^{(\ell)} \cdot \mathbf{h}_{\mu}^{(\ell)} 
    \nonumber\\&\quad - \frac{1}{2} \sum_{\ell=1}^{d-1} \frac{1}{n_{\ell}} \sum_{\mu,\nu=1}^{p} \bigg[J^{(\ell)} + \sum_{\ell'=\ell+1}^{d} \sigma_{\ell',\ell}^{2}  (\mathbf{q}^{(\ell')}_{\mu} \cdot \mathbf{q}^{(\ell')}_{\nu}) \bigg] (\mathbf{h}^{(\ell)}_{\mu} \cdot \mathbf{h}^{(\ell)}_{\nu})
    \nonumber\\&\quad - \frac{1}{2} \sum_{\ell=1}^{d} \sigma_{\ell,0}^{2} \sum_{\mu,\nu=1}^{p} (G_{xx})_{\mu\nu} (\mathbf{q}^{(\ell)}_{\mu} \cdot \mathbf{q}^{(\ell)}_{\nu}).
\end{align}
Applying the result of Appendix \ref{app:subsec:layerintegral} with
\begin{align}
\begin{split}
    G &= \sigma_{1,0}^{2} G_{xx},
    \\
    \mathbf{j}_{\mu} &= \mathbf{0},
    \\
    A &= J^{(1)} + \sum_{\ell'=2}^{d} \sigma_{\ell',1}^{2}  Q^{(\ell')}, 
    \\
    B &= 0, \quad \textrm{and}
    \\
    g &= 0,
\end{split}
\end{align}
we find that the effective action after integrating out the first layer is
\begin{align}
    S^{(1)} &= -\beta \sum_{\mu=1}^{p} \varepsilon(\mathbf{h}^{(d)}_{\mu}, \mathbf{y}_{\mu}) + \sum_{\mu=1}^{p} \sum_{\ell=2}^{d} i \mathbf{q}_{\mu}^{(\ell)} \cdot \mathbf{h}_{\mu}^{(\ell)} 
    \nonumber\\&\quad - \frac{1}{2} \sum_{\ell=2}^{d-1} \frac{1}{n_{\ell}} \sum_{\mu,\nu=1}^{p} \bigg[J^{(\ell)} + \sum_{\ell'=\ell+1}^{d} \sigma_{\ell',\ell}^{2}  (\mathbf{q}^{(\ell')}_{\mu} \cdot \mathbf{q}^{(\ell')}_{\nu}) \bigg] (\mathbf{h}^{(\ell)}_{\mu} \cdot \mathbf{h}^{(\ell)}_{\nu})
    \nonumber\\&\quad 
    - \frac{1}{2} \sum_{\ell=2}^{d} m_{\ell,1}^{2} \tr( G_{xx} Q^{(\ell)} )
    + \frac{1}{4} \frac{1}{n_1} \sum_{\ell,\ell'=2}^{d} g_{\ell,\ell',1} \tr(G_{xx} Q^{(\ell)} G_{xx} Q^{(\ell')} )
    \nonumber\\&\quad 
    + \frac{1}{2} \frac{1}{n_1} \sum_{\ell=2}^{d} \tr(G_{xx} \tilde{J}_{\ell,1} G_{xx} Q^{(\ell)} )
    \nonumber\\&\quad
    - \frac{1}{2} m_{1,0}^{2} \tr(G_{xx} J^{(1)})
    \nonumber\\&\quad + \frac{1}{4} \frac{1}{n_1} \sigma_{1,0}^{4} \tr(G_{xx} J^{(1)} G_{xx} J^{(1)})
    \nonumber\\&\quad + \mathcal{O}(n^{-2}),
\end{align}
where we have defined
\begin{align}
    m_{\ell,0}^2 &\equiv \sigma_{\ell,0}^{2}, 
    \\
    m_{\ell,1}^{2} &\equiv m_{\ell,0}^{2} + \sigma_{\ell,1}^{2} m_{1,0}^{2}
    \\
    g_{\ell,\ell',1} &\equiv \sigma_{\ell,1}^{2} \sigma_{\ell',1}^{2} \sigma_{1,0}^{4}, \quad \textrm{and}
    \\
    \tilde{J}_{\ell,1} &\equiv \sigma_{\ell,1}^{2} \sigma_{1,0}^{4} J^{(1)},
\end{align}
where $\ell,\ell' > 1$ for all cases but $m_{1,0}^{2}$. Assuming the network has more than one hidden layer, if we now again apply the results of Appendix \ref{app:subsec:layerintegral} with
\begin{align}
\begin{split}
    G &= m_{2,1}^{2} G_{xx},
    \\
    \mathbf{j}_{\mu} &= \mathbf{0},
    \\
    A &= J^{(2)} + \sum_{\ell'=3}^{d} \sigma_{\ell',2}^{2} Q^{(\ell')},
    \\
    B &= G_{xx} \tilde{J}_{2,1} G_{xx} + \sum_{\ell'=3}^{d} g_{\ell',2,1} G_{xx} Q^{(\ell')} G_{xx}, \quad \textrm{and}
    \\
    g &= g_{2,2,1} / m_{2,1}^{4},
\end{split}
\end{align}
we find that the effective action after integrating out the first two layers of the network is
\begin{align}
    S^{(2)} &= -\beta \sum_{\mu=1}^{p} \varepsilon(\mathbf{h}^{(d)}_{\mu}, \mathbf{y}_{\mu}) + \sum_{\mu=1}^{p} \sum_{\ell=1}^{d} i \mathbf{q}_{\mu}^{(\ell)} \cdot \mathbf{h}_{\mu}^{(\ell)} 
    \nonumber\\&\quad - \frac{1}{2} \sum_{\ell=3}^{d-1} \frac{1}{n_{\ell}} \sum_{\mu,\nu=1}^{p} \bigg[J_{\mu\nu}^{(\ell)} + \sum_{\ell'=\ell+1}^{d} \sigma_{\ell',\ell}^{2}  (\mathbf{q}^{(\ell')}_{\mu} \cdot \mathbf{q}^{(\ell')}_{\nu}) \bigg] (\mathbf{h}^{(\ell)}_{\mu} \cdot \mathbf{h}^{(\ell)}_{\nu})
    \nonumber\\&\quad - \frac{1}{2} \sum_{\ell=3}^{d} m_{\ell,2}^{2} \tr(G_{xx} Q^{(\ell)}) + \frac{1}{4} \frac{1}{n_{2}} \sum_{\ell,\ell'=3}^{d} g_{\ell,\ell',2} \tr( G_{xx} Q^{(\ell)} G_{xx}   Q^{(\ell')} )
    \nonumber\\&\quad + \frac{1}{2} \frac{1}{n_{2}} \sum_{\ell=3}^{d} \tr(G_{xx} \tilde{J}_{\ell,2} G_{xx} Q^{(\ell)} )
    \nonumber\\&\quad - \frac{1}{2} \sum_{\ell=1}^{2} m_{\ell,\ell-1}^{2} \tr( G_{xx} J^{(\ell)} )
    \nonumber\\&\quad + \frac{1}{4} \frac{1}{n_{1}} m_{1,0}^{4} \tr( G_{xx} J^{(1)} G_{xx} J^{(1)} ) + \frac{1}{4} \frac{1}{n_2} \left(m_{2,1}^{4} + \frac{n_2}{n_1} g_{2,2,1}\right) \tr(G_{xx} J^{(2)} G_{xx} J^{(2)} ) 
    \nonumber\\&\quad  + \frac{1}{2} \frac{1}{n_{1}} \tr(G_{xx} \tilde{J}_{2,1} G_{xx} J^{(2)})
    \nonumber\\&\quad + \mathcal{O}(n^{-2}),
\end{align}
where we now define
\begin{align}
    m_{\ell,2}^{2} &\equiv m_{\ell,1}^{2} + m_{2,1}^{2} \sigma_{\ell,2}^{2},
    \\
    g_{\ell,\ell',2} &\equiv m_{2,1}^{4} + \frac{n_2}{n_1} \bigg(g_{\ell,\ell',1} + g_{2,2,1} \sigma_{\ell,2}^{2} \sigma_{\ell',2}^{2} + g_{\ell,2,1} \sigma_{\ell',2}^{2} + \sigma_{\ell,2}^{2}  g_{2,\ell',1} \bigg), \quad \textrm{and}
    \\
    \tilde{J}_{\ell,2} &\equiv \frac{n_2}{n_1} \tilde{J}_{\ell,1} + \left(m_{2,1}^{4} + \frac{n_2}{n_1} g_{2,2,1}\right) \sigma_{\ell,2}^{2} J^{(2)} + \frac{n_2}{n_1} \sigma_{\ell,2}^{2} \tilde{J}_{2,1} + \frac{n_2}{n_1} g_{\ell,2,1} J^{(2)}
\end{align}
for $\ell,\ell' > 2$. We can now see that we can repeat this procedure to integrate out all of the hidden layers of the network, yielding an effective action of 
\begin{align}
    S^{(d-1)} &= -\beta \sum_{\mu=1}^{p} \varepsilon(\mathbf{h}^{(d)}_{\mu}, \mathbf{y}_{\mu}) + \sum_{\mu=1}^{p} i \mathbf{q}_{\mu}^{(d)} \cdot \mathbf{h}_{\mu}^{(d)} 
    \nonumber\\&\quad - \frac{1}{2} m_{d,d-1}^{2} \tr(G_{xx} Q^{(d)})  + \frac{1}{4} \frac{1}{n_{d-1}} g_{d,d,d-1} \tr( G_{xx} Q^{(d)} G_{xx}   Q^{(d)} )
    \nonumber\\&\quad + \frac{1}{2} \frac{1}{n_{d-1}} \tr(G_{xx} \tilde{J}_{d,d-1} G_{xx} Q^{(d)} )
    \nonumber\\&\quad - \frac{1}{2} \sum_{\tau=1}^{d-1} m_{\tau,\tau-1}^{2} \tr(G_{xx} J^{(\tau)}) 
    \nonumber\\&\quad + \frac{1}{4} \sum_{\tau=2}^{d-1} \left( \frac{1}{n_{\tau}} m_{\tau,\tau-1}^{4} + \frac{1}{n_{\tau-1}} g_{\tau,\tau,\tau-1}\right) \tr(G_{xx} J^{(\tau)} G_{xx} J^{(\tau)} ) 
    \nonumber\\&\quad  + \frac{1}{2} \sum_{\tau=2}^{d-1} \frac{1}{n_{\tau-1}} \tr(G_{xx} \tilde{J}_{\tau,\tau-1} G_{xx} J^{(\tau)})
    \nonumber\\&\quad + \mathcal{O}(n^{-2}),
\end{align}
where the coupling constants and effective source obey the recurrences
\begin{align}
    m_{\ell,\tau}^{2} &\equiv m_{\ell,\tau-1}^{2} + m_{\tau,\tau-1}^{2} \sigma_{\ell,\tau}^{2}, 
    \\
    g_{\ell,\ell',\tau} &\equiv m_{\tau,\tau-1}^{4} \sigma_{\ell,\tau}^{2} \sigma_{\ell',\tau}^{2} \nonumber\\&\quad + \frac{n_{\tau}}{n_{\tau-1}} \bigg(g_{\ell,\ell',\tau-1} + g_{\tau,\tau,\tau-1} \sigma_{\ell,\tau}^{2} \sigma_{\ell',\tau}^{2} + g_{\ell,\tau,\tau-1} \sigma_{\ell',\tau}^{2} + \sigma_{\ell,\tau}^{2}  g_{\tau,\ell',\tau-1} \bigg), \quad \textrm{and}
    \\
    \tilde{J}_{\ell,\tau} &\equiv \frac{n_{\tau}}{n_{\tau-1}} \tilde{J}_{\ell,\tau-1} + \frac{n_{\tau}}{n_{\tau-1}} \sigma_{\ell,\tau}^{2} \tilde{J}_{\tau,\tau-1} \nonumber\\&\quad + \left(m_{\tau,\tau-1}^{4} \sigma_{\ell,\tau}^{2} + \frac{n_{\tau}}{n_{\tau-1}} g_{\tau,\tau,\tau-1} \sigma_{\ell,\tau}^{2} + \frac{n_{\tau}}{n_{\tau-1}} g_{\ell,\tau,\tau-1} \right) J^{(\tau)}
\end{align}
for $\ell,\ell' > \tau$. Applying the results of Appendix \ref{app:subsec:layerintegral} once more with
\begin{align}
\begin{split}
    G &= m_{d,d-1}^{2} G_{xx},
    \\
    \mathbf{j}_{\mu} &= \beta \mathbf{y}_{\mu},
    \\
    A &= \beta n_{d} I_{p},
    \\
    B &= G_{xx} \tilde{J}_{d,d-1} G_{xx}, \quad \textrm{and}
    \\
    g &= g_{d,d,d-1}/m_{d,d-1}^{4},
\end{split}
\end{align}
we find the source-dependent terms in the logarithm of the partition function are
\begin{align}
    \log Z 
    &\supset  - \frac{1}{2} \frac{1}{n_{1}} \beta^2 n_{d} \tr(\Gamma^{-1} G_{xx} \tilde{J}_{d,d-1} G_{xx} \Gamma^{-1} G_{yy}) + \frac{1}{2} \frac{1}{n_{d-1}} \beta n_{d} \tr(\Gamma^{-1} G_{xx} \tilde{J}_{d,d-1} G_{xx})
    \nonumber\\&\quad - \frac{1}{2} \sum_{\tau=1}^{d-1} m_{\tau,\tau-1}^{2} \tr(G_{xx} J^{(\tau)}) 
    \nonumber\\&\quad + \frac{1}{4} \sum_{\tau=2}^{d-1} \left( \frac{1}{n_{\tau}} m_{\tau,\tau-1}^{4} + \frac{1}{n_{\tau-1}} g_{\tau,\tau,\tau-1}\right) \tr(G_{xx} J^{(\tau)} G_{xx} J^{(\tau)} ) 
    \nonumber\\&\quad  + \frac{1}{2} \sum_{\tau=2}^{d-1} \frac{1}{n_{\tau-1}} \tr(G_{xx} \tilde{J}_{\tau,\tau-1} G_{xx} J^{(\tau)})
    \nonumber\\&\quad + \mathcal{O}(n^{-2}),
\end{align}
where
\begin{align}
    \Gamma \equiv I_{p} + \beta m_{d,d-1}^{2} G_{xx}.
\end{align}

\subsection{Computing the average hidden layer kernels}

With the source-dependent terms of the relevant partition function in hand, we can compute the average hidden layer kernels for a feedforward linear network with arbitrary skip connections. We can immediately read off that
\begin{align}
    \langle K^{(\ell)} \rangle &= m_{\ell,\ell-1}^{2} G_{xx} \nonumber\\&\quad + \frac{n_{d}}{n_{d-1}}  \tr\left[\bigg(\beta^{2} G_{xx} \Gamma^{-1} G_{yy} \Gamma^{-1} G_{xx} - \beta G_{xx} \Gamma^{-1} G_{xx}\bigg)  \frac{\delta \tilde{J}_{d,d-1}}{\delta J^{(\ell)}} \bigg|_{J^{(\ell)} = 0}\right]  \nonumber\\&\quad + \mathcal{O}(n^{-2}),
\end{align}
hence our only task is to compute the derivative of the effective source $\tilde{J}_{d,d-1}$ with respect to the source for the $\ell$-th hidden layer. Singling out the $\ell$-th layer, we can set all sources except $J^{(\ell)}$ to zero. Then, the `earliest' effective source to be non-zero is
\begin{align}
    \tilde{J}_{\ell',\ell}
    = \left(m_{\ell,\ell-1}^{4} \sigma_{\ell',\ell}^{2} + \frac{n_{\ell}}{n_{\ell-1}} g_{\ell,\ell,\ell-1} \sigma_{\ell',\ell}^{2} + \frac{n_{\ell}}{n_{\ell-1}} g_{\ell',\ell,\ell-1}\right) J^{(\ell)},
\end{align}
for $\ell' > \ell$, and the recurrence relation for $\tau > \ell$ is
\begin{align}
    \tilde{J}_{\ell',\tau} = \frac{n_{\tau}}{n_{\tau-1}} \bigg( \tilde{J}_{\ell',\tau-1} + \sigma_{\ell',\tau}^{2} \tilde{J}_{\tau,\tau-1} \bigg).
\end{align}
From the form of these recurrences, we can see that
\begin{align}
    \langle K^{(\ell)} \rangle &= m_{\ell,\ell-1}^{2} G_{xx} \nonumber\\&\quad + \frac{n_{d}}{n_{d-1}} \tilde{g}_{\ell} G_{xx} (\beta^{2} \Gamma^{-1} G_{yy} \Gamma^{-1} - \beta \Gamma^{-1} ) G_{xx}  \nonumber\\&\quad + \mathcal{O}(n^{-2}),
\end{align}
where $\tilde{g}_{\ell}$ is a layer-dependent scalar. Even without explicitly solving the recurrences to obtain $\tilde{g}_{\ell}$, this shows that Conjecture \ref{conj1} holds perturbatively for linear networks with arbitrary skip connections. We leave detailed study of these recurrences---and therefore of the precise dependence of the corrections on width, depth, and skip connection structure---as an interesting objective for future work.

\section[Comparison to the results of Aitchison (2020) and Li \& Sompolinsky (2020)]{Comparison to the results of \citet{aitchison2020bigger} and \citet{li2021statistical}}\label{app:sec:comp}

In this appendix, we compare our results for the average kernels of deep linear networks to those of \citet{aitchison2020bigger} and \citet{li2021statistical}. 

\subsection[Comparison to the results of Aitchison (2020)]{Comparison to the results of \citet{aitchison2020bigger}}

We first show that our result \eqref{eqn:lowtemplinear} for the low-temperature limit of the average kernels of a deep linear network can be recovered from the results of \citet{aitchison2020bigger}. Working in what corresponds to the zero-temperature limit of our setup, Aitchison derives the following implicit recurrence
\begin{align}
    0 = - (n_{\ell+1}-n_{\ell}) (K^{(\ell)})^{-1} + n_{\ell+1} (K^{(\ell)})^{-1} (K^{(\ell+1)}) (K^{(\ell)})^{-1} - n_{\ell} (K^{(\ell-1)})^{-1},
\end{align}
for $\ell=1,\ldots,d-1$, where the boundary conditions of the recurrence are $K^{(0)} = G_{xx}$ and $K^{(d)}=G_{yy}$. We will self-consistently solve this recurrence relation in the limit $n_{1},\ldots,n_{d-1}\to\infty$, $n_{0},n_{d},p=\mathcal{O}(1)$. Concretely, we make the ansatz that the zero-temperature kernels are of the form
\begin{align}
    K^{(\ell)} = K^{(\ell)}_{\infty} + \frac{1}{n_{\ell}} K^{(\ell)}_{1} + \mathcal{O}(n_{\ell}^{-2}),
\end{align}
and solve the recurrence relations order-by-order using the resulting Neumann series
\begin{align}
    (K^{(\ell)})^{-1} &= (K^{(\ell)}_{\infty})^{-1} - \frac{1}{n_{\ell}} (K^{(\ell)}_{\infty})^{-1} K^{(\ell)}_{1} (K^{(\ell)}_{\infty})^{-1} + \mathcal{O}(n_{\ell}^{-2}).
\end{align}
The leading-order recurrence is simply
\begin{align}
    0 
    &= \left(1 - \frac{n_{\ell+1}}{n_{\ell}}\right) (K^{(\ell)}_{\infty})^{-1} + \frac{n_{\ell+1}}{n_{\ell}} (K^{(\ell)}_{\infty})^{-1} (K^{(\ell+1)}_{\infty}) (K^{(\ell)}_{\infty})^{-1} - (K^{(\ell-1)}_{\infty})^{-1},
\end{align}
with boundary conditions $K^{(0)}_{\infty} = G_{xx}$ and $K^{(d)}_{\infty} = G_{yy}$. For the last hidden layer, we have $n_{\ell+1}/n_{\ell} = n_{d}/n_{d-1} \to 0$, hence the recurrence reduces to
\begin{align}
    K^{(d-1)}_{\infty} = K^{(d-2)}_{\infty}. 
\end{align}
If we iterate this procedure backwards through the network, it is easy to see that the $n_{\ell+1}/n_{\ell}$-dependent terms at each layer will cancel, leaving
\begin{align}
    K^{(d-1)}_{\infty} = K^{(d-2)}_{\infty} = \cdots = K^{(1)}_{\infty} = G_{xx}.
\end{align}
We now consider the leading finite-width correction. For the last hidden layer, we obtain 
\begin{align}
    0 = n_{d} (G_{yy}-G_{xx}) - K^{(d-1)}_{1} + \frac{n_{d-1}}{n_{d-2}} K^{(d-2)}_{1}
\end{align}
after dropping all terms that are of $\mathcal{O}(n^{-2})$ and multiplying on the left and right by $G_{xx}$. For the first hidden layer, we have
\begin{align}
    0 = K^{(2)}_{1} - \left(1 + \frac{n_{2}}{n_{1}}\right) K^{(1)}_{1}.
\end{align}
Finally, for intermediate hidden layers (i.e., $\ell=2,3,\ldots,d-2$), we have
\begin{align}
    0 = K^{(\ell+1)}_{1} - \left(1 + \frac{n_{\ell+1}}{n_{\ell}}\right) K^{(\ell)}_{1} + \frac{n_{\ell}}{n_{\ell-1}} K^{(\ell-1)}_{1}.
\end{align}
Based on the form of these recurrences, we make the ansatz that the solution is of the form
\begin{align}
    K_{1}^{(\ell)} = n_{d} a_{\ell} (G_{yy}-G_{xx})
\end{align}
for some sequence $a_{\ell}$, where we assume that $G_{yy} \neq G_{xx}$. Then, the recurrence for the last hidden layer is satisfied provided that
\begin{align}
    a_{d-1} = 1 + \frac{n_{d-1}}{n_{d-2}} a_{d-2},
\end{align}
those for the intermediate layers if
\begin{align}
    0 = a_{\ell+1} - \left(1 + \frac{n_{\ell+1}}{n_{\ell}}\right) a_{\ell} + \frac{n_{\ell}}{n_{\ell-1}} a_{\ell-1},
\end{align}
and that for the first hidden layer if 
\begin{align}
    a_{2} = \left(1 + \frac{n_{2}}{n_{1}}\right) a_{1}.
\end{align}
Substituting the expression for $a_{d-1}$ into the condition resulting from the recurrence relation centered on $a_{d-2}$, we find that we must have
\begin{align}
    a_{d-2} = 1 + \frac{n_{d-2}}{n_{d-3}} a_{d-3},
\end{align}
hence we can iterate this process backwards to the second hidden layer, yielding
\begin{align}
    a_{\ell} = 1 + \frac{n_{\ell}}{n_{\ell-1}} a_{\ell-1}
\end{align}
for $\ell=2,3,\ldots,d-1$. Then, the condition relating $a_{2}$ and $a_{1}$ resulting from the recurrence relation for the first layer implies that we must have $a_{1} = 1$. Thus, we recover our zero-temperature result from solving Aitchison's recurrence relations order-by-order. 

\subsection[Comparison to the results of Li and Sompolinsky (2020)]{Comparison to the results of \citet{li2021statistical}}

We now show that our result \eqref{eqn:lowtemplinear} for the low-temperature limit of the average kernels of a deep linear network can be recovered as a limiting case of the result of \citet{li2021statistical}. Their result for the zero-temperature kernel in the limit $n_{0}, n, p \to \infty$ with $n_{1} = n_{2} = \cdots = n_{d-1} = n$, $n_{0}/n \in (0,\infty)$, $\alpha \equiv p/n \in (0,\infty)$, and $\sigma_{1} = \cdots = \sigma_{d} = \sigma$ is, in our notation,
\begin{align}
    \sigma^{-2(\ell+1)} \langle K^{(\ell)} \rangle \sim  \left(1 - \frac{n_d}{n}\right)^{\ell} G_{xx} + \frac{1}{n} \sigma^{-2d} Y V M_{\ell} V^{\top} Y^{\top},
\end{align}
where $Y \in \mathbb{R}^{p \times n_{d}}$ is the matrix of targets and $M_{\ell} \in \mathbb{R}^{n_{d} \times n_{d}}$ is a diagonal matrix with non-zero elements 
\begin{align}
    [M_{\ell}]_{kk} = z_{k}^{-(d-1)} \frac{z_{k}^{\ell} - 1}{z_{k}-1}.
\end{align}
Here, the orthogonal matrix $V$ is the matrix of eigenvectors of
\begin{align}
    R = \frac{1}{\sigma^{2} p} Y^{\top} G_{xx}^{+} Y = V \Omega V^{\top},
\end{align}
for $G_{xx}^{+}$ the pseudoinverse of $G_{xx}$, and the scalars $z_{k}$ are in turn defined in terms of the eigenvalues $\Omega_{kk} = \omega_{k}$ as
\begin{align}
    1-\alpha = z_{k} - \alpha \sigma^{-2(d-1)} z_{k}^{-(d-1)} \omega_{k};
\end{align}
we note that \citet{li2021statistical} use variables $u_{k0} = \sigma^2 z_{k}$. 

As we are interested in the limit $\alpha \downarrow 0$, it is useful to write the implicit equation for $z_{k}$ as
\begin{align}
    z_{k} = 1 + \alpha (\sigma^{-2L} z_{k}^{-(d-1)} \omega_{k} - 1),
\end{align}
hence we expect $z_{k} \to 1$ as $\alpha \downarrow 0$. Thus, we have 
\begin{align}
    [M_{\ell}]_{kk} \to \ell,
\end{align}
which gives
\begin{align}
    V M_{\ell} V^{\top} \to \ell I_{n_d}.
\end{align}
Using the expansion $(1-n_d/n)^{\ell} = 1 - n_{d}\ell/n + \mathcal{O}(n^{-2})$, we therefore find that
\begin{align}
    \sigma^{-2(\ell+1)} \langle K^{(\ell)} \rangle \sim G_{xx} + \frac{n_{d} \ell}{n} (\sigma^{-2 d} G_{yy} - G_{xx})
\end{align}
in the limit in which $n_{d}/n \downarrow 0$ and $p/n \downarrow 0$. Therefore, combining this result with that of the previous subsection, our result \eqref{eqn:lowtemplinear} agrees with those of \citet{aitchison2020bigger} and of \citet{li2021statistical} in the appropriate limit. Whether the full result of \citet{li2021statistical} agrees with that of \citet{aitchison2020bigger} is an interesting question, but is well beyond the scope of the present work.

\section{Predictor statistics and generalization in deep linear networks}\label{app:sec:predictor}

Though the main focus of our work is on the asymptotics of representation learning, we have also computed the leading finite-width corrections to the predictor statistics. Though one can derive the analogy of Conjecture \ref{conj1} for the predictor statistics of a general BNN with linear readout, the resulting formula is not particularly illuminating. We will therefore present results only for linear networks. As was true of the hidden layer kernels of deep linear networks, this calculation can be performed either using methods similar to those described in Appendix \ref{app:sec:perturbation} or Appendix \ref{app:sec:deeplinearkernel}. As the steps are largely identical to those calculations, we only briefly summarize the results. 

In short, we fix a test dataset $\hat{\mathcal{D}} = \{(\hat{\mathbf{x}}_{\mu},\hat{\mathbf{y}}_{\mu})\}_{\mu=1}^{\hat{p}}$ of $\hat{p}$ examples, and define the Gram matrices
\begin{align}
    (G_{\hat{x}\hat{x}})_{\hat{\mu}\hat{\nu}} &\equiv n_{0}^{-1} \hat{\mathbf{x}}_{\hat{\mu}} \cdot \hat{\mathbf{x}}_{\hat{\nu}},
    \\
    (G_{\hat{y}\hat{y}})_{\hat{\mu}\hat{\nu}} &\equiv n_{d}^{-1} \hat{\mathbf{y}}_{\hat{\mu}} \cdot \hat{\mathbf{y}}_{\hat{\nu}},
    \\
    (G_{x\hat{x}})_{\mu\hat{\mu}} &\equiv n_{0}^{-1} \mathbf{x}_{\mu} \cdot \hat{\mathbf{x}}_{\hat{\mu}}, \quad \textrm{and}
    \\
    (G_{y\hat{y}})_{\mu\hat{\nu}} &\equiv n_{d}^{-1} \mathbf{y}_{\mu} \cdot \hat{\mathbf{y}}_{\hat{\nu}}.
\end{align}
Introducing appropriate source terms to allow us to compute predictor statistics, we then proceed perturbatively as before, assuming that the combined input Gram matrix
\begin{align}
    \begin{bmatrix} G_{xx} & G_{x\hat{x}} \\ G_{x\hat{x}}^{\top} & G_{\hat{x}\hat{x}} \end{bmatrix}
\end{align}
is invertible. Again, the final result can be extended to the case in which this matrix is not invertible by a continuity argument. 

Our notation in this appendix will follow that of Appendix \ref{app:sec:perturbation} rather than Appendix \ref{app:sec:deeplinearkernel} in that we will introduce matrices
\begin{align}
    K_{\infty} &\equiv \sigma_{1}^{2} \cdots \sigma_{d-1}^{2} G_{xx},
    \\
    \hat{R}_{\infty} &\equiv  \sigma_{1}^{2} \cdots \sigma_{d-1}^{2} G_{x\hat{x}}, \quad \textrm{and} 
    \\
    \hat{K}_{\infty} &\equiv \sigma_{1}^{2} \cdots \sigma_{d-1}^{2} G_{\hat{x}\hat{x}}
\end{align}
to denote the blocks of the infinite-width kernel of the last hidden layer, rather than introducing scalar parameters to represent the products of variances. This will make our expressions somewhat more compact than they would be under the conventions of Appendix \ref{app:sec:deeplinearkernel}. 

\subsection{Predictor statistics}

Defining the matrix $\hat{F}_{\hat{\mu} j} \equiv f_{j}(\hat{\mathbf{x}}_{\hat{\mu}})$, we find that the mean predictor can be written compactly as
\begin{align}
    \langle \hat{F} \rangle = \hat{R}_{\infty}^{\top} \left[\Gamma^{-1} - \frac{1}{\beta \sigma_{d}^{2}} \left(\sum_{\ell=1}^{d-1} \frac{1}{n_{\ell}}\right) \Gamma^{-1} M \Gamma^{-1} \right] Y + \mathcal{O}(n^{-2})
\end{align}
for
\begin{align}
    M \equiv \Gamma^{-1} K_{\infty} + \tr(\Gamma^{-1} K_{\infty}) I_{p} - n_{d}(\sigma_{d}^{-2} \Gamma^{-1}  G_{yy} \Gamma^{-1} - \Gamma^{-1})  K_{\infty} .
\end{align}
The predictor covariance is given as
\begin{align}
    &\sigma_{d}^{-2} \cov(\hat{F}_{\hat{\mu}j},\hat{F}_{\hat{\nu}k})
    \nonumber\\&\quad= (\hat{K}_{\infty} - \hat{R}_{\infty}^{\top} \Gamma^{-1} \hat{R}_{\infty})_{\hat{\mu}\hat{\nu}} \delta_{jk} 
    \nonumber\\&\qquad + \left(\sum_{\ell=1}^{d-1} \frac{1}{n_{\ell}}\right) \bigg[ \hat{M}_{\hat{\mu}\hat{\nu}} \delta_{jk}
    \nonumber\\&\qquad\qquad\qquad\qquad + \sigma_{d}^{-2} (Y^{\top} \Gamma^{-1} K_{\infty} \Gamma^{-1} Y)_{j k}  (\hat{K}_{\infty} - \hat{R}_{\infty}^{\top} \Gamma^{-1} \hat{R}_{\infty} )_{\hat{\mu}\hat{\nu}} 
     \nonumber\\&\qquad\qquad\qquad\qquad - \frac{1}{\beta \sigma_{d}^{4}} (Y^{\top} \Gamma^{-1} K_{\infty} \Gamma^{-1} Y)_{j k} (  \hat{R}_{\infty}^{\top} \Gamma^{-2} \hat{R}_{\infty} )_{\hat{\mu}\hat{\nu}} 
    \nonumber\\&\qquad\qquad\qquad\qquad + \frac{1}{\beta^{2} \sigma_{d}^{6}} (Y^{\top} \Gamma^{-2} \hat{R}_{\infty})_{j \hat{\nu}} (Y^{\top} \Gamma^{-2} \hat{R}_{\infty})_{k \hat{\mu}} \bigg]
    \nonumber\\&\qquad + \mathcal{O}(n^{-2})
\end{align}
for
\begin{align}
    \hat{M} &\equiv - \tr(\Gamma^{-1} K_{\infty}) (\hat{K}_{\infty} - \hat{R}_{\infty}^{\top}  \Gamma^{-1} \hat{R}_{\infty}) + \frac{1}{\beta \sigma_{d}^{2}} \tr(\Gamma^{-1} K_{\infty}) \hat{R}_{\infty}^{\top} \Gamma^{-2} \hat{R}_{\infty} - \frac{1}{\beta^2 \sigma_{d}^{4}} \hat{R}_{\infty}^{\top} \Gamma^{-3} \hat{R}_{\infty} \nonumber\\&\quad + n_{d}  \frac{1}{\beta^{2} \sigma_{d}^{4}} \hat{R}_{\infty}^{\top} \Gamma^{-1} (\sigma_{d}^{-2} \Gamma^{-1}  G_{yy} \Gamma^{-1} - \Gamma^{-1}) \Gamma^{-1} \hat{R}_{\infty} .
\end{align}
The mean and covariance of the training set predictor $F_{\mu j} \equiv f_{j}(\mathbf{x}_{\mu})$ can be obtained by setting $\hat{R}_{\infty}$ and $\hat{K}_{\infty}$ to $K_{\infty}$ in the above expressions. 

\subsection{Bias-variance decompositions and the low-temperature limit}

These results allow us to define thermal bias-variance decompositions of the form
\begin{align}
    \langle E \rangle &= \frac{1}{2} \sum_{\mu=1}^{p} \Vert \langle \mathbf{f}(\mathbf{x}_{\mu}) \rangle - \mathbf{y}_{\mu} \Vert_{2}^{2}  + \frac{1}{2} \sum_{\mu=1}^{p} \sum_{k=1}^{n_d} \cov[f_{k}(\mathbf{x}_{\mu}),f_{k}(\mathbf{x}_{\mu})] 
    \equiv E_{b} + E_{v}
\end{align}
for the mean training and test errors. However, the resulting expressions are not particularly illuminating except in the low-temperature limit $\beta \to \infty$. We will focus on the regime in which $G_{xx}$ (and thus $K_{\infty}$) is invertible, in which the underlying linear system $X W = Y$ is underdetermined and the training set can be interpolated. In this regime, $\Gamma^{-1} = K_{\infty}^{-1} + \mathcal{O}(\beta^{-1})$, and the mean predictor reduces to the least-norm pseudoinverse solution to the linear system, with mean training and test predictions of 
\begin{align}
    \langle F \rangle = Y + \mathcal{O}(\beta^{-1})
\end{align}
and
\begin{align}
    \langle \hat{F} \rangle = \hat{R}_{\infty}^{\top} K_{\infty}^{-1} Y + \mathcal{O}(\beta^{-1}) = G_{\hat{x} x}^{\top} G_{xx}^{-1} Y + \mathcal{O}(\beta^{-1}) = \hat{X} X^{\top} (X X^{\top})^{-1} Y + \mathcal{O}(\beta^{-1}),
\end{align}
respectively. The training and test set covariances have low-temperature limits of
\begin{align}
    \cov(F_{\mu j}, F_{\nu k}) = \mathcal{O}(\beta^{-1})
\end{align}
and
\begin{align}
    \cov(\hat{F}_{\hat{\mu}j},\hat{F}_{\hat{\nu}k}) &= \sigma_{d}^{2} (\hat{K}_{\infty} - \hat{R}_{\infty}^{\top} K_{\infty}^{-1} \hat{R}_{\infty})_{\hat{\mu}\hat{\nu}} \left[ \delta_{jk} + \left(\sum_{\ell=1}^{d-1} \frac{1}{n_{\ell}}\right) (\sigma_{d}^{-2} Y^{\top} K_{\infty}^{-1} Y - p I_{n_2})_{j k} \right]
    \nonumber\\&\qquad + \mathcal{O}(\beta^{-1},n^{-2}),
\end{align}
respectively. Then, it is easy to see that both $E_{b}$ and $E_{v}$ are $\mathcal{O}(\beta^{-1})$, while
\begin{align}
    \hat{E}_{b} = \frac{1}{2} \Vert \hat{R}_{\infty}^{\top} K_{\infty}^{-1} Y - \hat{Y} \Vert_{F}^{2}  + \mathcal{O}(\beta^{-1})
\end{align}
and
\begin{align}
    \hat{E}_{v} &= \frac{1}{2} n_{d} \sigma_{d}^{2} \tr(\hat{K}_{\infty} - \hat{R}_{\infty}^{\top} K_{\infty}^{-1} \hat{R}_{\infty}) \left[ 1 + \left(\sum_{\ell=1}^{d-1} \frac{1}{n_{\ell}}\right) (\sigma_{d}^{-2} \tr( K_{\infty}^{-1} G_{yy}) - p ) \right]
    \nonumber\\&\qquad + \mathcal{O}(\beta^{-1},n^{-2}) .
\end{align}
Thus, at least to leading order, width affects the low-temperature test error only through the variance term. Substituting in the definition of $K_{\infty}$, we find that to leading order the test error decreases with increasing width if
\begin{align}
    \frac{1}{p} \tr(G_{xx}^{-1} G_{yy}) > \sigma_{1}^{2} \cdots \sigma_{d}^{2}
\end{align}
and increases with increasing width otherwise. This small-initialization condition is the generalization of that found by \citet{li2021statistical} to our asymptotic regime.

\subsection{Effects of alternative regularization temperature-dependence} 

In this appendix, we comment on the possibility of alternative temperature-dependent posteriors. This possibility arises from the interpretation of the Bayes posterior \eqref{eqn:posterior} as the equilibrium distribution of the Langevin dynamics 
\begin{align}\label{eq:langevin_sampling}
    d\Theta^{(\ell)}(t) = - (\lambda(\beta) \Sigma \Theta + \nabla_{\Theta} E) dt + \sqrt{2 \beta^{-1}} dB^{(\ell)}(t)
\end{align}
at inverse temperature $\beta$, where $B^{(\ell)}(t)$ is a standard Wiener process, $\Sigma$ is the diagonal matrix of prior variances, and $\lambda(\beta) = 1/\beta$. As elsewhere, we focus on the regime in which the training dataset can be linearly interpolated, in which the thermal variance of the test set predictions need not vanish. Moreover, it suffices to consider only the GP contributions; the finite-width corrections computed above do not change the qualitative results. In these statistics, the case of general $\lambda(\beta)$ is related to $\lambda(\beta) = 1/\beta$ by the replacement
\begin{align}
    \sigma_{1}^{2} \cdots \sigma_{d}^{2} \gets \frac{\sigma_{1}^{2} \cdots \sigma_{d}^{2}}{\beta^{d} \lambda(\beta)^{d}}.
\end{align}
Then, if we assume a low-temperature power-law dependence $\lambda(\beta) \sim \beta^{\omega}$ for simplicity, we find that the zero-temperature limits of the training set predictor mean and covariance are 
\begin{align}
    \lim_{\beta \to \infty} \langle F \rangle = 
    \begin{cases}
        0 & \omega > 1/d - 1
        \\
         K_{\infty} (\sigma_{d}^{-2} I_{p} + K_{\infty})^{-1} Y & \omega = 1/d - 1
        \\
        Y & \omega < 1/d - 1
    \end{cases}
\end{align}
and
\begin{align}
    \lim_{\beta \to \infty} \cov(F_{\mu j}, F_{\nu k}) = 0,
\end{align}
respectively, while those of the test set mean and covariance are 
\begin{align}
    \lim_{\beta \to \infty} \langle \hat{F} \rangle = 
    \begin{cases}
        0 & \omega > 1/d - 1
        \\
        \hat{R}_{\infty}^{\top} (\sigma_{d}^{-2} I_{p} + K_{\infty})^{-1} Y & \omega = 1/d - 1
        \\
        \hat{R}_{\infty}^{\top} K_{\infty}^{-1} Y & \omega < 1/d - 1
    \end{cases}
\end{align}
and 
\begin{align}
    \lim_{\beta \to \infty} \cov(\hat{F}_{\hat{\mu} j},\hat{F}_{\hat{\nu} k}) = 
    \begin{cases}
        0 & \omega > -1
        \\
        \sigma_{d}^{2} (\hat{K}_{\infty} - \hat{R}^{\top} K_{\infty}^{-1} \hat{R}_{\infty} )_{\hat{\mu}\hat{\nu}} \delta_{jk} & \omega = -1
        \\
        \infty & \omega < -1,
    \end{cases}
\end{align}
respectively. Therefore, taking $\lambda(\beta) = 1/\beta$ yields sensible zero-temperature infinite-width behavior for a linear network of any depth in the underdetermined regime.

\section{Derivation of the average kernels for a depth-two network}\label{app:sec:nonlinkernel}

In this appendix, we derive the average feature kernel for a network with a single (possibly nonlinear) hidden layer and a linear readout. This derivation is a simple extension of the perturbative derivation of Conjecture \ref{conj1} in Appendix \ref{app:sec:perturbation}, using the fact that the size of the terms in the expansion for two-layer networks can be directly controlled in terms of the inverse hidden layer width. 

Concretely, we consider a network defined as
\begin{align}
    \mathbf{h}^{(1)} &= \frac{\sigma_1}{\sqrt{n_0}} W^{(1)} \mathbf{x}
    \\
    \mathbf{h}^{(2)} &= \frac{\sigma_2}{\sqrt{n_1}} W^{(2)} \phi(\mathbf{h}^{(1)})
    \\
    \mathbf{f} &= \mathbf{h}^{(2)}.
\end{align}
Our task is to control the prior cumulants of the hidden layer feature kernel
\begin{align}
    K_{\mu\nu} \equiv \frac{1}{n_{1}} \phi(\mathbf{h}^{(1)}_{\mu}) \cdot \phi(\mathbf{h}^{(1)}_{\nu}).
\end{align}
We can use the fact that the rows $[\mathbf{w}^{(1)}_{j}]^{\top}$ of $W^{(1)}$ are independent and identically distributed under the prior to obtain
\begin{align}
    [K_{\infty}]_{\mu\nu} 
    &= \mathbb{E}_{\mathcal{W}} K_{\mu\nu} 
    \\
    &= \frac{1}{n_1} \sum_{j=1}^{n_1} \mathbb{E}_{\mathbf{w}^{(1)}_{j}}\left[ \phi\left(\frac{\sigma_1}{\sqrt{n_0}} \mathbf{w}^{(1)}_{j} \cdot \mathbf{x}_{\mu}\right) \phi\left(\frac{\sigma_1}{\sqrt{n_0}} \mathbf{w}^{(1)}_{j} \cdot \mathbf{x}_{\nu}\right)\right]
    \\
    &= \mathbb{E}[ \phi(h_{\mu}^{(1)}) \phi(h_{\nu}^{(1)}) \,:\,\mathbf{h}^{(1)} \sim \mathcal{N}(\mathbf{0}, \sigma_{1}^{2} G_{xx})]
\end{align}
at any hidden layer width \cite{neal1996priors,williams1997computing}. Similarly, we can easily see that
\begin{align}
    \cov_{\mathcal{W}}(K_{\mu\nu},K_{\rho\lambda}) = \frac{1}{n_1} \left( \mathbb{E}[ \phi(h_{\mu}^{(1)}) \phi(h_{\nu}^{(1)}) \phi(h_{\rho}^{(1)}) \phi(h_{\lambda}^{(1)})] - [K_{\infty}]_{\mu\nu} [K_{\infty}]_{\rho\lambda}  \right),
\end{align}
where $\mathbf{h}^{(1)} \sim \mathcal{N}(\mathbf{0}, \sigma_{1}^{2} G_{xx})$, and that higher cumulants are $\mathcal{O}(n_{1}^{-2})$. Then, we can directly apply the result of Appendix \ref{app:sec:perturbation} to conclude that
\begin{align}
    \langle K_{\mu\nu} \rangle = [K_{\infty}]_{\mu\nu} + \frac{1}{2} n_{d} \sum_{\rho,\lambda=1}^{p} (\sigma_{d}^{-2} \Gamma^{-1} G_{yy} \Gamma^{-1} - \Gamma^{-1})_{\rho\lambda} \cov_{\mathcal{W}}(K_{\mu\nu},K_{\rho\lambda}) + \mathcal{O}(n_{1}^{-2})
\end{align}
for $\Gamma = \sigma_{1}^{2} K_{\infty} + I_{p}/\beta \sigma_{2}^{2}$. Depending on the nonlinearity, this result may be continuous in $G_{xx}$, and therefore extensible to the non-invertible case via a continuity argument. In particular, as noted in Appendix \ref{app:sec:deeplinearkernel}, this holds for a linear network.

To gain some intuition for how different choices of nonlinear activation function affect the learned representations, we consider the case in which $G_{xx}$ is diagonal. In this special case, the four-point term simplifies dramatically. In particular, we have
\begin{align}
    (K_{\infty})_{\mu\nu} = \var[\phi(h_{\mu}^{(1)})] \delta_{\mu\nu} + \mathbb{E}[\phi(h_{\mu}^{(1)})] \mathbb{E}[ \phi(h_{\nu}^{(1)}) ] 
\end{align}
and
\begin{align}
    \cov_{\mathcal{W}}(K_{\mu\nu},K_{\rho\lambda}) = \frac{1}{n_1} \bigg(& \var[\phi(h_{\mu}^{(1)})^2] \delta_{\mu\nu} \delta_{\mu\rho} \delta_{\mu\lambda} \nonumber\\& + \var[\phi(h_{\mu}^{(1)})] \var[\phi(h_{\nu}^{(1)})] (1-\delta_{\mu\nu}) (\delta_{\mu\rho} \delta_{\nu\lambda} + \delta_{\mu\lambda} \delta_{\nu\rho}) \bigg),
\end{align}
which yields
\begin{align}
    \langle K_{\mu\nu} \rangle 
    &= (K_{\infty})_{\mu\nu} + \frac{1}{2} \frac{n_{2}}{n_1}  (\sigma_{2}^{-2} \Gamma^{-1} G_{yy} \Gamma^{-1} - \Gamma^{-1})_{\mu\nu} \nonumber\\&\qquad\qquad\qquad\times  \bigg[\var[\phi(h_{\mu}^{(1)})^2] \delta_{\mu\nu} + 2 \var[\phi(h_{\mu}^{(1)})] \var[\phi(h_{\nu}^{(1)})] (1-\delta_{\mu\nu})  \bigg] + \mathcal{O}(n_{1}^{-2}).
\end{align}
Moreover, applying the Sherman-Morrison formula \cite{horn2012matrix}, we have
\begin{align}
    \frac{1}{\beta \sigma_{2}^{2}} \Gamma^{-1}_{\mu\nu} = \frac{\delta_{\mu\nu} }{\gamma_{\mu}} - \frac{1}{1 + \sum_{\rho=1}^{p} \mathbb{E}[\phi(h_{\rho}^{(1)})]^2/\gamma_{\rho}} \frac{\mathbb{E}[\phi(h_{\mu}^{(1)})]}{\gamma_{\mu}} \frac{\mathbb{E}[ \phi(h_{\nu}^{(1)}) ]}{\gamma_{\nu}},
\end{align}
where we have defined the vector $\gamma_{\mu} \equiv 1 + \beta \sigma_{2}^{2} \var[\phi(h_{\mu}^{(1)})]$ for brevity. Thus, in this simple setting, activation functions with $\mathbb{E}\phi(h) \neq 0$ yield qualitatively different behavior from those with $\mathbb{E}\phi(h) = 0$: non-vanishing $\mathbb{E}\phi(h)$ introduces a rank-1 component in the GP kernel, which in turn couples elements of $G_{yy}$ in the leading finite-width correction.

\section{Numerical methods}\label{app:sec:numeric}

In this appendix, we describe the numerical methods used in our experiments. We perform our simulations by sampling network parameters at each time step of the Langevin update \ref{eq:langevin_sampling} after some large burn-in period when the loss function stabilizes around a fixed number. We used Euler-Maruyama method \cite{kloeden1992stochastic} to obtain the discretized Langevin equation:
\begin{align}
    \Theta(t+1) - \Theta(t) = - \beta^{-1} \Theta(t) d t- \nabla_\Theta E(t) dt + \xi \sqrt{2\beta^{-1}dt},
\end{align}
where $\xi \sim \mathcal{N}(0,1)$ is a standard Gaussian random variable sampled i.i.d. at each time step and $dt$ is the time step. The first, second and last terms represent the weight decay, the gradient descent update and the stochastic Wiener process, respectively.

We used the Neural Tangents framework \cite{neuraltangents2020} and PyTorch deep learning library \cite{paske2019pytorch} to generate the neural networks and trained them according to the discretized full-batch Langevin update rule. A typical burn-in time was $\sim 2\times 10^6$ iterations and after that the parameters were sampled over $\sim 2\times 10^6$ iterations where we chose a learning rate of $dt \sim 10^{-4}$. Simulations have been performed on a cluster with NVIDIA Tesla V100 GPU's with 32 GB RAM and a typical simulation run took $\sim 2-6 \text{ hr}$ depending on the architecture and the network width. All code used throughout this work can be reached at \url{https://github.com/Pehlevan-Group/finite-width-bayesian/}.

All figures shown here are results of a single instance of a trained neural network on a fixed dataset. Since we performed all our experiments with $\beta = 1$, we observed that the different initializations of a network did not influence the final posterior mean due to the weight decay and long burn-in periods.

Throughout all experiments, the MNIST digits were downsized from $28\times 28$ pixels to $10 \times 10$ pixels without distorting the original digits. This was done to accelerate the training process since large input dimensions would take an order of magnitude more time to obtain well estimated posterior means. We considered $10$-dimensional outputs corresponding to one-hot encoded digits. Both inputs and labels were ordered according to their class. Figure \ref{fig:mnist_and_cnn1d} shows an example of MNIST digits and the input $G_{xx}$ and output $G_{yy}$ Gram matrices.

\end{document}